\documentclass[11pt,a4paper]{article}

\usepackage[margin=1in]{geometry}
\usepackage{amsmath,amssymb}
\usepackage{bm} 
\usepackage{booktabs}
\usepackage{tabularx}
\usepackage{array}
\usepackage{graphicx}
\usepackage{xcolor}
\usepackage{enumitem}
\usepackage{hyperref}
\usepackage{caption}
\usepackage{subcaption}
\usepackage{pgfplots} 
\usepackage{pdflscape}
\usepackage{longtable}
\usepackage{multirow}
\usepackage{comment}
\usepackage{natbib}

\hypersetup{
  colorlinks=true,
  linkcolor=blue!55!black,
  citecolor=blue!55!black,
  urlcolor=blue!55!black
}

\setlist[itemize]{leftmargin=*,topsep=2pt,itemsep=2pt}
\setlist[enumerate]{leftmargin=*,topsep=2pt,itemsep=2pt}
\renewcommand{\arraystretch}{1.12}

\usepackage[T1]{fontenc}
\usepackage{lmodern}
\usepackage{tikz}
\usetikzlibrary{arrows.meta, positioning, calc, fit, shapes.geometric,shapes.symbols, backgrounds, decorations.pathreplacing}

\usepackage{cleveref}

\usepackage{CJKutf8}
\newcommand{\zhname}{\begin{CJK*}{UTF8}{gbsn}知境\end{CJK*}}
\newcommand{\projectname}{ZenGen (\zhname)}
\newcommand{\projectnameen}{ZenGen}
\newcommand{\modelname}{ZenGen}
\newcommand{\agentname}{Actio}
\newcommand{\datasetname}{SoMBench}
\newcommand{\evalname}{SoMEval}
\newcommand{\evalurl}{https://github.com/Zhijing-AI/SoMEval}

\usepackage{fancyhdr}
\usepackage{graphicx}

\usepackage{titlesec}

\titleclass{\subsubsubsection}{straight}[\subsubsection]

\newcounter{subsubsubsection}[subsubsection]

\renewcommand\thesubsubsubsection
{\thesubsubsection.\arabic{subsubsubsection}}

\titleformat{\subsubsubsection}
{\normalfont\normalsize\bfseries}
{\thesubsubsubsection}{1em}{}

\titlespacing*{\subsubsubsection}
{0pt}{3ex}{1em}

\fancypagestyle{plain}{
    \fancyhf{}
    \fancyhead[L]{\includegraphics[height=16pt]{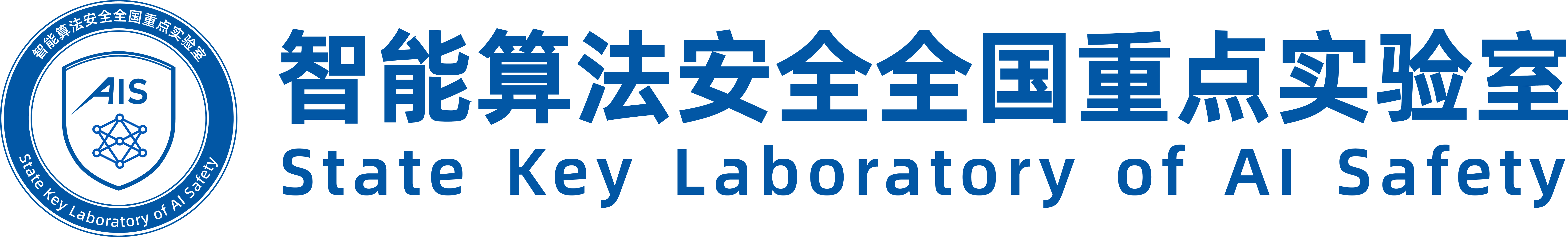}}
    \fancyhead[R]{\small \projectname{} Technical Report}
    \fancyfoot[C]{\thepage}
    
}

\title{\textbf{\projectnameen: Social Mind for LLMs}}
\author{
  \projectnameen{} Team
}
\date{}

\begin{document}
\maketitle

\begin{abstract}

As large language models move from isolated task solving toward long-term service in human environments, they require social intelligence: the ability to infer mental states, track social relations, reason over norms, and adapt behavior under situated context. This report presents \projectname{}, an integrated framework for measuring social intelligence, internalizing social reasoning, and grounding it at deployment time.

For measurement, we introduce \datasetname{}, a psychology-grounded and comprehensive benchmark for social intelligence. It defines social intelligence as a structured capability space, covering 3 primary dimensions, 17 secondary dimensions, and 71 task paradigms. It further controls question format, narrative perspective, and context length across 284 shared scenarios and 3,481 expert-verified instances. Evaluation over 20 representative LLMs shows substantial remaining headroom: the best model reaches only 72.08\% overall accuracy, and none of the 17 secondary dimensions reaches the 90\% near-ceiling band.

For internalization, we develop \modelname{}, a diagnosis-driven training recipe using supervised fine-tuning, on-policy distillation, and rubric-based reinforcement learning. Across five social-cognition benchmarks, \modelname{} consistently improves over its base models, with \modelname{}-27B-Stage2 achieving the best average score among all compared models and \modelname{}-32B-Stage2 showing competitive performance against DeepSeek-V4-Pro.  These results indicate that social reasoning can be strengthened as a stable model capability through diagnosis-driven staged training.

For deployment-time grounding, we build \agentname{}, a harness-controlled inference architecture that wraps a base LLM and routes four typed supports into reasoning: PRISM for procedural guidance, Starling for runtime mental-state representation, SAGE for reusable experience, and gated RAG for external social and normative knowledge. Across five base models and three social-cognition benchmarks, the full harness improves 14 of 15 model-benchmark pairs and is best or tied for best in 8 pairs, showing that typed runtime support can systematically strengthen social reasoning at inference time. Module-level analyses attribute these gains to selective activation of complementary supports, establishing \agentname{} as an effective path for grounding social reasoning at deployment time.

Together, these results show that social mind LLMs require coordinated progress in evaluation, parametric internalization, and deployment-time grounding.


\end{abstract}

\begin{center}
    \includegraphics[width=\linewidth]{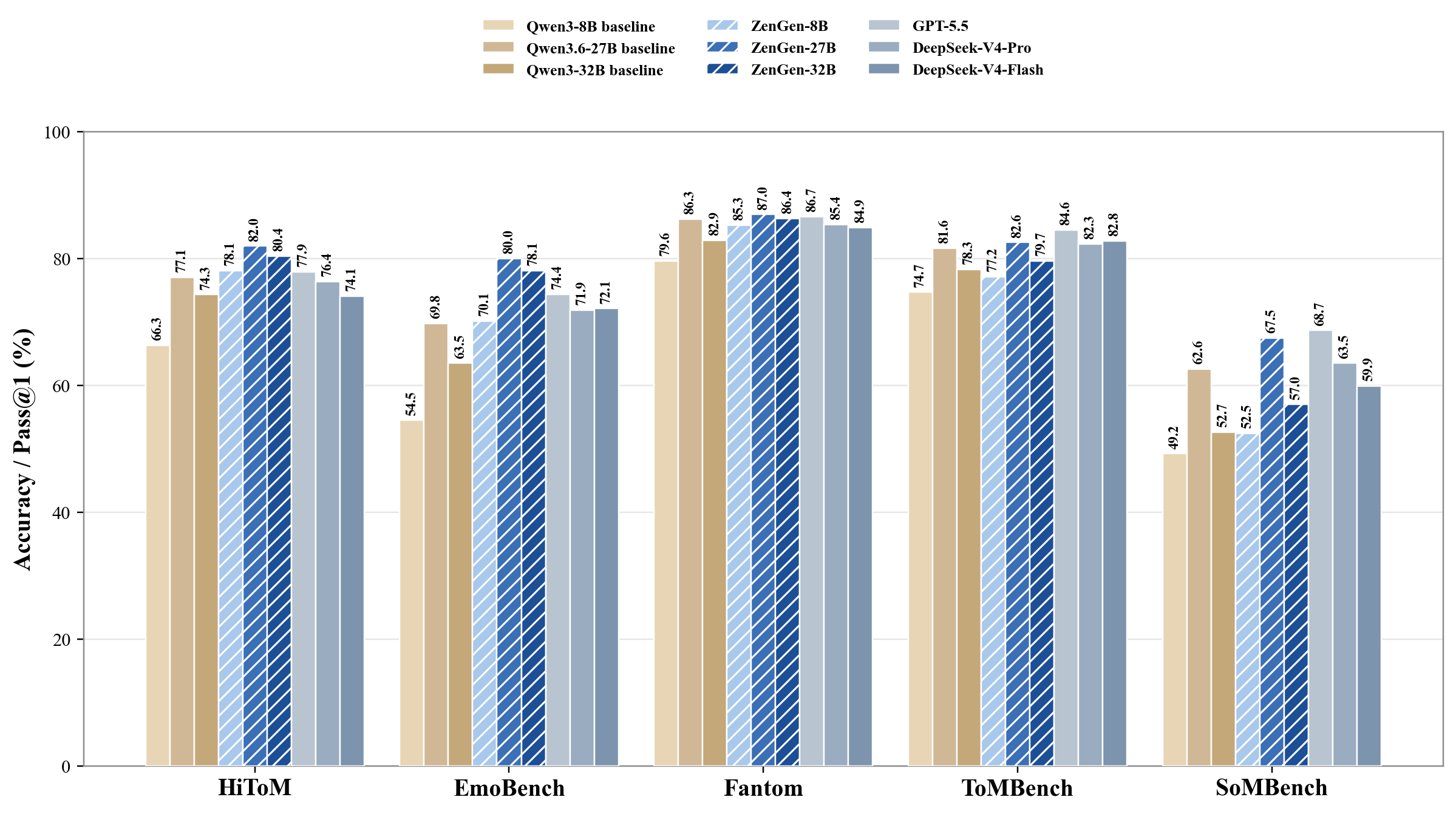}
    \captionof*{figure}{Benchmark performance of \projectname{}-series and baselines}
\end{center}

\section{Introduction}
\label{sec:introduction}

Modern large language models are increasingly used, and often envisioned, as a path toward general-purpose artificial intelligence. This vision has been supported by a clear capability progression. Early LLMs demonstrated strong \emph{language intelligence} through fluent generation, knowledge recall, and instruction following; more recent models have moved toward \emph{task intelligence}, showing increasingly strong abilities in reasoning, coding, planning, and tool use.

Now consider a stronger deployment scenario. An LLM is no longer a passive assistant that answers isolated questions, but a long-term collaborator embedded in open social environments. It must interact with users, groups, and other agents while reading social cues such as implicit goals, emotions, relationships, and norms. It must also act under incomplete information and evolving context. In such a setting, task intelligence alone is insufficient. A response can be factually correct but socially inappropriate; an action can be efficient but unsafe; a short-term objective can be satisfied while long-term trust is damaged. The central challenge therefore shifts from solving tasks to acting appropriately, reliably, and safely in social contexts.

This reveals a missing layer of intelligence. Just as embodied agents must understand physical structure and dynamics to interact with the physical environment, LLMs must understand mental states, social relations, and normative expectations to interact with human environments. We call this capability \emph{social intelligence}: the capacity to reason about people and social contexts, and to use such reasoning to guide context-sensitive behavior. Social intelligence is therefore a core foundation for \emph{social mind large language models}, which aim to move LLMs from task executors toward socially aware collaborators.

Recent work has begun to evaluate and improve this capability. On the evaluation side, theory-of-mind and social reasoning benchmarks have expanded from classical false-belief or story-based tasks to more systematic settings, including controlled belief-reasoning problems in MindGames~\cite{mindgames}, higher-order recursive belief reasoning in HI-TOM~\cite{hitom}, information-asymmetric conversational reasoning in FANToM~\cite{fantom}, multi-dimensional evaluation in ToMBench~\cite{tombench}, and open-ended role-play interaction in SOTOPIA~\cite{sotopia}. On the enhancement side, researchers have explored instruction tuning, chain-of-thought prompting, self-consistency, preference optimization, behavior cloning, and reinforcement-style training with model-based feedback; SOTOPIA-$\pi$, for example, improves social agents through behavior cloning and self-reinforcement on filtered interaction trajectories~\cite{sotopiapi}.

Despite this progress, the social intelligence of current LLMs remains partial and fragile. Existing models may perform well on static theory-of-mind questions, social commonsense benchmarks, or role-playing prompts, suggesting a degree of local emergence. Yet such performance often reflects surface-level social cognition: it can rely on semantic pattern matching and degrade sharply once the setting becomes dynamic, interactive, or long-horizon. This brittleness is not only a model-level limitation; it also reflects limitations in how social intelligence is currently evaluated, trained, and supported at deployment time. Existing benchmarks remain fragmented, model enhancement methods often lack fine-grained supervision over the reasoning process, and many social intelligence tasks require psychological priors, external norms, and historical context that are difficult to store reliably in model parameters alone.

These limitations point to a more fundamental question: how can an LLM move from task execution to socially situated intelligence, where it can understand people, reason about social context, and act appropriately in open-ended environments? Answering this question requires treating social intelligence as a layered capability rather than a single skill or benchmark score. At the evaluation level, it must be measurable: we need to identify which social abilities are present, missing, or brittle across different contexts. At the model level, it must be internalized: social reasoning should become a stable capability of the base model rather than only a prompt-induced behavior. At the deployment level, it must be grounded: social reasoning should remain anchored in relevant knowledge, context, and feedback from the environment.


Guided by this view, we organize our work, namely \projectname{}, around three complementary components. For evaluation, we construct the psychology-grounded benchmark \datasetname{}, which defines a structured capability space for social intelligence and supports controlled, metadata-based diagnosis. For model training, we develop \modelname{}, a staged training recipe that uses capability diagnosis to construct targeted supervision, followed by supervised fine-tuning, on-policy distillation, and rubric-based reinforcement learning to strengthen social reasoning. For deployment-time reasoning, we design \agentname{}, a harness-controlled architecture that routes typed supports into inference, including procedural guidance, runtime mental-state representation, reusable experience, and external social and normative knowledge. Together, these components connect capability diagnosis, parametric learning, and deployment-time grounding into one technical route toward social mind LLMs.
  
\paragraph{Contributions and Takeaways.}

This report offers three main takeaways.

  \begin{itemize}
    \item \textbf{A psychology-grounded taxonomy supports more diagnostic social-intelligence evaluation.}
    \datasetname{} provides a structured capability space for evaluating social intelligence and shows that current
    LLMs still have substantial headroom: the best model reaches only 72.08\% overall accuracy, and none of the 17
    secondary dimensions reaches the 90\% near-ceiling band.

    \item \textbf{Social reasoning can be internalized through diagnosis-driven supervision and staged training.}
    The cross-scale results of the \modelname{} family show consistent gains over base models, with
    \modelname{}-27B-Stage2 achieving the best average score among all compared models and \modelname{}-32B-Stage2
    showing competitive text-only performance against DeepSeek-V4-Pro.

    \item \textbf{Deployment-time grounding benefits from selective routing over typed supports.}
    The module-level behavior of \agentname{} shows that effective grounding is not achieved by uniform context
    expansion, but by activating complementary supports according to the reasoning needs of each case. Across five
    base models and three benchmarks, the full harness improves 14 of 15 model--benchmark pairs and is best or tied
    for best in 8 pairs.
  \end{itemize}

\section{Technical Overview}
\label{sec:technical-architecture}

The following sections develop the three components of \projectname{} introduced above as the main technical body of this report. The evaluation line starts from the need to keep the evaluated capability identifiable throughout benchmark construction and scoring. \datasetname{} formulates this as three principles---capability-grounded design, controlled evidence and perspective, and diagnostic difficulty with traceability---and implements them through taxonomy-guided shared scenarios, controlled rewriting and filtering, expert verification, and metadata-preserved evaluation. The training line focuses on progressive internalization under a structured and uneven capability space. \modelname{} uses capability diagnosis to drive data iteration, and uses a staged training recipe to move from broad ToM foundations toward specialized social-mind reasoning. The deployment-time line starts from a different requirement: social reasoning at inference time needs controllable support, not merely a larger prompt. \agentname{} wraps a frozen base model in a harness-controlled runtime that decides when procedural guidance, attributed state, experience, or external social knowledge should enter the context, checks the resulting reasoning, and records execution traces for later inspection.

\begin{figure}[t]
  \centering
  \includegraphics[width=\linewidth]{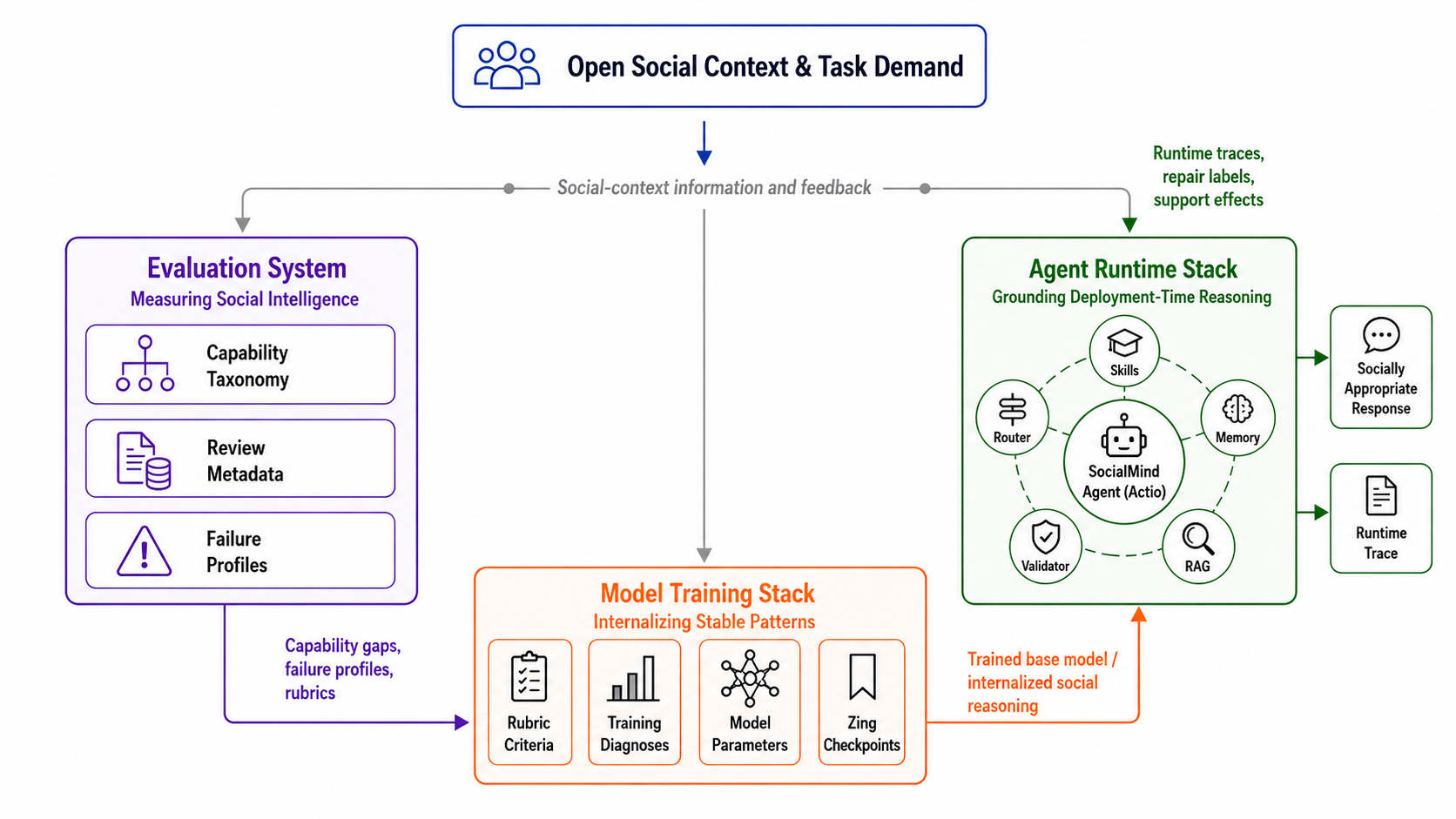}
  \caption{Technical overview of \projectname{} across evaluation, training, and deployment-time grounding, highlighting the main artifacts exposed by each implementation line.}
  \label{fig:technical-architecture}
\end{figure}

Figure~\ref{fig:technical-architecture} summarizes these three implementation lines. They are connected through the diagnostic artifacts they produce and reuse: benchmark labels and failure profiles can inform training data construction and agent support selection; training checkpoints and failure analyses can guide later evaluation and deployment; and agent execution traces can reveal context-sensitive failures that are useful for further analysis.


\section{\datasetname{}: Measuring Social Intelligence}
\label{sec:socialmind}

\subsection{Motivation and Benchmark Principles}
\label{sec:socialmind-motivation}

For human-involved applications, language models need to infer people's mental states, including their knowledge, beliefs, desires, intentions, and emotions, while simultaneously tracking shifts in interpersonal relationships and information states throughout an interaction. They must also judge how roles, power, group boundaries, and social norms constrain possible actions. These interconnected demands define this chapter’s evaluation objective: determining whether a model possesses a usable ``social mind'', operationalized here as the capacity to reason over mental states, interactional dynamics, and norm-conditioned judgment in situated contexts.

Existing theory-of-mind benchmarks have pinpointed essential dimensions of this capacity, ranging from false-belief attribution, higher-order belief reasoning, information-asymmetry tracking, emotion and desire inference, and faux-pas recognition \cite{tom,frith,hitom,fantom,tombench}. These tasks are valuable because mental-state attribution is foundational to social cognition. Nevertheless, these tests remain predominantly decontextualized mentalizing exercises.. They typically probe whether a model can correctly identify a belief or emotion in a brief narrative vignette, yet they rarely evaluate whether that inferred mental state can subsequently inform adaptive behaviors, such as strategic action, conflict de-escalation, trust-building, deception detection, or norm- constrained judgment in an ongoing social situation.

Benchmarks for social reasoning and social interaction, spanning social commonsense question answering, role-play, and interactive agent evaluation, have moved closer to ecologically realistic use cases \cite{socialiqa,sotopia,sotopiapi}. Yet, they introduce a limitation of a different kind: their performance metrics typically hinge on global outcomes, such as whether an agent completes a goal, delivers a socially acceptable response, or sustains cooperative dialogue. While practically informative, these aggregate scores conflate multiple underlying faculties into a single measure.A model may earn credit for being agreeable, cautious, or generically prosocial, even when it has not accurately represented the relationship structure, information asymmetries, or the strategic incentives in the scenario. As a result,it becomes difficult to diagnose whether a failure stems from misunderstanding the situation, choosing a ill-suited interaction strategy, or misapplying  a social norm.

Norm and value benchmarks supply a vital ethical component, covering moral acceptability, rules of thumb, situated social norms, and action-consequence reasoning \cite{ethics,socialchemistry,normbank,moralstories}. Yet they often cast social norms as static classification targets: asking whether an action is acceptable, a rule is violated, or option is morally superior. In interactive settings, however, norm reasoning is far less fixed: the identical action can take on entirely different significance depending on participants' identities, roles, relationship histories, power imbalances, institutional contexts, or group affiliations \cite{cialdini,foucault,scott}. A robust benchmark for social mind must therefore go beyond assessing norm knowledge; , it must test whether a model can balance competing norms under relational and perspectival constraints.


Collectively, these limitations point to a foundational gap. What is needed is not simply a larger pool of social questions, but a benchmark that defines the target social capabilities before item construction and preserves those definitions throughout evaluation. Such a benchmark must pre-specify the capability under measurement, tightly constrain the evidence and perspective available in each scenario, and retain the metadata required to trace model failures back to interpretable social-cognitive mechanisms. Following this rationale, \datasetname{} f is organized around three principles: capability-grounded design, controlled evidence and perspective, and diagnostic difficulty with traceability.
  
The first principle is \textbf{capability-grounded design}. Many social benchmarks often blend task formats, topics, and capability targets, making it difficult to interpret whether a correct or incorrect response actually reflects a specific social cognitive faculty. To address this, \datasetname{} anchors itself in a capability taxonomy drawn from theory-of-mind research, social-interaction theory, and norm reasoning, further informed by established inventories of mentalizing measures \cite{atoms}, social-emotional-behavioral skill taxonomies \cite{bessi}, and pragmatic and cooperative interaction theory \cite{searle,grice,axelrod}. It comprises three primary dimensions, 17 secondary dimensions, and 71 fine-grained task paradigms. For every paradigm, we predefine the target construct, the evidence the scenario must supply, the distractor that a fluent but socially weak model is prone to follow, and the expected failure mode. By specifying the target construct upfront, we ensure that every item remains tied to a well-defined capability, rather than drifting into an open-ended collection of plausible social stories.


The second principle is \textbf{controlled evidence and perspective}. Social reasoning is inherently contingent on both the evidence available and the vantage point adopted: the same action or utterance may warrant divergent interpretations depending on which information is accessible and whose epistemic state anchors the reasoning problem. To address this, \datasetname{} organizes each test case around a shared social scenario, then poses multiple question types over that identical context, while also varying the framing between first-person and third-person perspectives. As the reasoning target shifts across questions, the shared context holds the evidentiary basis constant; simultaneously, the perspective manipulation specifies whether the model must reason from a situated participant's knowledge boundary or from an observer's view over multiple agents. This design isolates failures in the queried reasoning operation from changes in scenario content, and renders perspective leakage empirically detectable when a model exploits information unavailable to the relevant character.
  

The third principle is \textbf{diagnostic difficulty and traceability}. A benchmark for the social mind should do more than simply test model accuracy; its primary purpose is to pinpoint specific capability deficits. For difficulty to be genuinely diagnostic, it must stem from the intended social-cognitive challenge itself, rather than from confusing wording, incomplete evidence, or ambiguous answers. To this end, \datasetname{} implements a multi-stage construction pipeline that combines human-in-the-loop seed generation, controlled context and question rewriting, model-based filtering, and final human verification. This process retains items that stress the target capability while remaining clearly answerable, contextually grounded, and perspective-consistent. In addition, each item annotated with detailed construction and evaluation metadata, making it possible to trace model errors back to interpretable mechanisms, such as knowledge-boundary leakage, failures in deception recognition, norm conflicts, or misjudgments arising from power asymmetries.

\subsection{Benchmark Construction Overview}
\label{sec:socialmind-overview}

\begin{figure}[!h]
  \centering
  \includegraphics[width=\linewidth]{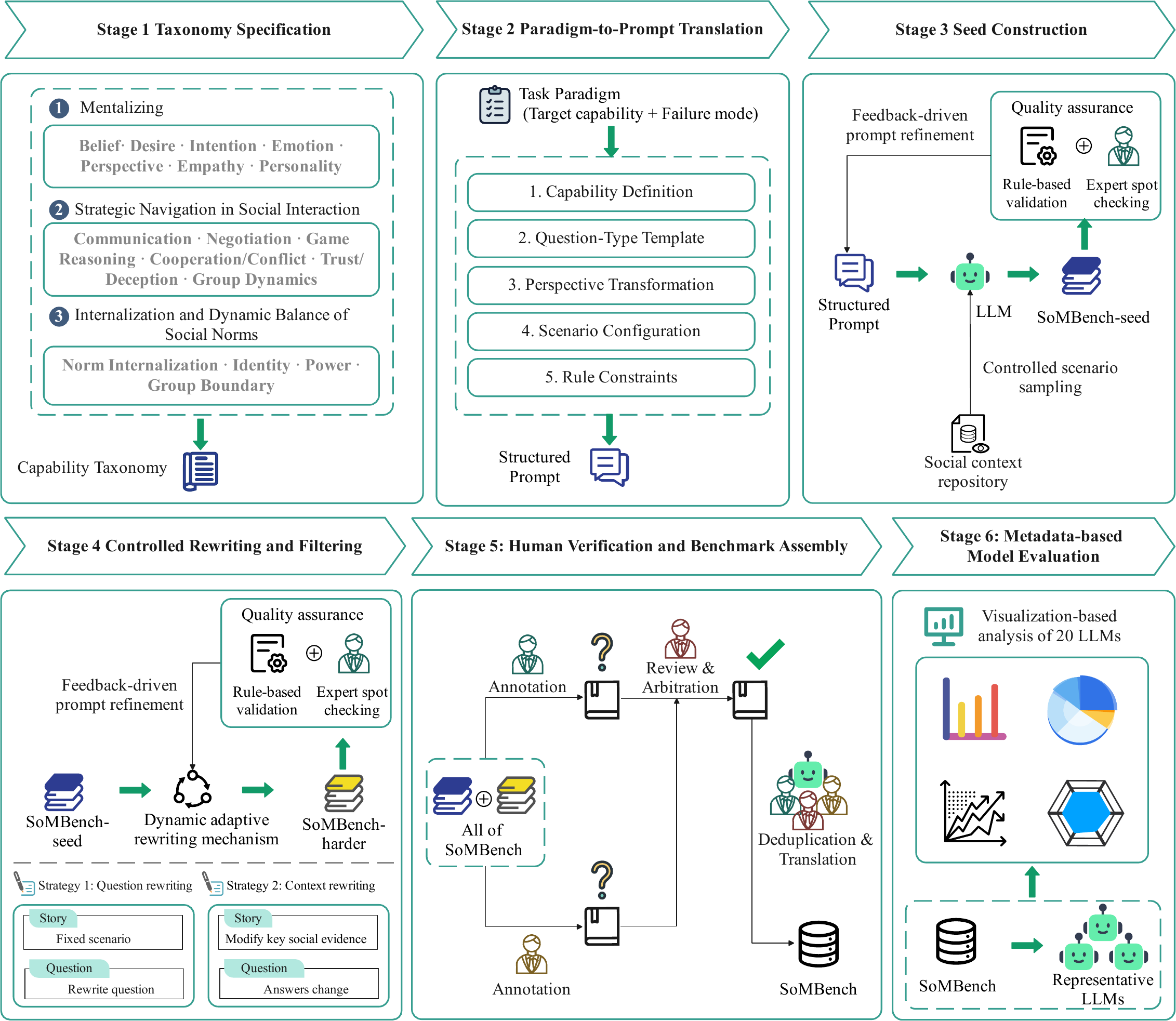}
  \caption{Overview of the \datasetname{} benchmark workflow. The workflow starts from a capability taxonomy specification, translates task paradigms into seed-generation prompts, constructs shared-context seed cases, creates and filters harder variants through controlled rewriting, verifies and assembles the benchmark, and finally evaluates representative LLMs with metadata-based diagnostic analysis.}
  \label{fig:data-pipeline}
\end{figure}

We build \datasetname{} through a workflow that explicitly connects benchmark construction with metadata-based evaluation, as depicted in Figure~\ref{fig:data-pipeline}. The process comprises six sequential stages: the first five handle the construction, hardening, verification, and assembly of the dataset, while the final stage uses the completed benchmark and its metadata to evaluate representative LLMs. In this subsection, we provide a chapter-level overview of each stage, and Table~\ref{tab:principle-stage} summarizes how the three principles introduced in Section~\ref{sec:socialmind-motivation} are instantiated across the workflow.

\paragraph{Stage 1: taxonomy specification.}
The workflow starts by defining the capability space. We structure social mind domain into three primary dimensions, 17 secondary dimensions, and 71 fine-grained task paradigms. Each paradigm specifies the target construct, required evidence, distractor logic, and expected failure mode, which collectively establish the measurement specification for all subsequent stages.

\paragraph{Stage 2: paradigm-to-prompt translation.}
Each task paradigm is then operationalized as a set of seed-generation prompts. This stage turns the abstract taxonomy into concrete operational constraints, covering required social evidence, question format, perspective setting, answer boundaries, distractor rules, and exclusion criteria. The goal is to ensure that every prompt faithfully embodies the measurement intent of the taxonomy before any candidate item is generated.

\paragraph{Stage 3: seed construction.}
The workflow next generates initial seed cases under the specified constraints. Each seed is anchored in a shared social scenario, incorporates multiple question types, maintains controlled context length, and adopts either first- or third-person framing. A human-in-the-loop review then verifies that each candidate remains faithful to its intended paradigm, supplies sufficient evidentiary grounding, avoid ambiguous answers, and respects the specified perspective.

\paragraph{Stage 4: controlled rewriting and filtering.}
Seed cases are subsequently expanded into more challenging variants through two targeted operations: context rewriting and question rewriting. The former modifies the scenario evidence while preserving the question surface; the latter keeps the scenario fixed while altering the queried reasoning operation. The resulting variants are then passed through model-based filtering, which retains only those that effectively stress the target capability while remaining answerable, internally coherent, and stable on the non-target questions.

\paragraph{Stage 5: human verification and benchmark assembly.}
Retained cases are subjected to a comprehensive verification process covering evidentiary grounding, answer determinacy, distractor validity, paradigm alignment, perspective consistency, rewrite coherence, and safety compliance. Verified cases are then assembled into the final benchmark, with metadata attached to each item, including capability labels, question type, perspective, construction source, rewrite mode, difficulty information, and verification status.

\paragraph{Stage 6: metadata-based model evaluation.}
The final stage marks the transition from construction to evaluation, deploying the completed benchmark to assess representative LLMs. The accompanying metadata enables fine-grained performance disaggregation across primary and secondary dimensions, task paradigms, question types, perspectives, construction sources, and rewrite mode. This design ensures that evaluation outputs support targeted failure diagnosis rather than devolving into aggregate ranking alone.

\begin{table}[t]
  \centering
  \small
  \renewcommand{\arraystretch}{1.22}
  \begin{tabularx}{\linewidth}{@{}p{3.1cm}p{2.0cm}X@{}}
  \toprule
  \textbf{Design principle} & \textbf{Relevant stages} & \textbf{Implementation in the workflow} \\
  \midrule
Capability-grounded design &
  Stages 1--2, 5 &
  Defines task paradigms before generation, translates them into prompts, and verifies paradigm alignment. \\
  Controlled evidence and perspective &
  Stages 2--5 &
  Controls evidence, question format, context length, narrative viewpoint, rewriting changes, and perspective
  consistency. \\
  Diagnostic difficulty and traceability &
  Stages 4--6 &
  Creates harder variants, filters and verifies valid difficulty, assembles metadata, and uses it for diagnostic
  model evaluation. \\
\bottomrule
  \end{tabularx}
  \caption{How the workflow stages operationalize the three benchmark principles.}
  \label{tab:principle-stage}
\end{table}

The rest of the chapter walks through this pipeline. We begin in Section~\ref{sec:socialmind-taxonomy} by laying out the capability taxonomy. Section~\ref{sec:socialmind-construction} then covers the shared-context design, seed construction, controlled rewriting, and model-based filtering. In Section~\ref{sec:socialmind-protocol}, we present benchmark verification procedures, quality control measures, and the final benchmark statistics. Section~\ref{sec:socialmind-findings} closes with the evaluation protocol and an interpretation of model results.

\subsection{Taxonomy}
\label{sec:socialmind-taxonomy}

\begin{figure}[!h]
    \centering
    \includegraphics[width=\linewidth]{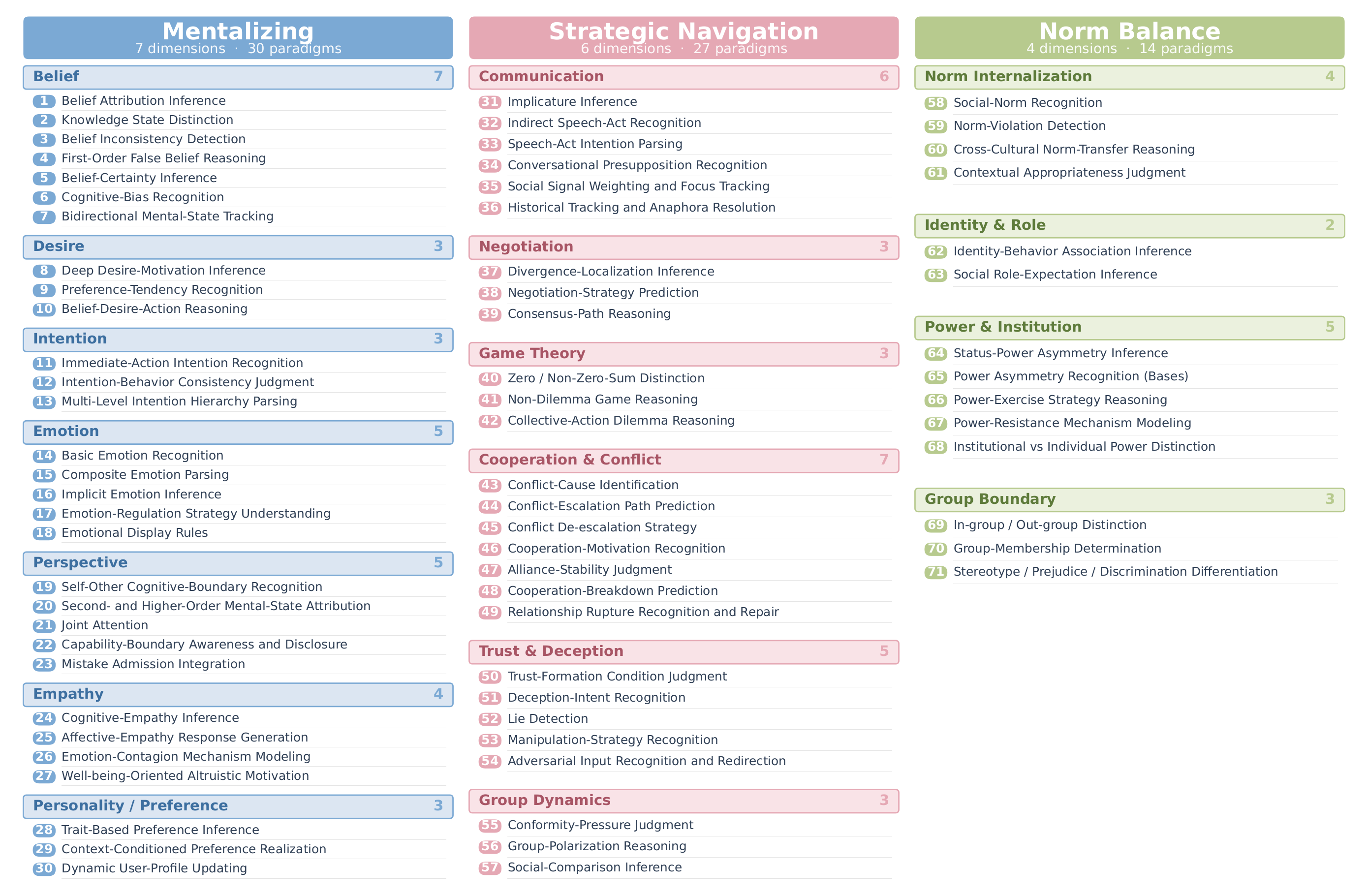}
    \caption{\textbf{A three-level overview of the \datasetname{} taxonomy.} The
    figure presents the complete structure, in which 3 first-level dimensions
    are refined into 17 second-level dimensions and further into 71 task
    paradigms. Under each second-level dimension, the figure lists the
    fine-grained task paradigms it contains.}
    \label{fig:framework_overview}
\end{figure}

The social mind refers to the suite of capacities enabling individuals to infer mental states, coordinate social interaction, and reason about norms in situated contexts. Although anchored in individual cognition, these capacities extend across interpersonal relationships and cultural symbols, rendering the construct fundamentally relational and multi-scalar~\cite{buber1923ithou}. Consistent with this view, the taxonomy presented below serves as a measurement map rather than a list of surface-level tasks, which systematically decomposes the domain into 3 primary dimensions, 17 secondary dimensions, and 71 granular task paradigms, thereby establishing a unified coordinate framework of the target abilities. Figure~\ref{fig:framework_overview} provides an overview of this structure. 

These three first-level dimensions are \textbf{Mentalizing}, \textbf{Strategic Navigation in Social Interaction}, and \textbf{Internalization and Dynamic Balance of Social Norms}, corresponding to three complementary levels of analysis: individual mental-state representation, multi-agent interaction, and norm-conditioned social judgment. This tripartite structure follows Doise's classical levels of analysis in social psychology~\cite{doise1986levels}. At the micro, intra-mental level, inquiry centers on how one mind represents the mental states of itself and others. At the meso, inter-mental level, it concerns how multiple minds interact strategically in dynamic processes. At the macro, extra-mental level, it examines how external social structures and rules shape cognition, and in turn internalized by the mind. Crucially, these levels are mutually constitutive: mental-state attribution abstracted from interaction and norms is incomplete, just as interactional or normative reasoning detached from mental-state tracking lacks a genuine cognitive foundation.

\paragraph{Mentalizing}
corresponds to the micro intra-mental level. It captures the ability to infer unobservable mental states in others and in oneself~\cite{schaafsma2015deconstructing, happe2017structure, schurz2021toward, mcdonald2024hits}. Its core diagnostic challenge is twofold: testing whether a model can distinguish what it knows from what another agent knows, and whether it can still correctly track another's belief when reality and that belief diverge, rather than merely responding to the literal content of an utterance. As the cognitive bedrock of the social mind, mentalizing underpins all higher-order strategic and normative reasoning. Without stable mental-state representations, strategic choices and normative trade-offs in interaction would lose the very objects they act upon. This dimension is further organized into 7 second-level subdimensions, namely belief, desire, intention, emotion, perspective, empathy, and personality/preference, which encompass 30 task paradigms.

\paragraph{Strategic Navigation in Social Interaction}
corresponds to the meso inter-mental level, capturing the ability to select and adapt social actions within evolving multi-agent interactions~\cite{tomasello2009why, sotopia, mathur2024advancing}. It covers implicature in communication, concession and trade-off in negotiation and bargaining, and relational judgment in cooperation and conflict, trust and deception, and group dynamics. This dimension addresses a major limitation of existing evaluations, which tend to treat social interaction as a monolithic category and rely on coarse, outcome-oriented overall scores. Such a broad-brush approach neither isolates the specific strategic primitives nor resists the influence of prosocial bias, which allows models to achieve inflated performance through avoidance rather than genuine strategic engagement~\cite{zhou2025socialeval, huang2025sibench, chen2024socialbench}. To address this, our framework explicitly decomposes the general ``interaction'' label into 6 second-level dimensions, namely communication, negotiation, bargaining, cooperation and conflict, trust and deception, and group dynamics, supporting 27 task paradigms in total.

\paragraph{Internalization and Dynamic Balance of Social Norms}
corresponds to the macro extra-mental level, addressing how external structures and rules shape cognition, how such norms are internalized, and how individuals weigh conflicting norms in concrete situations~\cite{socialchemistry, normbank, bessi}. The target is not generic safety alignment or moral preference classification, but context-conditioned norm judgment. Specifically, a model's capacity to weigh competing expectations such as honesty versus politeness, loyalty versus fairness, or authority versus autonomy within a given social situation. These judgments are further conditioned by identity, role, institutional position, power relation, and group boundary. To capture this complexity, this dimension is organized into 4 second-level dimensions, namely norm internalization, identity and role, power and institution, and group boundary, comprising 14 task paradigms in total.

The 3 primary dimensions and 17 secondary dimensions define the broad conceptual space of the social mind. For benchmark construction, however, a finer-grained unit of analysis is required. Accordingly, we decompose the taxonomy into 71 specific task paradigms. Rather than serving as surface-level question formats, these paradigms function as construct-isolation units: each prescribes both the target capability under examination and the evidentiary conditions necessary to ensure answer determinacy.

Each paradigm is defined by a uniform quadruple: target construct, required evidence, distractor logic, and expected failure mode. The target construct specifies the latent social-cognitive capability being measured. The required evidence specifies what information the scenario must furnish to enable a meaningful test of that capacity. The distractor logic, in turn, outlines a plausible but flawed reasoning path, such as collapsing higher-order mentalizing into first-order reasoning or retreating to the literal semantics of an utterance, that a linguistically fluent but socially shallow model is prone to follow. Finally, the expected failure mode records the interpretable signature of the resulting error.

This quadruple turns the taxonomy into an operational specification for data construction. It instructs generation prompts on the necessary evidence, guides rewriting and filtering toward challenging the predicted failure pattern, and establishes the criteria for human verification regarding alignment and answerability. Crucially, it also makes evaluation results interpretable: an incorrect answer can be traced to a specific predicted failure mechanism rather than educed to mere binary errors. Due to space constraints, we do not enumerate all 71 paradigms. Table~\ref{tab:quadruple} provides representative examples.

\begin{table}[!h]
    \centering
    \caption{Examples of task paradigms defined by the quadruple.}
    \label{tab:quadruple}
    \small
    \renewcommand{\arraystretch}{1.3}
    \begin{tabularx}{\textwidth}{@{}p{2.3cm}XXXX@{}}
        \toprule
        \textbf{Paradigm} & \textbf{Construct} & \textbf{Evidence} & \textbf{Distractor} & \textbf{Diagnosed Failure} \\
        \midrule
        1.5.2 Second-order and higher-order mental attribution
        & Representing ``A thinks that B thinks X'' and higher-order recursive minds\textsuperscript{1}
        & Tracking nested beliefs at second order or above to predict behavior
        & Collapsing to first order, judging only whether ``X is true''
        & Level collapse; loss of recursive depth \\
        \addlinespace
        2.1.1 Inference of implicit meaning
        & Identifying implicature beyond the literal via the cooperative principle\textsuperscript{2}
        & Recovering intent from a maxim violation, answering by implied meaning
        & Retreating to literal semantics, echoing the surface proposition
        & Literalization collapse; implicature confused with literal meaning \\
        \addlinespace
        3.1.2 Detection of norm violation
        & Judging a violation and its domain and severity\textsuperscript{3}
        & Locating the violation's domain first, then judging severity and changeability
        & Homogeneous criticism, or elevating a conventional/personal matter to a moral violation
        & Domain confusion; homogenized moralization \\
        \bottomrule
    \end{tabularx}
    \vspace{2pt}
    {\footnotesize
    \begin{flushleft}
        \textsuperscript{1}\,Anchored in Wimmer \& Perner's second-order false belief~\cite{wimmer1983beliefs}.\\
        \textsuperscript{2}\,Anchored in Grice's cooperative principle~\cite{grice}.\\
        \textsuperscript{3}\,Anchored in Turiel's social domain theory~\cite{turiel1983development}.
    \end{flushleft}}
\end{table}

This taxonomy supplies the control specification for data construction: generation prompts, rewriting operations, filtering criteria, and metadata labels are all anchored to these paradigm definitions.

\subsection{Data Construction}
\label{sec:socialmind-construction}

\begin{figure}[!h]
    \centering
    \includegraphics[width=\linewidth]{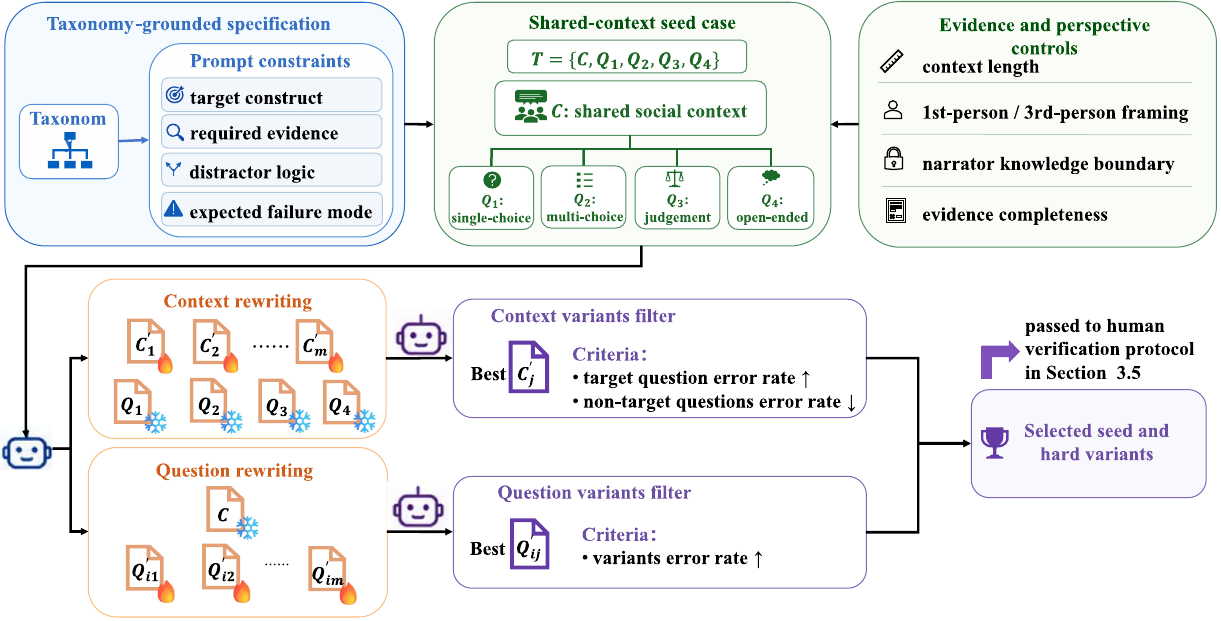}
    \caption{\textbf{Overview of the \datasetname{} data-construction pipeline.}
    First, taxonomy-grounded specification derives construction constraints from fine-grained paradigms. Second, a shared-context seed case is constructed, where the same social context supports single-choice, multiple-choice, judgment, and open-ended questions. Third, controlled variant construction produces two types of candidates. Finally, model-based filtering selects seed cases and harder variants according to failure patterns, validity, evidence support, and paradigm alignment,}
    \label{fig:data_construction_pipeline}
\end{figure}

\datasetname{} operationalizes the taxonomy from Section~\ref{sec:socialmind-taxonomy} as concrete test cases through a carefully controlled construction pipeline. The pipeline is anchored in a shared-context test-case schema, where each social scenario supports multiple question types under explicit evidence and perspective controls. Building on this schema, we generate taxonomy-grounded seed cases, derive more challenging variants through context and question rewriting, and apply model-based filtering to retain diagnostically useful cases. The selected seed and its variants are then passed to the human verification protocol described in Section~\ref{sec:socialmind-protocol}. Figure~\ref{fig:data_construction_pipeline} provides an overview of this data-construction component.

\paragraph{Shared-context test-case schema.}
The shared-context test case constitutes the basic unit of \datasetname{}. Instead of posing isolated questions, each case anchors multiple question types within a single social scenario. By holding characters, relationships, timelines, and task-relevant evidence constant, the shared context minimizes scenario-level confounds and enables direct comparisons across question types, allowing performance differences to be  more cleanly attributed to the specific reasoning operation under assessment. Formally, each test case is denoted as
\[
T=\{C,Q_1,Q_2,Q_3,Q_4\},
\]
where $C$ denotes the shared narrative context, and $Q_1$--$Q_4$ are four questions derived from it. $Q_1$ is a \emph{single-choice} question that probes focused discrimination among mutually exclusive interpretations or actions. $Q_2$ is a \emph{multiple-choice} question that requests all applicable interpretations, evidence types, or social-cognitive conditions. $Q_3$ is a \emph{judgment} question requiring the model to assess whether a stated claim is supported by the context and to provide a rationale. $Q_4$ is an \emph{open-ended} question that elicits a synthesized response, such as an explanation, a predicted social trajectory, or a proposed interaction strategy. Because all four questions share the same context $C$, performance differences across question types can be interpreted as reflecting variation in reasoning demand rather than differences in the underlying scenario.

\paragraph{Evidence and perspective controls.}
Shared contexts are further governed by two construction parameters: context length and narrative perspective. Context length modulates the quantity and distribution of available evidence, ranging from brief vignettes to extended scenarios involving multiple agents and complex timelines. Narrative perspective, in turn, determines whose epistemic state anchors the correct response: third-person contexts p offer an observer's view, while first-person contexts impose a narrator knowledge boundary. Together, these parameters render evidence availability explicit and enforceable: a first-person case is invalid if it relies on information unavailable to the narrator, and a long-context case is invalid if the task-relevant evidence is missing, contradictory, or falls outside the intended reasoning trajectory.

\paragraph{Taxonomy-grounded seed construction.}
Seed construction instantiates each fine-grained paradigm as a shared-context test case. For each paradigm, we build a modular generation prompt that specifies the target construct, required evidence, distractor logic, question-type constraints, perspective, context length, answer format, and exclusion criteria. Guided by these specifications, the generator produces a complete case with context, four questions, reference answers, supporting evidence, and rationales, so that the gold answer is verifiable as a coherent reasoning path rather than a bare categorical label. At this formative stage, domain reviewers focus on prompt refinement rather than final case acceptance. They inspect candidates for paradigm drift, evidential insufficiency, ambiguous answers, shallow lexical cues, weak distractors, and perspective violations. Recurring issues are fed back into the prompt and validation rules before the subsequent generation rounds. The resulting output, \datasetname{}-original, comprises taxonomy-aligned, evidence-grounded seed cases that are well positioned for the controlled rewriting phase.

\paragraph{Controlled variant construction.}
To generate difficult test cases and explore the mental weaknesses of LLMs, we perform an adaptive rewriting process~\cite{li2026deepbias}.
Specifically, once seed cases establish an answerable, evidence-grounded, and taxonomy-aligned foundation, controlled rewriting introduces further diagnostic pressure through systematic manipulation. The two passes differ in what remains invariant: context rewriting keeps the questions fixed and modifies the scenario evidence, whereas question rewriting preserves the context and modifies the queried reasoning operation.

Context rewriting tests whether a model can adjust its reasoning in response to altered social evidence rather than reusing the answer pattern of the seed case. Given a seed $T$, we select one target question $Q_t$ from $Q_1$--$Q_3$ and prompt the generator to produce a new context $C'$ such that the correct option for $Q_t$ shifts to a different choice, while the answers to the non-target questions stay as stable as possible. With the questions and options held fixed, difficulty is introduced solely through modifications to the scenario evidence by weakening direct cues, adding plausible distractors, or introducing more subtle social-cognitive clues. We restrict this pass to $Q_1$--$Q_3$, as their discrete answers formats enable straightforward automatic comparison, and we generate roughly 15 to 20 candidate variants per seed.

Question rewriting targets the complementary weakness: a valid context may contain questions that are overly broad, too local, or solvable through lexical overlap and conspicuous distractors. In this pass, the context remains fixed while underperforming questions are re-authored within the same format and paradigm. The new questions are not not paraphrases or expanded distractor sets, but freshly constructed items  whose difficulty arises from mechanisms such as multi-cue integration, finer distinctions between adjacent interpretations, stricter perspective tracking, and resistance to shallow shortcuts. Each candidate is accompanied by structured answer information and supporting evidence, allowing for a preliminary validation that screens out formatting errors, unsupported or ambiguous answers, evidence mismatches, and paradigm drift prior to model evaluation.

\paragraph{Variant filtering and selection.}
Since the rewriting phase generates an excess of candidates, filtering is used to retain variants that exhibit diagnostic difficulty rather than spurious instability. For context-rewritten cases, we prefer variants that elicit failure on the target question while preserving stability on non-target questions. To operationalize this, we evaluate each variant across several models and compute two metrics: a target-question wrong rate and a non-target wrong rate. The filtering score accordingly rewards target-item errors and penalizes excessive non-target instability. From the scored pool we select four variants per seed using temperature-controlled softmax sampling rather than a purely greedy top-$k$ rule, ensuring diversity across target questions and rewritten dimensions. For question-rewritten variants, the selection criterion shifts to maximizing  model error while maintaining format validity, evidential grounding, and paradigm alignment. All selected variants, together with the original seed cases, are then forwarded to the human verification stage described in Section~\ref{sec:socialmind-protocol}.

\subsection{Benchmark Verification and Statistics}
\label{sec:socialmind-protocol}

Once candidate instances have been generated, rewritten, and filtered, \datasetname{} applies a crucial human verification stage before final benchmark assembly. This stage serves as the quality gate between automatic construction and meaningful model evaluation, verifying that each candidate satisfies key criteria, i.e., answerability, evidential grounding, alignment with its intended paradigm, and adherence to its designated perspective. This subsection first describes the human review workflow and the criteria applied, then presents the composition of verified benchmark and compares its coverage with existing social-cognition benchmarks.

\paragraph{Human verification workflow.}
Every candidate instance generated by the construction pipeline underwent human verification before inclusion in the final benchmark. We recruited 15 undergraduate and graduate students with psychology backgrounds as reviewers, using a custom web-based annotation platform. Each instance was independently assessed by two reviewers. If the two reviewers reached the same decision, the instance was accepted, marked for revision, or rejected accordingly. Disagreements were resolved by a third reviewer. Instances marked for revision were corrected and rechecked, so that the released benchmark includes only cases that passed the final verification.

\paragraph{Review guidelines and quality criteria.}
Reviewers followed a standardized guideline that combines scenario-level validity review, blind solving, and question-level consistency verification. They first examined whether the scenario was complete, internally consistent, and aligned with the intended difficulty strategy, including the character relationships, timeline, information flow, and task-relevant evidence. For first-person narratives, they additionally checked whether the narrator's knowledge boundary was preserved. Reviewers then answered the questions without access to the reference answers, which helped expose ambiguous wording, unsupported gold labels, and annotation errors. After blind solving, they inspected the full question--answer chain, including the scenario, question, options, rationale, and reference answer. This step checked evidence grounding, answer determinacy, distractor validity, paradigm alignment, and perspective consistency. Rewritten variants received an additional propagation check to ensure that perturbations remained coherent, that unchanged questions stayed valid, and that the claimed answer shifts were justified without downstream contradictions. Finally, reviewers performed safety and compliance checks before issuing the final decision.

\paragraph{Verified benchmark composition and coverage.}
Of the $3,484$ instances generated by the pipeline, the verification process retained $3,034$ with an overall retention rate of $87.08\%$, and identified the remaining instances as requiring rejection, uncertainty resolution, or revision.
Among the non-retained instances, $3$ could not be rewritten and were discarded, while all other instances were revised. As a result, the final benchmark contains $3,481$ expert-verified instances.
The final set covers the full \datasetname{} taxonomy, spanning all 3 primary dimensions, 17 secondary dimensions, and 71 task paradigms. The distribution across the three primary dimensions is as follows: \textbf{Mentalizing} contains $1,452$ instances, \textbf{Strategic Navigation in Social Interaction} has $1,325$ instances, and \textbf{Internalization and Dynamic Balance of Social Norms} consists of $704$ instances. In terms of construction source, the benchmark comprises $1,704$ original instances and $1,777$ rewritten or harder variants. Each retained instance is accompanied by structured metadata, including taxonomy labels, question type, narrative perspective, construction source, rewriting strategy, verification history, and annotation details. This rich metadata infrastructure enables performance disaggregation across primary and secondary dimensions, task paradigms, question formats, perspectives, and data sources, moving well beyond the limitations of a single aggregate evaluation score.

Table~\ref{tab:benchmark-comparison} positions \datasetname{} relative to representative social-cognition benchmarks reviewed in Sections~\ref{sec:zing} and~\ref{sec:actio}. The comparison centers on data form and coverage, i.e., target capability,  question format, context length, narrative perspective, and scenario source. Existing resources typically concentrate on narrower subsets,  such as single-choice questions, short vignettes, or third-person narratives alone. \datasetname{} deliberately broadens the design space: it spans four question types, i.e., single-choice, multi-select, judgment, and open-ended questions, both short and long contexts, and both first- and third-person perspectives. Furthermore, it provides a larger verified corpus than several prior benchmarks, with $284$ scenarios and $3,481$ expert-verified instances.

\begin{table}[t]
\centering
\scriptsize
\caption{Comparison between \datasetname{} and representative social-cognition benchmarks in task format, context length, narrative perspective, scenario coverage, and question scale. A $\checkmark$ indicates coverage, $\times$ indicates that the dimension is not covered. 
Short contexts are less than $500$ characters, and long contexts are more than $500$ characters.
}
\label{tab:benchmark-comparison}
\resizebox{\textwidth}{!}{%
\begin{tabular}{@{}lcccccccccc@{}}
\toprule
 & \multicolumn{4}{c}{Question formats} & \multicolumn{2}{c}{Context length} & \multicolumn{2}{c}{Perspective} & \multicolumn{2}{c}{Scale} \\
\cmidrule(lr){2-5}\cmidrule(lr){6-7}\cmidrule(lr){8-9}\cmidrule(l){10-11}
Benchmark & Single-choice & Multi-select & Judgment & Open-ended & Short & Long & First-person & Third-person & Scenarios & Questions \\
\midrule
ToMBench~\cite{tombench}
& $\checkmark$ & $\times$ & $\checkmark$ & $\times$ & $\checkmark$ & $\times$ & $\times$ & $\checkmark$
& 1,584 & 2,860 \\
EmoBench~\cite{sabour2024emobench}
& $\checkmark$ & $\times$ & $\times$ & $\times$ & $\checkmark$ & $\times$ & $\checkmark$ & $\checkmark$
& 321 & 400 \\
FANToM~\cite{fantom}
& $\checkmark$ & $\times$ & $\checkmark$ & $\checkmark$ & $\checkmark$ & $\checkmark$ & $\checkmark$ & $\checkmark$
& 256 & 10,317 \\
HI-TOM~\cite{hitom}
& $\checkmark$ & $\times$ & $\times$ & $\times$ & $\checkmark$ & $\times$ & $\times$ & $\checkmark$
& 240 & 1,200 \\
SOTOPIA~\cite{sotopia}
& $\times$ & $\times$ & $\times$ & $\checkmark$
& $\times$ & $\checkmark$
& $\checkmark$ & $\checkmark$
& 90 & 90 \\

SocialBench~\cite{chen2024socialbench}
& $\checkmark$ & $\checkmark$ & $\times$ & $\checkmark$
& $\checkmark$ & $\checkmark$
& $\checkmark$ & $\checkmark$
& 7,264 & 7,702 \\

SocialEval~\cite{zhou2025socialeval}
& $\checkmark$ & $\times$ & $\times$ & $\checkmark$
& $\times$ & $\checkmark$
& $\checkmark$ & $\checkmark$
& 153 & 2,493 \\

SI-Bench~\cite{huang2025sibench}
& $\times$ & $\times$ & $\times$ & $\checkmark$
& $\checkmark$ & $\times$
& $\checkmark$ & $\checkmark$
& 2,221 & 2,221 \\

SocialIQA~\cite{socialiqa}
& $\checkmark$ & $\times$ & $\times$ & $\times$
& $\checkmark$ & $\times$
& $\times$ & $\checkmark$
& 37,132 & 37,588 \\

DecodingTrust~\cite{wang2023decodingtrust}
& $\checkmark$ & $\times$ & $\checkmark$ & $\checkmark$
& $\checkmark$ & $\checkmark$
& $\checkmark$ & $\checkmark$
& 536,009 & 536,009 \\

Social Chemistry 101~\cite{socialchemistry}
& $\times$ & $\times$ & $\checkmark$ & $\checkmark$
& $\checkmark$ & $\times$
& $\checkmark$ & $\checkmark$
& 103,296 & 355,922 \\

MT-Bench-101~\cite{bai2024mt}
& $\times$ & $\times$ & $\times$ & $\checkmark$
& $\checkmark$ & $\checkmark$
& $\checkmark$ & $\times$
& 1,388 & 4,208 \\

ELEPHANT~\cite{cheng2025elephant}
& $\times$ & $\times$ & $\times$ & $\checkmark$
& $\checkmark$ & $\checkmark$
& $\checkmark$ & $\times$
& 10,395 & 11,986 \\

BenchPreS~\cite{yoon2026benchpres}
& $\times$ & $\times$ & $\times$ & $\checkmark$
& $\times$ & $\checkmark$
& $\checkmark$ & $\checkmark$
& 390 & 1,950 \\

RealPref~\cite{guo2026realpref}
& $\checkmark$ & $\times$ & $\checkmark$ & $\checkmark$
& $\times$ & $\checkmark$
& $\checkmark$ & $\checkmark$
& 100 & 1,300 \\

LongMemEval~\cite{wu2024longmemeval}
& $\times$ & $\times$ & $\times$ & $\checkmark$
& $\times$ & $\checkmark$
& $\checkmark$ & $\times$
& 500 & 500 \\

SafetyBench~\cite{zhang2024safetybench}
& $\checkmark$ & $\times$ & $\times$ & $\times$
& $\checkmark$ & $\times$
& $\checkmark$ & $\checkmark$
& 11,435 & 11,435 \\

GoEmotions~\cite{demszky2020goemotions}
& $\times$ & $\checkmark$ & $\times$ & $\times$
& $\checkmark$ & $\times$
& $\checkmark$ & $\checkmark$
& 58,009 & 58,009 \\

ETHICS~\cite{ethics}
& $\times$ & $\times$ & $\checkmark$ & $\times$
& $\checkmark$ & $\checkmark$
& $\checkmark$ & $\checkmark$
& 134,420 & 134,420 \\

NormBank~\cite{normbank}
& $\times$ & $\times$ & $\checkmark$ & $\checkmark$
& $\checkmark$ & $\times$
& $\times$ & $\checkmark$
& 70,215 & 155,423 \\

PUB~\cite{sravanthi2024pub}
& $\checkmark$ & $\times$ & $\checkmark$ & $\times$
& $\checkmark$ & $\checkmark$
& $\checkmark$ & $\checkmark$
& 18,159 & 26,743 \\
\midrule
\textbf{\datasetname{} (ours)}
& $\checkmark$ & $\checkmark$ & $\checkmark$ & $\checkmark$ & $\checkmark$ & $\checkmark$ & $\checkmark$ & $\checkmark$
& 284 & 3,481 \\
\bottomrule
\end{tabular}%
}
\end{table}

\begin{figure}[t]
    \centering
    \includegraphics[width=0.9\linewidth]{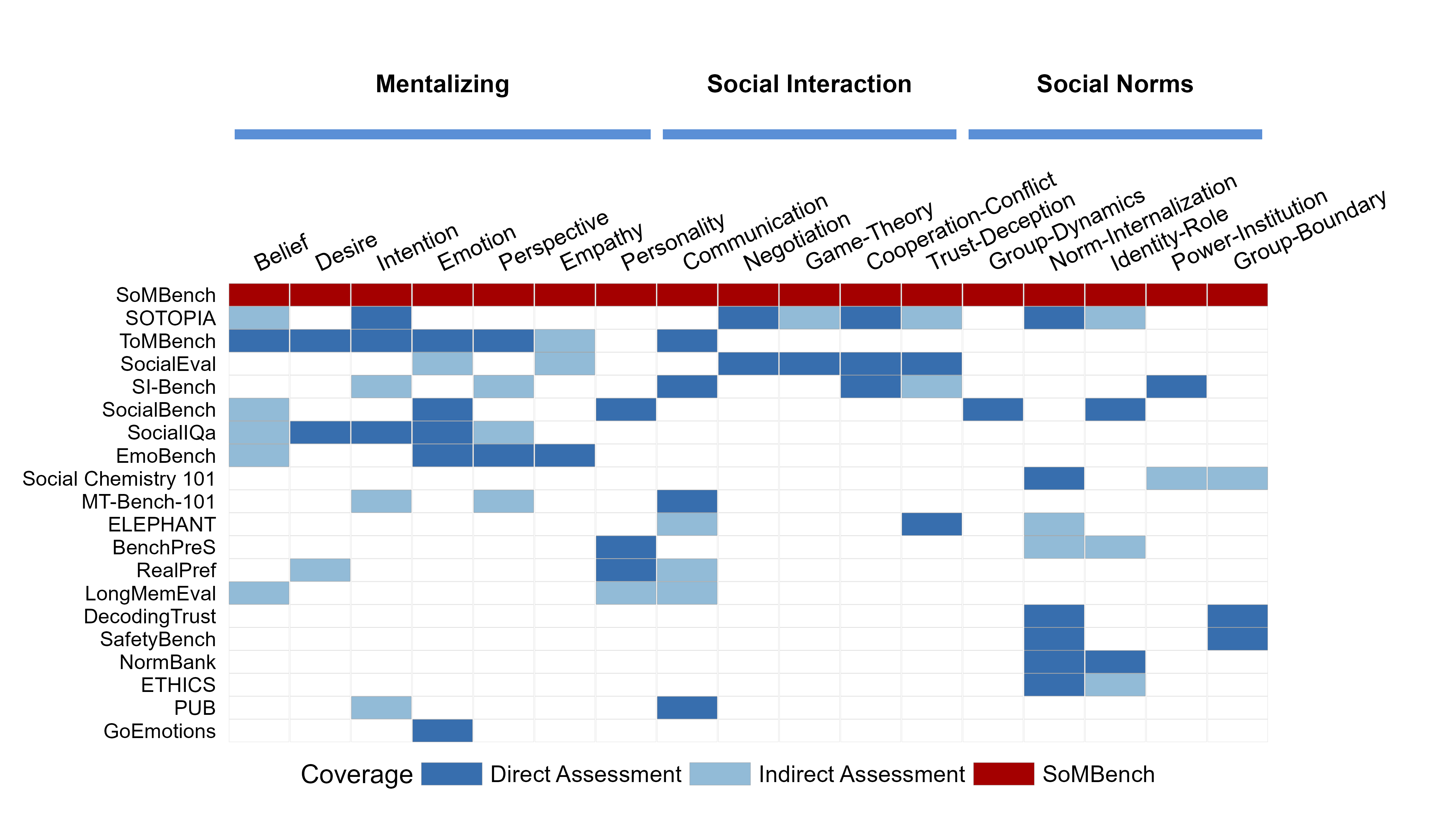}
    \caption{\textbf{Coverage comparison across social-cognition benchmarks.} Columns correspond to the second-level dimensions in the \datasetname{} taxonomy and are grouped into Mentalizing, Social Interaction, and Social Norms. Dark blue marks primary coverage, light blue marks secondary coverage, and blank cells indicate that the dimension is not a central target of the benchmark. \datasetname{} provides systematic primary coverage across the full taxonomy, while existing benchmarks cover more selective subsets.}
    \label{fig:benchmark_coverage}
\end{figure}

The scale comparison above clarifies what each dataset contains; the coverage and evaluation results further show why this design matters. Figure~\ref{fig:benchmark_coverage} compares the breadth of our benchmark coverage with others. Rows represent representative benchmarks, columns correspond to second-level social-cognition dimensions grouped into Mentalizing, Social Interaction, and Social Norms, and colored cells indicate whether a dimension is treated as a primary or secondary focus. The matrix shows that prior benchmarks usually concentrate on a narrower subset of mental-state reasoning, interactional, or normative  abilities, whereas \datasetname{} is designed to provide primary coverage across all three families and their fine-grained dimensions.

\subsection{Evaluation Protocol and Model Findings}
\label{sec:socialmind-findings}

\paragraph{Evaluation setup.}
We evaluated 20 representative LLMs on the verified evaluation split of \datasetname{}, spanning flagship systems, high-performance models, mid-sized architectures, and lightweight variants. The evaluated pool includes \texttt{claude-opus-4-8}, \texttt{gpt-5.5}, \texttt{gpt-5.4}, \texttt{gemini-pro-agent} (reported as \texttt{gemini-3.1-pro-high}), and models from the DeepSeek, Qwen, Doubao, and Claude families. Every model was evaluated on the same 3,481 scored instances. The evaluation harness retained item-level metadata, including primary and secondary dimensions, task-paradigm identifiers, question type, perspective, and context-length category, to support fine-grained diagnostic analysis.

Evaluation follows a two-track scoring protocol. For closed-form questions, correctness is computed by exact match against the reference label. For open-ended questions, two automated judges produce a continuous quality score, which is converted to a binary correctness judgment using the fixed pass threshold used in this release. The reported Q4 value is therefore the open-ended pass rate, rather than the mean judge score. The evaluated split contains 713 single-choice, 626 multiple-choice, 1,661 judgment, and 481 open-ended items. Retaining the item-level metadata permits comparisons by question type, first- versus third-person perspective, context length, primary and secondary dimensions, and individual task paradigms.

\begin{figure}[t]
    \centering
    \includegraphics[width=\linewidth]{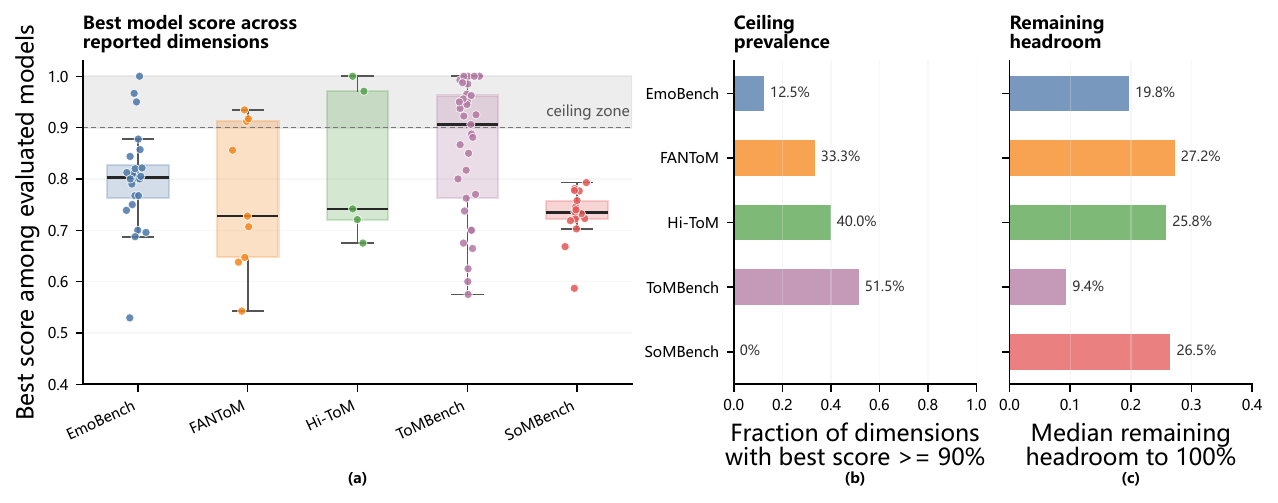}
    \caption{\textbf{Evaluation-depth comparison across social-cognition benchmarks.}
    (a) Each point is one reported metric or dimension, and its value is the best score among the evaluated models rather than the score of a particular model. The grey band marks the 90--100\% near-ceiling region. (b) Fraction of reported dimensions whose best-of-pool score reaches at least 90\%. (c) Median remaining headroom to 100\%, computed as $1-\mathrm{best\ score}$ for each reported dimension. \datasetname{} contains 17 reported secondary dimensions, none of which reaches the near-ceiling region.}
    \label{fig:benchmark-depth-comparison}
\end{figure}

\paragraph{Benchmark evaluation depth.}
Figure~\ref{fig:benchmark-depth-comparison} examines evaluation depth across the result tables in this study. Rather than reporting any single model's accuracy, the figure uses a best-of-pool metric: for each reported metric or dimension in a benchmark, we take the highest score achieved by any evaluated model in the local results. This ceiling score indicates whether at least one current model has already saturated for that dimension. We designate dimensions with a top score of at least 90\% as near-ceiling regions. Existing benchmarks already exhibit several such saturated regions. ToMBench has a median best score of 90.6\% across its reported dimensions, with 51.5\% of dimensions reaching the 90\% band; Hi-ToM and FANToM reach the same band on 40.0\% and 33.3\% of their dimensions, respectively. \datasetname{} presents a different profile: none of its 17 secondary dimensions reaches 90\%, its highest best-of-pool score is 80.5\%, and its median best score is 72.4\%. The resulting median headroom of 27.6\% indicates that the benchmark retains diagnostic separation for current models.

\begin{figure}[t]
    \centering
    \includegraphics[width=\linewidth]{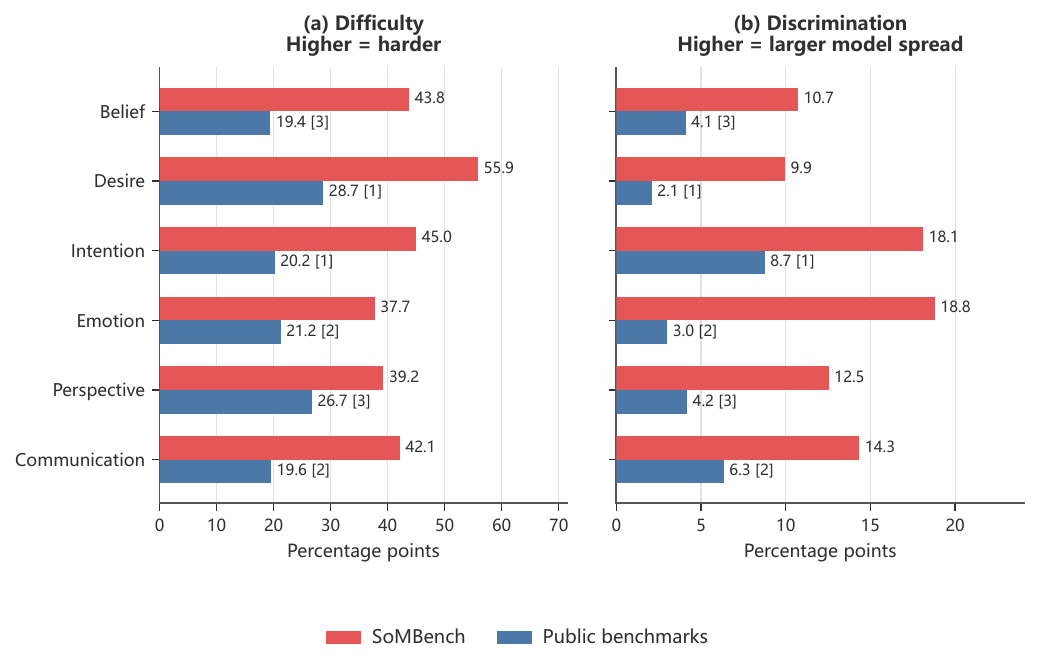}
    \caption{\textbf{Difficulty and discrimination on shared dimensions.}
    (a) Difficulty, defined as $100-\mathrm{accuracy}$, for six dimensions shared by \datasetname{} and public social-cognition benchmarks. (b) Discrimination, defined as the best-minus-worst score across the model pool. \datasetname{} values are equal-weight means over the four question types; public values are equal-weight means over the contributing benchmark-level metrics. The model pool contains GPT-5.5, DeepSeek-V4-Pro, and DeepSeek-V4-Flash ($n=3$). Bracketed values [k] indicate the number of public benchmarks contributing to each public aggregate.}
    \label{fig:shared-dimension-comparison}
\end{figure}

As shown in Figure~\ref{fig:shared-dimension-comparison}, \datasetname{} is consistently harder and more discriminative than the public benchmark aggregate across the six aligned dimensions. Difficulty ranges from 37.7 to 55.9 percentage points for \datasetname{}, compared with 19.4 to 28.7 percentage points for the public benchmarks. The largest \datasetname{} difficulty occurs on Desire, whereas Emotion is the least difficult aligned dimension. The model spread for \datasetname{} ranges from 9.9 to 18.8 percentage points, exceeding the 2.1 to 8.7 percentage-point range of the public aggregate, with the widest separation on Emotion and Intention. Because this comparison uses only three matched models and six shared dimensions, it provides evidence of aligned diagnostic depth rather than a claim about every task in every benchmark.

\begin{figure}[t]
  \centering
  \includegraphics[width=0.95\linewidth]{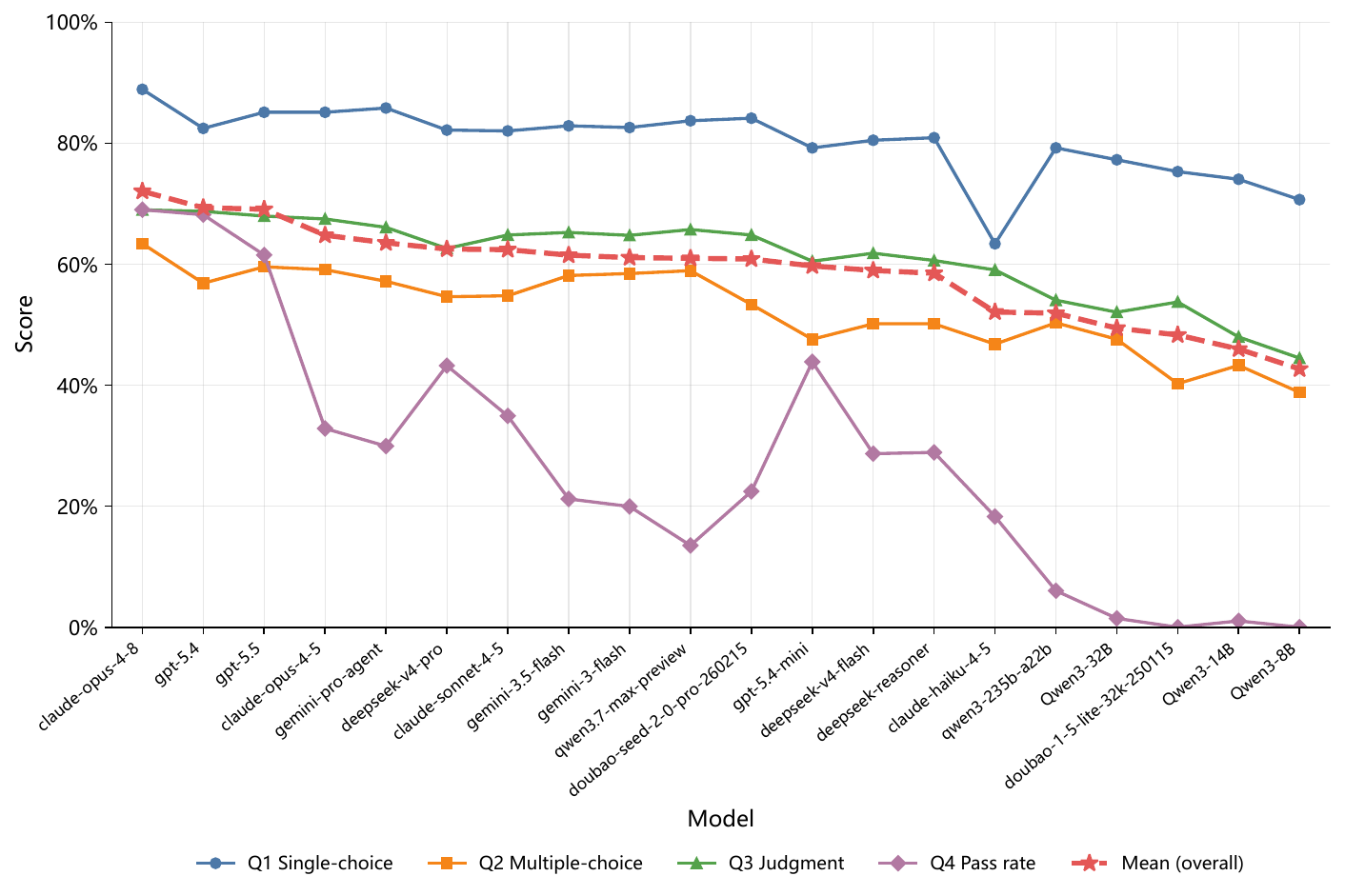}
  \caption{Overall accuracy of the 20 evaluated LLMs on \datasetname{}. Models are ordered by aggregate accuracy from high to low.}
  \label{fig:socialmind-overall-accuracy}
\end{figure}

\paragraph{Overall model ranking.}
Figure~\ref{fig:socialmind-overall-accuracy} reports the overall accuracy of the 20 evaluated models. \texttt{claude-opus-4-8} leads at 72.08\%, followed by \texttt{gpt-5.4} (69.35\%), \texttt{gpt-5.5} (69.09\%), \texttt{claude-opus-4-5} (64.81\%), and \texttt{gemini-3.1-pro-high} (63.54\%). At the lower end, \texttt{qwen3-8b} achieves 42.69\%, followed by \texttt{qwen3-14b} (45.99\%), \texttt{doubao-1-5-lite-32k-250115} (48.32\%), \texttt{qwen3-32b} (49.44\%), and \texttt{qwen3-235b-a22b} (51.91\%). The 29.39-point gap between the strongest and weakest systems confirms that \datasetname{} separates model capability without approaching aggregate saturation. 

The aggregate ranking suggests two conclusions. First, frontier and high-end commercial models occupy most of the top positions, indicating that general instruction following, long-context handling, and broad reasoning ability remain beneficial for social-cognitive tasks. Second, no model approaches saturation. Even the best performer misses roughly a quarter of the benchmark, which means the benchmark is not merely measuring whether models can recognize surface social commonsense. It continues to expose failures in belief tracking, norm conflict, conversational presupposition, and interaction dynamics.

\begin{figure}[t]
  \centering
  \includegraphics[width=0.95\linewidth]{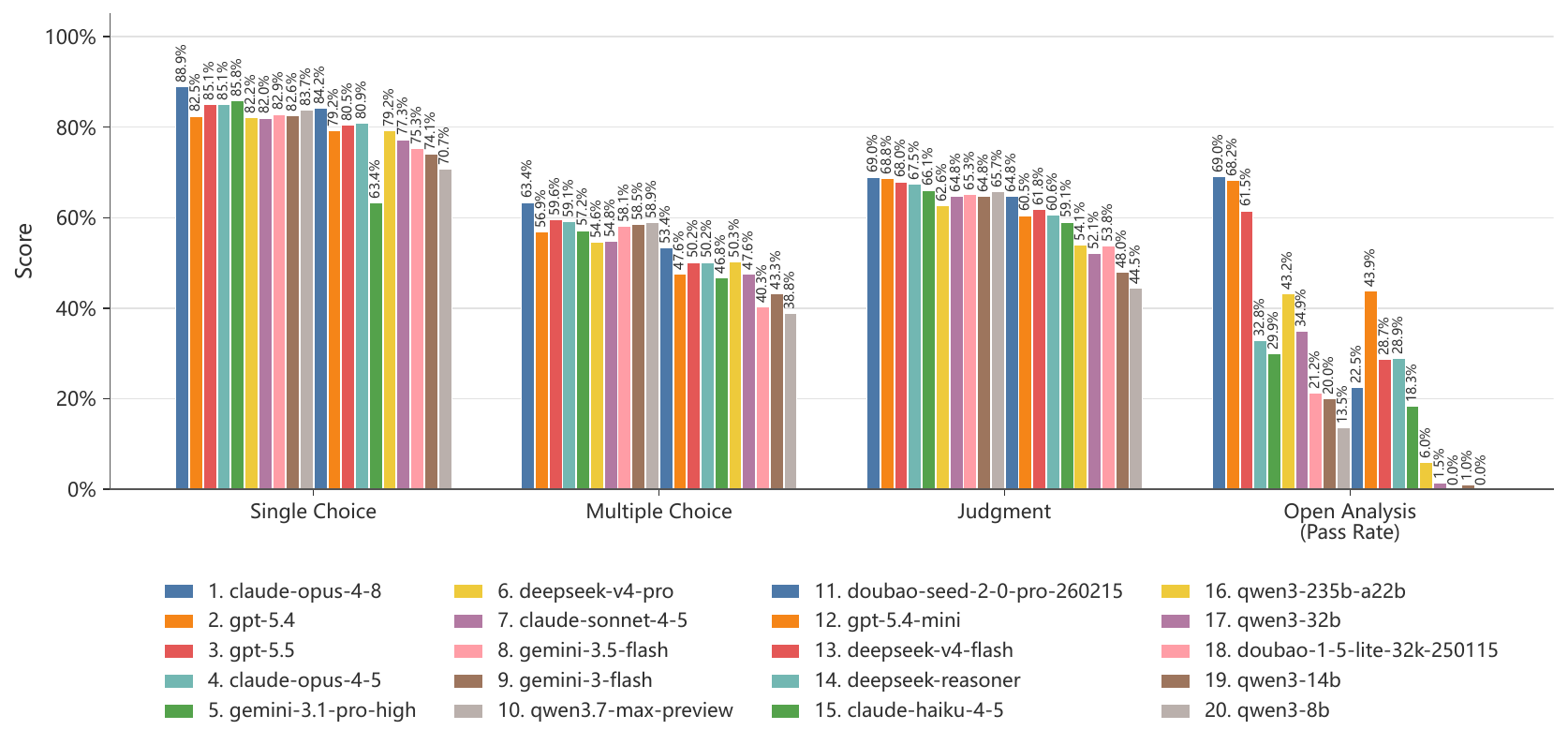}
  \caption{Accuracy by question type. Single-choice items are near-saturated for many models, whereas multiple-choice, judgment, and open-ended analysis expose stronger differences.}
  \label{fig:socialmind-qtype-accuracy}
\end{figure}

\paragraph{Question-format effects.}
Figure~\ref{fig:socialmind-qtype-accuracy} decomposes performance by question type. Single-choice questions are the easiest format, with a mean accuracy of 94.47\% across models. Multiple-choice questions are substantially harder, with a mean of 55.60\%, as they require models to identify multiple jointly correct alternatives rather than settling on a single plausible option. Judgment questions fall between these extremes, yielding  a mean of 64.06\%, while open-ended analysis questions average 62.86\% but show the widest model spread, from 34.47\% to 82.20\%.

The format-level pattern is diagnostically important. High performance on single-choice questions suggests that many models can identify a plausible social answer when the evidence is cleanly localized and the output space is constrained. The collapse on multiple-choice items suggests a different limitation: models often recognize one salient social cue but fail to maintain the full set of compatible mental states, strategic moves, or norm constraints. Open-ended analysis questions, meanwhile, produce the broadest variance, indicating that free-form social reasoning depends heavily on response planning and explanation quality, not merely on answer recognition. This finding is especially relevant for downstream agent training, where a model must not only select an action but justify it under perspective and norm constraints.

\begin{figure}[!tbp]
  \centering
  \includegraphics[width=0.95\linewidth]{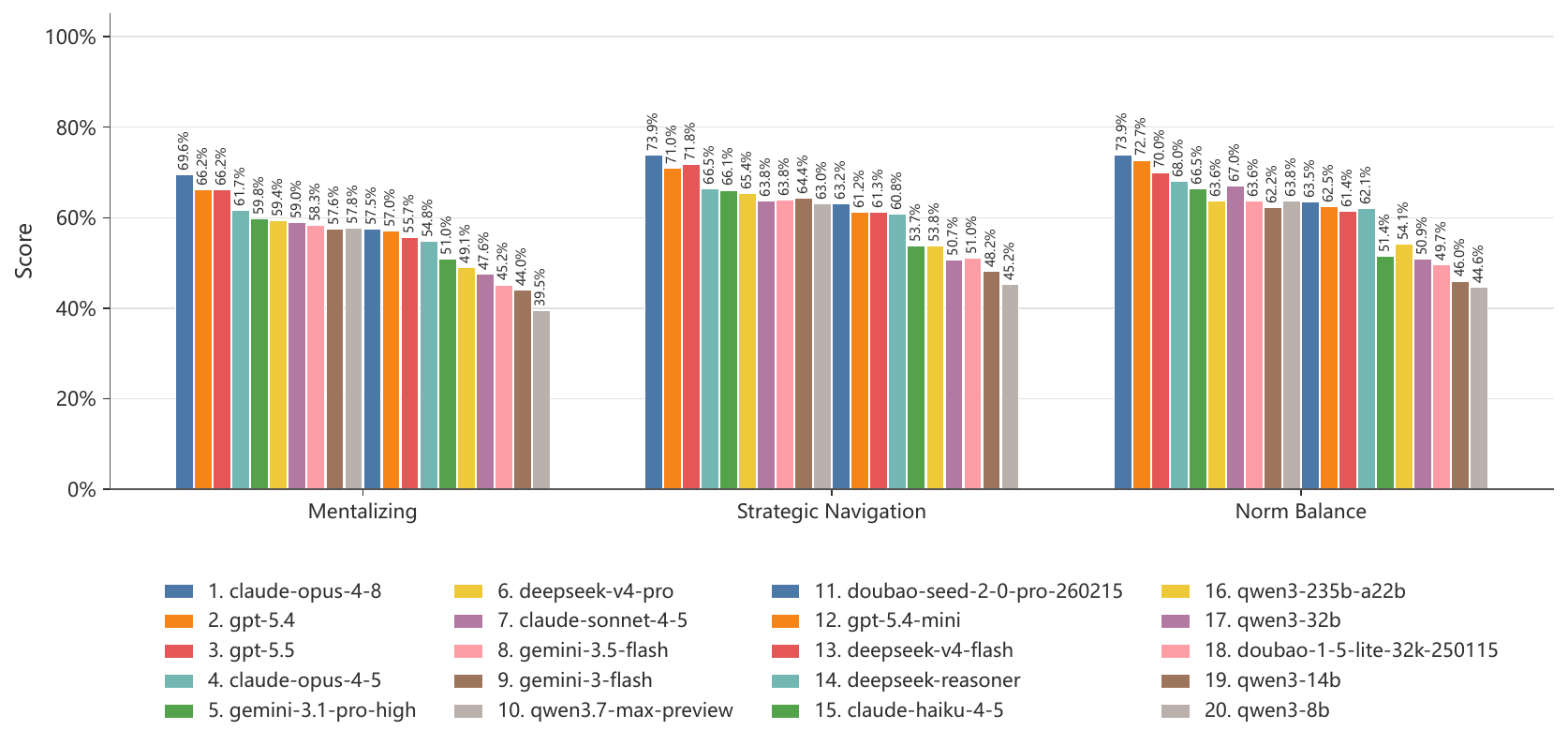}
  \caption{Accuracy across the three primary dimensions of \datasetname{}: Mentalizing, Strategic Navigation in Social Interaction, and Norm Internalization and Dynamic Balance.}
  \label{fig:socialmind-dim1-accuracy}
\end{figure}

\paragraph{Primary capability dimensions.}
Performance across the three primary dimensions is presented in Figure~\ref{fig:socialmind-dim1-accuracy}. Averaged across models, Strategic Navigation in Social Interaction is the highest-scoring dimension at 69.29\%, followed by Norm Internalization and Dynamic Balance at 67.65\%, and Mentalizing at 65.86\%. The relative ordering should not be read as indicating that strategic interaction is intrinsically easy. Instead, it reflects that models often benefit from familiar pragmatic and cooperative scripts, whereas mental-state attribution requires more precise separation between belief, desire, intention, perspective, and ground truth.

The primary-dimension results show that aggregate social performance obscures qualitatively distinct types of model weakness. Mentalizing is particularly sensitive to belief-truth separation and recursive perspective tracking. Strategic interaction, in turn,  rewards models capable of inferring conversational intent, negotiating cooperation and conflict, and reasoning about trust or deception. Norm reasoning requires models to avoid applying generic moral rules without regard to role, group boundary, or institutional context. The fact that no dimension is near saturation, even for the top- performing models, vindicates our design choice: the social mind is best treated as a multi-component construct, rather than reduced to a unitary theory-of-mind score.

\begin{figure}[t]
  \centering
  \includegraphics[width=0.95\linewidth]{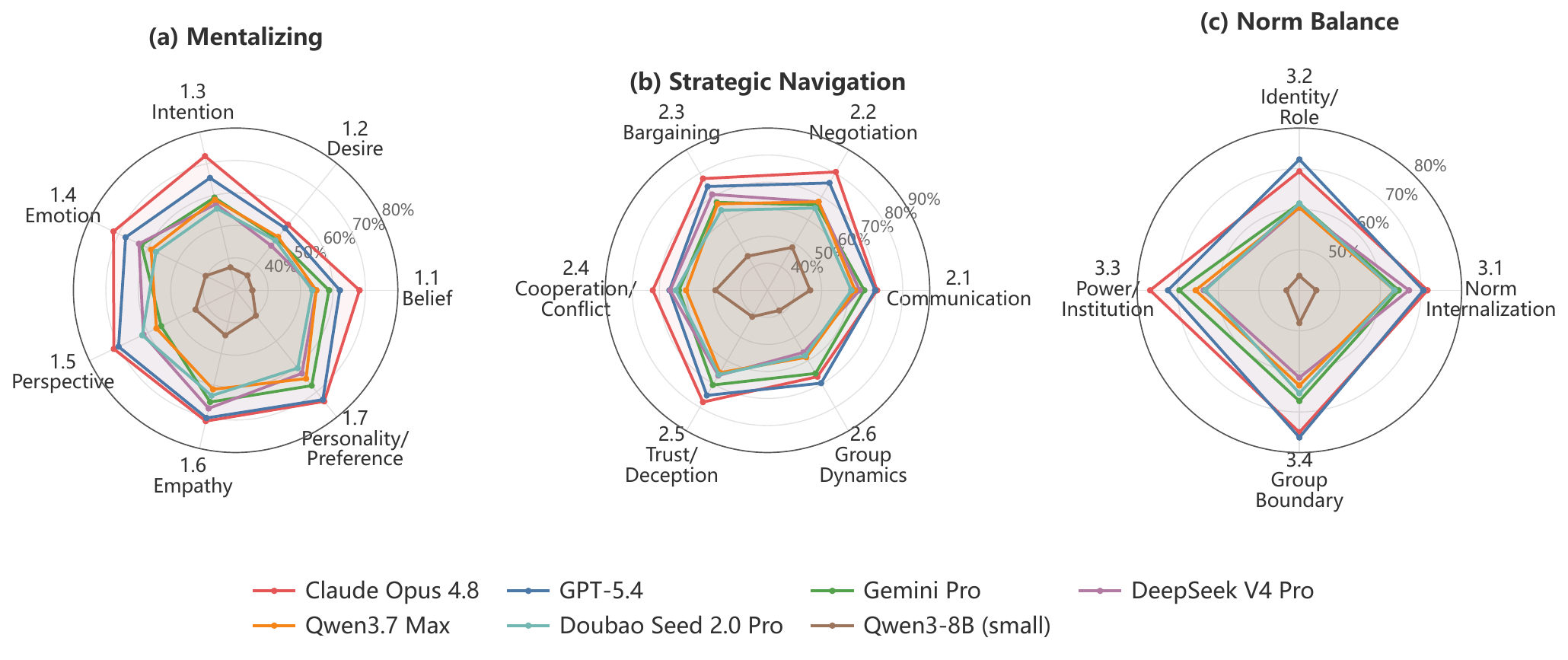}
  \caption{Secondary-dimension radar plots. The radial axis is truncated to emphasize model differences within each primary capability family.}
  \label{fig:socialmind-dim2-radar}
\end{figure}

\paragraph{Secondary-dimension structure.}
Figure~\ref{fig:socialmind-dim2-radar} shows secondary-dimension performance using radar plots. The radial scale is intentionally truncated rather than starting from zero, so that differences among models are visible within each primary dimension. Across the Mentalizing panel, models are relatively close on some emotion and perspective tasks but separate more clearly on belief and intention-related dimensions. In the Strategic Navigation panel, communication, negotiation, cooperation/conflict, and trust/deception produce different profiles, indicating that strong aggregate models do not share identical interactional strengths. The Norm Internalization panel reveals that identity/role, norm internalization, power/institution, and group boundary are not interchangeable: a model that succeeds on generic norm recognition may nonetheless struggle when the norm is conditioned on role or group membership.

\begin{figure}[!tbp]
  \centering
  \includegraphics[width=0.95\linewidth]{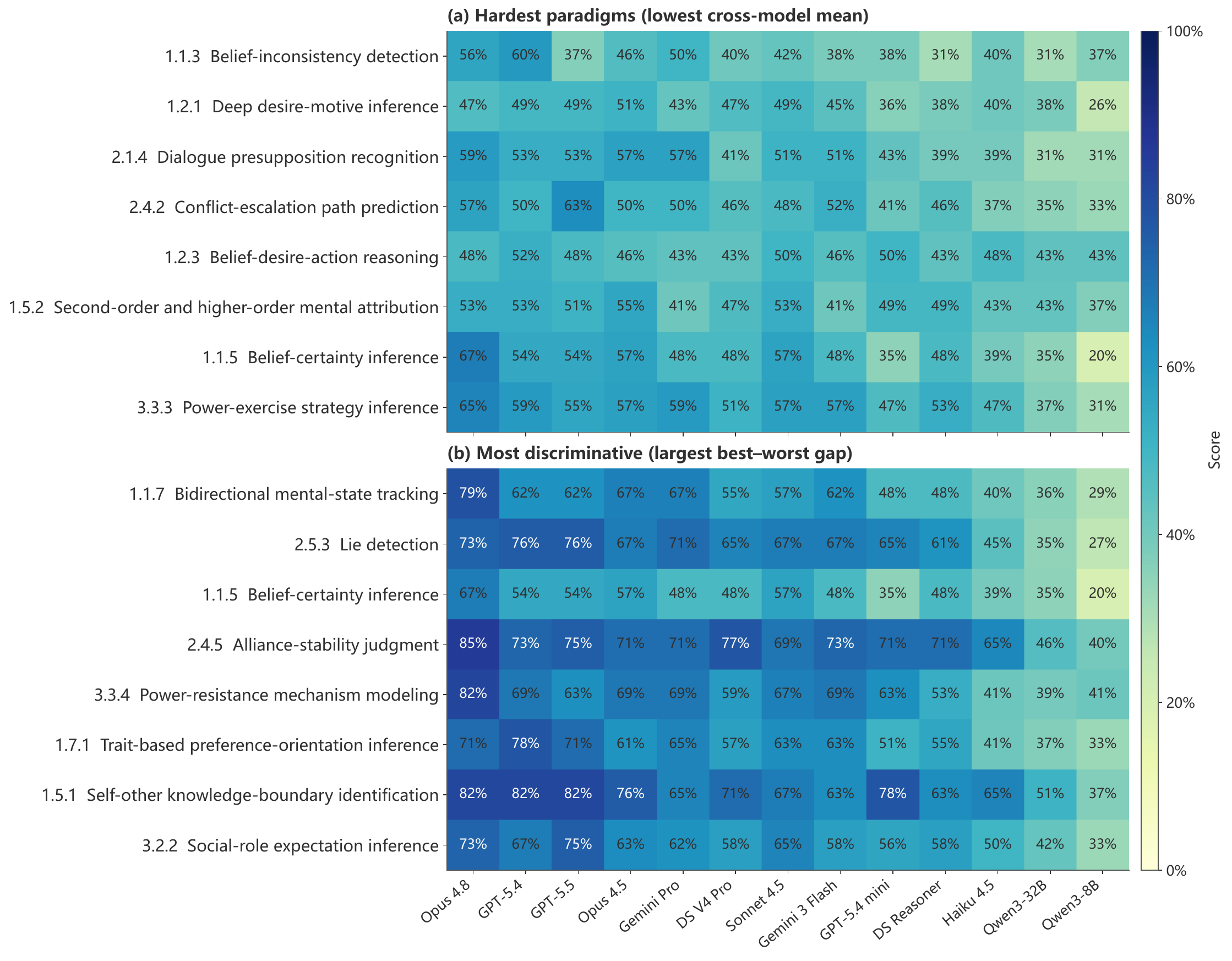}
  \caption{\textbf{Selected fine-grained task-paradigm scores.} are selected over all 20 evaluated models: (a) the eight paradigms with the lowest cross-model means and (b) the eight with the largest best-minus-worst gaps. Columns show 13 representative models ordered by overall accuracy.}
  \label{fig:socialmind-dim3-heatmap}
\end{figure}

\paragraph{Fine-grained task paradigms.}
Figure~\ref{fig:socialmind-dim3-heatmap} isolates two complementary diagnostic views of the fine-grained task paradigms. Panel~(a) selects the eight paradigms with the lowest cross-model means across all 20 evaluated models, whereas panel~(b) selects the eight with the largest best-minus-worst gaps. The lowest means occur for belief-inconsistency detection at 41.2\%, deep desire-motive inference at 42.4\%, dialogue presupposition recognition at 43.6\%, and conflict-escalation path prediction at 44.2\%. The largest model separations occur for bidirectional mental-state tracking at 50.0 points, lie detection at 49.0 points, belief-certainty inference at 47.8 points, and alliance-stability judgment at 45.8 points. Belief-certainty inference appears in both panels, showing that average difficulty and model discrimination can coincide but are not interchangeable. Together, the two views distinguish shared weaknesses from model-specific separation and show why overall accuracy alone is insufficient for mechanism-level diagnosis.

\begin{table}[!htbp]
  \centering
  \caption{\textbf{Difficult and discriminative fine-grained dimensions.} Rows are selected by low cross-model mean or large best-minus-worst spread. Accuracies are percentages over the 20 evaluated models; the model column lists four representative systems from higher to lower accuracy within each dimension.}
  \label{tab:socialmind-difficult-dimensions}
  \scriptsize
  \renewcommand{\arraystretch}{1.15}
  \begin{tabularx}{\textwidth}{@{}p{0.8cm}XrXr@{}}
    \toprule
    Dim & Task paradigm & Mean & Representative accuracy gradient & Gap \\
    \midrule
    1.1.3 & Belief-inconsistency detection & 41.2 & \texttt{gpt-5.4} (59.6) $>$ \texttt{claude-opus-4-8} (55.8) $>$ \texttt{gemini-3.1-pro-high} (50.0) $>$ \texttt{claude-opus-4-5} (46.2) & 28.8 \\
    1.1.5 & Belief-certainty inference & 46.2 & \texttt{claude-opus-4-8} (67.4) $>$ \texttt{claude-opus-4-5} (56.5) $>$ \texttt{claude-sonnet-4-5} (56.5) $>$ \texttt{doubao-seed-2-0-pro-260215} (56.5) & 47.8 \\
    1.1.7 & Bidirectional mental-state tracking & 53.0 & \texttt{claude-opus-4-8} (78.6) $>$ \texttt{claude-opus-4-5} (66.7) $>$ \texttt{gemini-3.1-pro-high} (66.7) $>$ \texttt{gemini-3.5-flash} (64.3) & 50.0 \\
    1.2.3 & Belief--desire--action reasoning & 45.0 & \texttt{gpt-5.4} (52.2) $>$ \texttt{claude-sonnet-4-5} (50.0) $>$ \texttt{gpt-5.4-mini} (50.0) $>$ \texttt{claude-opus-4-8} (47.8) & 15.2 \\
    2.1.4 & Dialogue presupposition recognition & 43.6 & \texttt{claude-opus-4-8} (58.8) $>$ \texttt{claude-opus-4-5} (56.9) $>$ \texttt{gemini-3.1-pro-high} (56.9) $>$ \texttt{gpt-5.4} (52.9) & 31.4 \\
    2.4.2 & Conflict escalation path prediction & 44.2 & \texttt{gpt-5.5} (63.0) $>$ \texttt{claude-opus-4-8} (56.5) $>$ \texttt{gemini-3.5-flash} (52.2) $>$ \texttt{gemini-3-flash} (52.2) & 37.0 \\
    3.1.1 & Social norm recognition & 52.3 & \texttt{claude-opus-4-8} (69.6) $>$ \texttt{gpt-5.4} (69.6) $>$ \texttt{gemini-3.5-flash} (63.0) $>$ \texttt{gpt-5.5} (60.9) & 39.1 \\
    3.1.2 & Norm-violation detection & 54.9 & \texttt{claude-opus-4-8} (72.0) $>$ \texttt{gemini-3.1-pro-high} (68.0) $>$ \texttt{claude-opus-4-5} (66.0) $>$ \texttt{claude-sonnet-4-5} (64.0) & 40.0 \\
    \bottomrule
  \end{tabularx}
\end{table}


Table~\ref{tab:socialmind-difficult-dimensions} highlights dimensions that are simultaneously difficult on average and discriminative across models, sorted by the taxonomy dimension rather than leaderboard ranking. The Mentalizing rows show that belief-centered subskills are not uniformly solved: belief-inconsistency detection, certainty inference, bidirectional mental-state tracking, and belief-desire-action reasoning each exhibit a visible gradient from strong to weak systems. Notably, the top model within a row is not always the highest aggregate model, suggesting that these mechanisms are partially specialized rather than reducible to overall benchmark strength. The Strategic Navigation rows show even sharper heterogeneity: dialogue presupposition recognition displays the largest gap, i.e., 42.1 points, indicating that some models recover implicit conversational assumptions while others remain anchored to literal utterances. Conflict escalation path prediction is the hardest selected paradigm, with a mean of 49.2\%, and separates models that can update relational dynamics from those that miss how local actions change future conflict trajectories. Finally, the Norm Internalization rows further show that norm recognition and norm-violation detection remain challenging even when the task is not framed as an abstract moral judgment; models vary substantially in whether they condition norms on role, context, and social domain.

\begin{figure}[!h]
  \centering
  \includegraphics[width=0.95\linewidth]{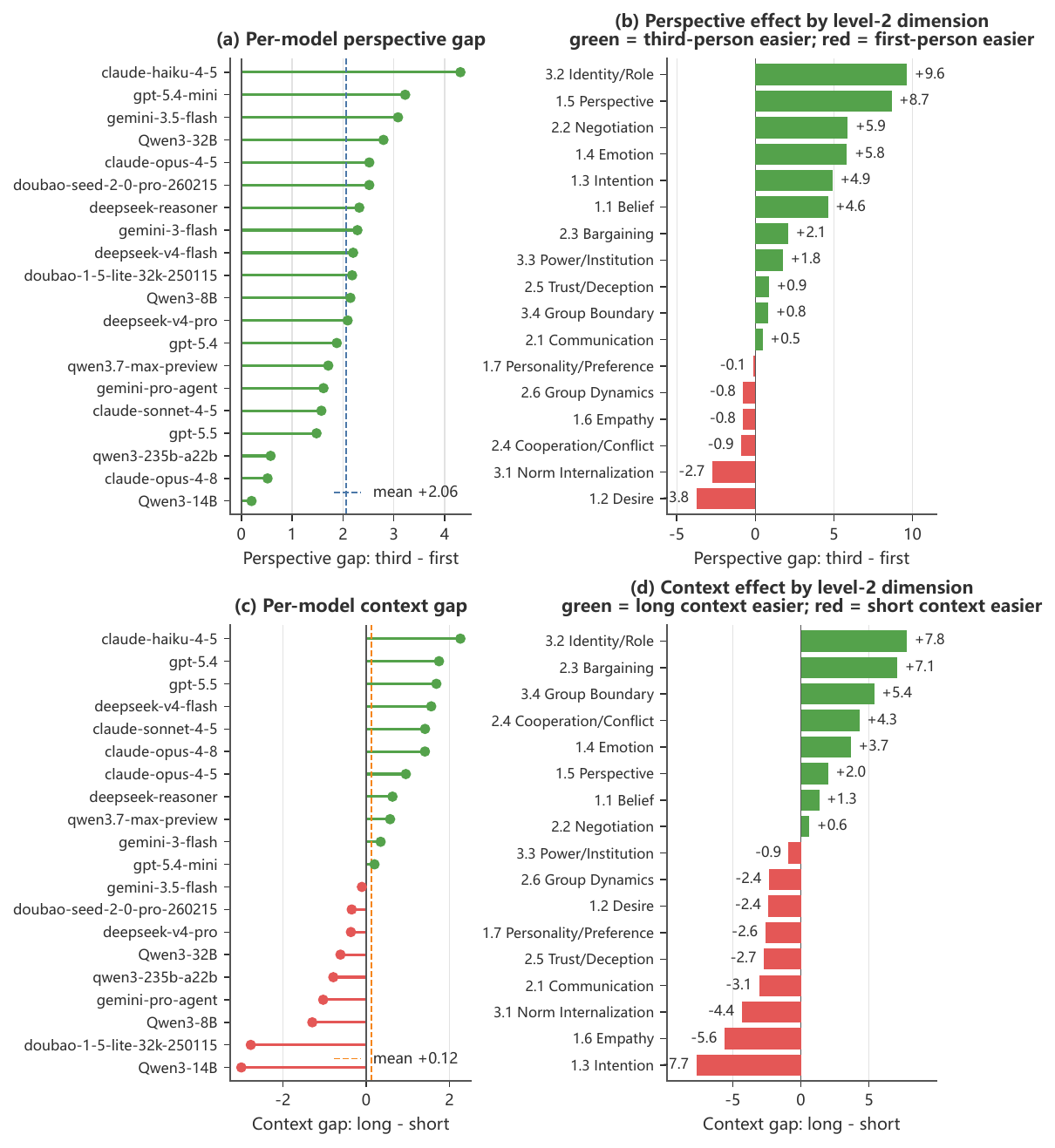}
  \caption{\textbf{Perspective and context-length effects on \datasetname{}.} (a) Per-model perspective gaps, computed as third-person minus first-person accuracy. (b) Perspective gaps averaged by secondary dimension. (c) Per-model context-length gaps, computed as long minus short accuracy. (d) Context-length gaps averaged by secondary dimension. Green indicates that the first condition in each difference is easier; red indicates the reverse. All gap values are reported in percentage points.}
  \label{fig:socialmind-perspective_context_combo}
\end{figure}

\paragraph{Perspective and context effects.}
Figure~\ref{fig:socialmind-perspective_context_combo} summarizes the effects of narrative perspective and context length. Across the 20 evaluated models, third-person items prove slightly easier than first-person items, yielding average accuracies of 59.82\% and 57.76\%, respectively. This 2.06-point mean difference is small but consistent across models, whose third-person advantage ranges from 0.2 to 4.3 points. The dimension-level results show that this effect is not uniform. Third-person framing provides the largest advantages for Identity/Role and Perspective, at 9.7 and 8.7 points, whereas first-person framing is easier for Desire and Norm Internalization by 3.8 and 2.7 points. Context length has an even smaller aggregate effect: long contexts average 58.86\%, compared with 58.74\% for short contexts, a difference of only 0.12 points. This near-zero aggregate masks opposing dimension-level patterns. Long contexts improve performance most for Identity/Role and Bargaining, by 7.8 and 7.1 points, whereas short contexts are easier for Intention and Empathy, by 7.7 and 5.6 points. These results show that perspective and context length do not impose a uniform difficulty shift; instead, their effects depend on the social-cognitive capability being tested.

\paragraph{Implications for training and agent evaluation.}
The findings reveal two distinct classes of capability gaps, i.e., stable gaps and context-sensitive gaps. Stable gaps are those that appear across many models and task formats: belief-truth separation, bidirectional mental-state tracking, conversational presupposition, conflict escalation, and norm-domain discrimination. These are good candidates for targeted training data, process supervision, and diagnostic reward modeling because the failure modes are tied to explicit metadata and can be converted into contrastive examples. Context-sensitive gaps are those that vary more strongly by model family, output format, or perspective, such as open-ended explanation quality and first-person viewpoint control. Addressing these may require not only additional supervised examples but also prompting strategies, tool-assisted memory, deliberation scaffolds, or explicit agent state tracking. Overall, \datasetname{} thus serves not merely as a social intelligence leaderboard, but as a diagnostic map of the social-cognitive mechanisms that current LLMs still fail to represent reliably.

\section{\modelname{}: Internalizing Social Intelligence}
\label{sec:zing}

\subsection{Motivation and Challenges}
\label{sec:zing-motivation}

The goal of \modelname{} is to internalize social mind reasoning as a stable parametric capability. The taxonomy in Section~\ref{sec:socialmind-taxonomy} defines this target as a structured and interdependent capability space, rather than a single social-reasoning skill. This view changes the training problem: the objective is not merely to obtain better task-level behavior, but to make the model acquire a coordinated set of social-cognitive abilities that can support reasoning across situated contexts. These properties make it nontrivial to decide where training should focus and how the capability space should be internalized.

The first challenge is adaptive capability diagnosis. In a structured capability space, aggregate performance gives only limited guidance about the model's capability profile and the supervision needed next. The capability gaps that matter for the next round of training are not fixed: as some parts of the space become better internalized, the useful supervision shifts toward the remaining weak or fragile regions. Therefore, the challenge is to interpret observed performance in terms of capability strengths and gaps, so that subsequent supervision can adapt to the model's current state.

The second challenge is hierarchical capability internalization. The capabilities in the taxonomy are interdependent rather than flat: complex social judgments presuppose more basic social-cognitive grounding. When this grounding is unstable, high-level examples may not provide effective signals for the intended higher-level ability. The model may instead rely on local task patterns or generic social priors. Therefore, the challenge is to make the internalization of complex social mind abilities build on the lower-level social-cognitive capacities they depend on, rather than bypass them.

These two challenges suggest two design principles for \modelname{}. First, supervision should be diagnosis-driven: training data should be selected, synthesized, and filtered according to the model's current capability gaps rather than treated as a fixed resource. Second, training should be stage-wise: broad social-cognitive foundations should be strengthened before later rounds focus on more fine-grained social mind abilities. The next section introduces the overall pipeline that instantiates these principles through data iteration and staged post-training.

\subsection{Training Pipeline Overview}
\label{sec:zing-pipeline}

\begin{figure}[t]
  \centering
  \includegraphics[width=0.95\linewidth]{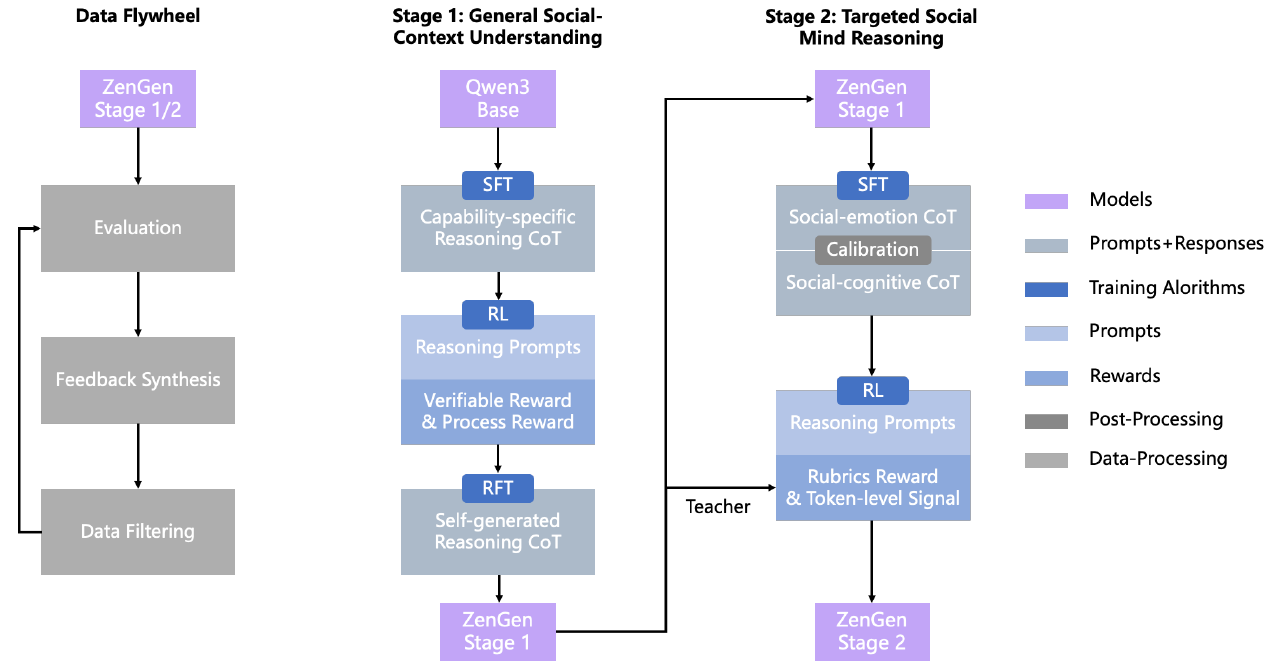}
  \caption{Training Pipeline Overview.}
  \label{fig:train_pipeline}
\end{figure}

\modelname{} implements the two principles above through a diagnosis-driven data flywheel and a stage-wise post-training path, instantiated on Qwen-family backbones. The data flywheel adapts supervision to the model's current capability gaps, while the stage-wise path organizes capability internalization from broad ToM foundations to more specialized social mind reasoning. Together, they turn the taxonomy-defined capability space into an iterative training recipe rather than a fixed corpus or a one-shot fine-tuning run. Figure~\ref{fig:train_pipeline} summarizes this architecture.

The data flywheel determines where each round of supervision should be concentrated. Starting from the current checkpoint, the pipeline evaluates model behavior on capability-labeled tasks and maps failures back to the corresponding social-cognitive dimensions. It then synthesizes or selects new samples that reconstruct the shared deficiency behind those failures, rather than merely paraphrasing individual bad cases. Candidate data are further filtered or repaired for correctness, near-duplicate control, answerability, difficulty, shortcut dependence, and distributional balance before entering the next training round. In this way, the latest checkpoint becomes the reference model for the next iteration, and the supervision pool progressively tracks the model's changing capability profile.

The stage-wise path determines how these capabilities are internalized. Stage~1 builds a broad foundation for general ToM and social-context understanding, using capability-balanced data construction, correctness-constrained reasoning traces, and verifiable and process-reward optimization to strengthen basic operations for tracking information access, updating belief states, and grounding intentions or emotions in context. Stage~2 starts from the Stage~1 checkpoint and targets residual fine-grained social mind gaps that require more compositional reasoning over affect, perspective, norms, and interaction. In this stage, capability-oriented SFT supplies targeted reasoning traces, while subsequent OPD~\cite{lu2025onpolicydistillation} and mixed-reward GRPO~\cite{grpo} refine answer stability and reasoning-process quality.

The remainder of this section first describes the data flywheel, then presents the two training stages, and finally reports the model-family evaluation.

\subsection{Mind-Capability-Driven Data Iteration Framework}
\label{sec:zing-data-iteration}

This section describes the data flywheel that supplies targeted supervision throughout \modelname{} training. The flywheel implements adaptive capability diagnosis by converting model behavior into capability-targeted data for the next training round. Rather than treating failed items as examples to be directly rewritten, it interprets failures through the social-mind capability taxonomy, abstracts shared weakness patterns, synthesizes new candidates that exercise those weaknesses, and filters the candidates according to their training value.

Existing ToM and social-cognition data provide useful references, but they do not by themselves solve the adaptive supervision problem. Human-annotated data are reliable but limited in scale, while automated synthesis is scalable but often fixed by predefined templates, topics, or capability prompts~\cite{emelin-etal-2021-moral,nematzadeh-etal-2018-evaluating,exploretom2025}. The key limitation is not only the initial coverage of the data, but the absence of a closed data-construction loop that updates the supervision signal as the model's capability profile changes. This motivates a flywheel-style process in which evaluation, diagnosis, synthesis, and filtering are coupled across iterations, with each updated checkpoint providing the behavioral signal for the next round.

\begin{figure}[t]
  \centering
  \includegraphics[width=\linewidth]{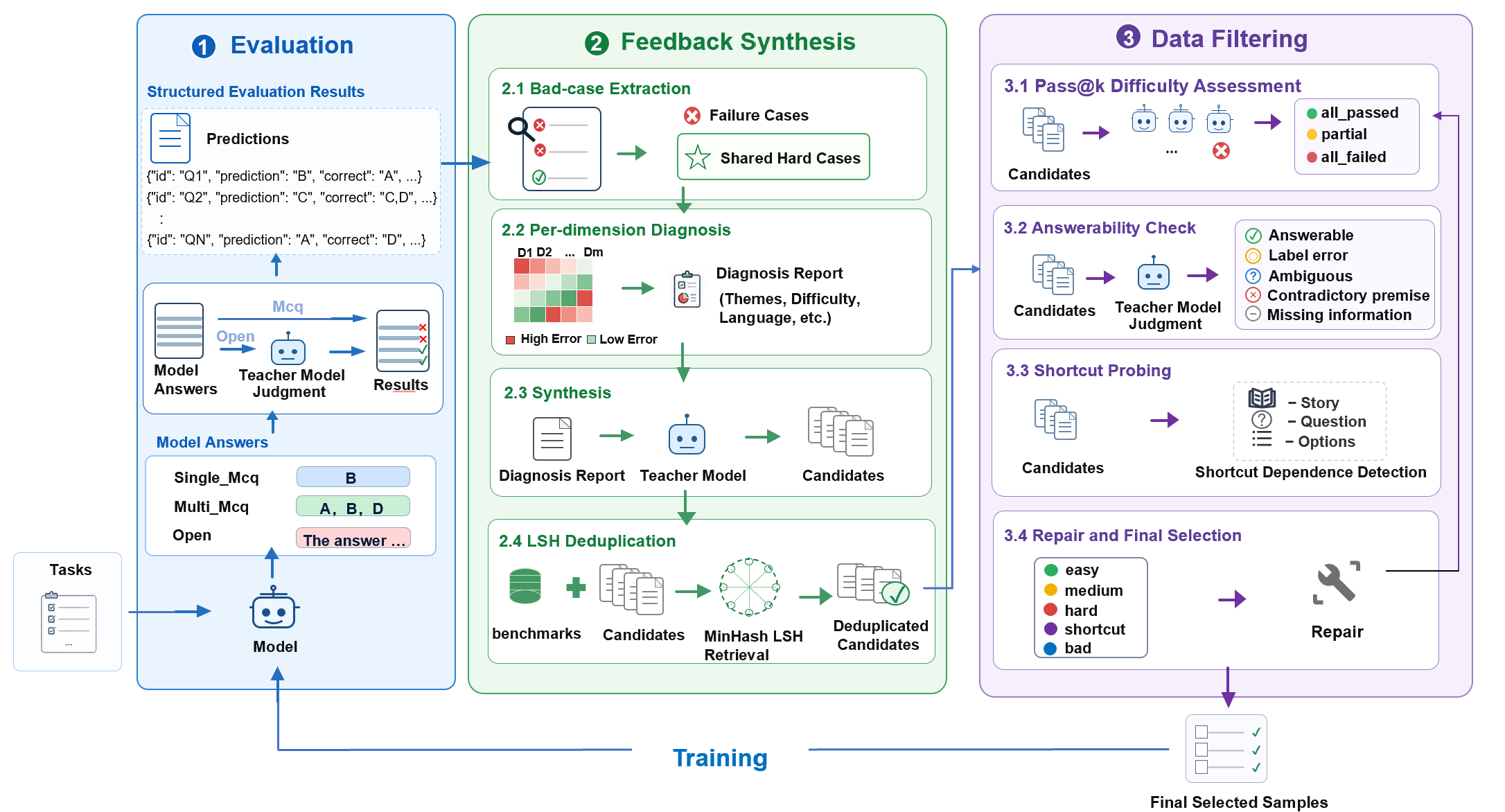}
  \caption{Capability diagnosis-driven data iteration framework.}
  \label{fig:data-iteration-strategy}
\end{figure}

We implement this loop as FLARE (Failure-Loop Augmented Refinement Engine), an integrated data engine built on SoMEval. FLARE uses a unified schema for capability-labeled evaluation, failure diagnosis, synthesis, and filtering. As shown in Figure~\ref{fig:data-iteration-strategy}, the framework proceeds through three stages: Evaluation, which produces structured model-behavior records; Feedback Synthesis, which turns shared failures into capability-oriented candidate data; and Data Filtering, which validates whether the candidates provide effective and reliable training supervision.

FLARE's iteration starts from a capability-labeled diagnostic pool constructed from our own data resources: taxonomy-guided seed cases and benchmark-inspired synthetic cases. The former directly instantiate the SocialMind capability taxonomy, while the latter follow the task formats and reasoning demands of existing ToM and social-cognition benchmarks without reusing their items. Each case carries capability metadata, allowing model behavior to be aggregated by dimension. As training proceeds, this pool is expanded and refined through the same diagnosis--synthesis--filtering loop. The following paragraphs describe the three FLARE stages in order.

\subsubsection{Evaluation}
The first stage converts model behavior into structured diagnostic inputs. We evaluate the current checkpoint on the capability-labeled diagnostic pool and collect its predictions. Each evaluation item is associated with a cognitive capability dimension, and each prediction record stores the item identifier, predicted answer, reference answer, correctness label, and capability metadata. Each item is evaluated multiple times to ensure robustness, resulting in multiple trial records. These structured records allow model errors to be analyzed by capability dimension rather than only as isolated failed questions, and provide the input to Feedback Synthesis.

\subsubsection{Feedback Synthesis}
Feedback Synthesis converts dispersed prediction failures into capability-oriented candidate data. Instead of synthesizing new samples through sample-level revisions or surface-level variations of individual bad cases, FLARE uses the shared cognitive deficiency mechanisms revealed across multiple failures. The stage consists of four steps.

\paragraph{Bad-case Extraction.}
To localize capability deficiencies that limit the current model, FLARE first organizes sample-level failures into structured inputs for capability-level analysis. We collect bad cases from the evaluation results, while recording how often each sample is failed across trials of the current checkpoint. Each failed prediction is retained, and each sample receives a difficulty score equal to the total number of incorrect predictions it receives. This score distinguishes broadly challenging cases from trial-specific sampling noise and serves as the sampling weight in the subsequent diagnosis stage. The weighted bad cases are then aggregated by capability dimension, and the error rate of each dimension is used to allocate diagnosis budgets.

\paragraph{Per-dimension Diagnosis.}
Bad cases indicate where the model fails, but not necessarily the shared capability deficiency behind those failures. Directly synthesizing from individual bad cases would keep generation tied to local semantic variations and produce sample-level augmentation rather than capability-oriented supervision. FLARE therefore randomly samples a batch of bad cases in the same dimension and summarizes the failures into a structured Dimension Diagnosis Report. Specifically, a teacher model analyzes common error patterns and records the primary and secondary failed cognitive operations, recommended synthesis themes, and target difficulty distribution of the sampled cases. This report decouples capability diagnosis from data generation: synthesis is conditioned on the diagnosed capability gap rather than on the surface content of the original failed examples.

\paragraph{Synthesis.}
Conditioned on the Dimension Diagnosis Report, teacher models synthesize new training candidates that target the corresponding cognitive capability deficiency. The goal is to reconstruct the underlying failure pattern in novel semantic contexts, so that solving the synthesized samples requires the same target capability without preserving the surface form of the original bad cases. Different data formats share this diagnosis-to-synthesis interface, with format-specific output constraints and language isolation.

\paragraph{LSH Deduplication.}
After synthesis, FLARE applies a dual deduplication strategy based on MinHash LSH. Candidate samples are first deduplicated against the accumulated diagnostic pool using approximate retrieval followed by 4-gram Jaccard verification ($\theta=0.6$), and then deduplicated within the candidate pool using a stricter threshold ($0.85$). Only samples passing both checks proceed to Data Filtering.

\subsubsection{Data Filtering}
Diagnosis-driven synthesis improves targeting, but candidate samples may still be too easy, unanswerable, mislabeled, ambiguous, or shortcut-dependent. FLARE therefore applies a decision-tree filtering pipeline with iterative repair to retain only samples with genuine training value. The pipeline consists of four steps: Pass@k Difficulty Assessment, Answerability Check, Shortcut Probing, and Repair and Final Selection. 

\paragraph{Pass@k Difficulty Assessment.}
This step estimates the training value of each candidate for the current student model. Each sample is evaluated over $k$ independent trials with randomized option ordering and categorized as \emph{all\_passed}, \emph{partial}, or \emph{all\_failed}. \emph{all\_passed} samples are too easy to provide meaningful training signal; \emph{partial} samples are learnable yet sufficiently challenging; and \emph{all\_failed} samples are either genuinely difficult or potentially defective.

\paragraph{Answerability Check.}
Because failures may arise from either reasoning difficulty or sample-quality defects, a teacher model further checks whether \emph{partial} and \emph{all\_failed} samples admit a consistent answer based solely on the story and question. Each sample is classified as answerable, label error, ambiguous, contradictory premise, or missing information. Only answerable samples are retained for subsequent filtering decisions.

\paragraph{Shortcut Probing.}
For \emph{partial} and answerable samples, FLARE tests whether the target reasoning process is actually required. We compare model performance after removing the story, the question, or the options. A sample that remains correctly answered after removing the story or question does not genuinely rely on the complete context, while a sample that succeeds with options but fails once options are removed suggests reliance on option formatting or elimination strategies. Shortcut-dependent samples are sent to repair.

\paragraph{Repair and Final Selection.}
Finally, FLARE integrates the preceding results and assigns each sample one of five decision labels: easy, medium, hard, shortcut, or bad. Only easy, shortcut, or bad samples enter the repair pipeline according to their respective defects. Successfully repaired samples are re-evaluated until they satisfy the quality criteria or reach the repair limit. The final training set retains all validated medium and hard samples together with the latest repaired version of each sample.

Across iterations, the latest checkpoint serves as the student model, while models with strong performance on the corresponding capability dimensions serve as teacher models. The accepted samples are used in subsequent training, and the updated checkpoint re-enters the loop. In this way, FLARE closes the data flywheel: evaluation reveals capability gaps, diagnosis abstracts them into supervision targets, synthesis constructs new candidates, and filtering validates their training value.

\subsection{Stage 1: General Social-Context Understanding}
\label{sec:zing-stage1}

Stage~1 is the foundation-building phase of our training pipeline. Its goal is to move the model from superficial recognition of mental-state concepts to robust mental-state inference across diverse social contexts. Rather than covering the full SocialMind taxonomy, Stage~1 focuses on foundational ToM-related operations that support later reasoning over interactional dynamics and norm-conditioned judgment. We organize these operations into seven training dimensions: Belief, Desire, Intention, Perspective, Empathy, Communication, and Knowledge. These dimensions strengthen the model's ability to track what different agents know, believe, intend, want, and feel, and to use these latent states when interpreting situated social contexts.

The design principles of Stage~1 are derived from three observations about building reliable social reasoning abilities. First, broad ToM capability depends on reasoning depth: the model must learn explicit intermediate mental-state transitions rather than merely memorizing labels such as ``belief'' or ``intention''. Second, the composition and weighting of training data are more important than raw data volume, since excessive training on already-mastered capabilities yields diminishing returns, while insufficient coverage of weak capabilities preserves systematic reasoning failures. Third, effective foundation building requires preserving previously acquired capabilities through balanced supervision and capability-aware optimization, as improving one aspect of ToM reasoning at the expense of another would result in an unbalanced social reasoner. These observations motivate the data construction, supervised fine-tuning, and reinforcement learning strategies described in the following sections.

\subsubsection{Stage-1 Data Construction}
\label{sec:zing-stage1-data}

Stage-1 training data are constructed with the FLARE data flywheel described in Section~\ref{sec:zing-data-iteration}, which provides capability-targeted story--question--option instances for the current model. Stage~1, however, requires more than reliable question--answer pairs: it also needs reasoning traces that expose the intermediate mental-state updates leading to the answer. We therefore extend FLARE-generated items with capability-specific Chain-of-Thought (CoT)~\cite{wei2022chain} traces and apply trace-level quality control before using them for post-training.

 We instantiate this process over the seven foundational ToM-related training dimensions defined above: Belief, Desire, Intention, Perspective, Empathy, Communication, and Knowledge~\cite{baron1997mindblindness}. These dimensions cover the basic operations needed for later specialization, including belief attribution and false-belief reasoning~\cite{wimmer1983beliefs}, desire and goal understanding, intention inference, perspective-taking, empathic emotion understanding, pragmatic and non-literal language comprehension, and knowledge-state tracking. The Evaluation and Bad-case Extraction stages provide the sampling signal for Stage~1: sample difficulty is estimated from multi-trial performance of the current checkpoint, and errors are aggregated by ToM capability dimension. Consistently solved samples are removed, while weak but learnable dimensions or subcategories receive larger sampling weights through inverse-accuracy reweighting.

During Feedback Synthesis, Stage~1 uses multi-teacher capability routing to construct capability-specific CoT supervision. This routing is needed because different ToM dimensions require different reasoning operations: false-belief items require tracking information access and belief persistence, intention items require linking goals to actions, and higher-order belief items require maintaining nested mental-state representations. Rather than relying on a single general-purpose teacher, each ToM capability is assigned to the teacher model or specialized checkpoint with the strongest diagnostic performance on that dimension. The selected teacher then generates or validates both the final answer and the intermediate reasoning trace, so that each training instance provides supervision from the most reliable source for the corresponding cognitive operation.

Since Stage~1 learns from reasoning traces, accepted FLARE items must be paired with traces that are suitable for supervision and optimization. We therefore apply trace-level quality control after synthesis. Correctness filtering retains only traces whose teacher-generated final answer matches the gold label, preventing post-hoc rationales for incorrect solutions from being distilled or reinforced. Reasoning-depth filtering removes traces that are too shallow to expose intermediate mental-state transitions or excessively long and noisy. Trace deduplication reduces repeated reasoning templates, while capability balancing prevents the training set from being dominated by a small number of easier ToM dimensions. Together, these filters ensure that the retained data supervise explicit mental-state reasoning processes, not only final-answer labels.

\subsubsection{Stage-1 Supervised Fine-Tuning}
\label{sec:zing-stage1-sft}

Given the CoT-augmented Stage-1 data constructed above, supervised fine-tuning trains the model to imitate capability-specific mental-state reasoning trajectories rather than directly mapping inputs to final labels. Each retained trajectory contains a story--question--option instance, a teacher-generated reasoning trace, and a final answer, with the trace explicitly grounding the answer in intermediate updates over agents' mental states, accessible information, intentions, and perspectives.

The objective of Stage-1 SFT is to make these reasoning operations available during generation. The goal is not merely to standardize the response format, but to internalize the intermediate operations needed for broad ToM reasoning: evidence grounding, perspective separation, belief-state updating, intention inference, and knowledge-state tracking. This SFT stage provides the initial parametric foundation that is then further stabilized through reinforcement learning.

\subsubsection{Stage-1 Reinforcement Learning}
\label{sec:zing-stage1-rl}

Stage-1 RL is used to stabilize the broad ToM foundation learned through supervised fine-tuning (SFT). While SFT provides offline reasoning traces that demonstrate desirable mental-state inference patterns, RL exposes the model to its own sampled reasoning trajectories~\cite{instructgpt} and rewards responses that remain valid under generation-time variation. The goal of Stage-1 RL is therefore not to optimize fine-grained social-emotional behaviors, but to improve the reliability and consistency of broad ToM reasoning.

We employ two complementary reward signals within a unified GRPO. For open-ended analytical tasks without explicit ground-truth answers, we adopt a process-reward formulation~\cite{lightman2024let} where an external LLM serves as a judge. We focus the process reward on belief- and knowledge-centered reasoning because these are the most diagnostic failure mechanisms among the Stage-1 foundational operations and serve as prerequisites for higher-level ToM and social reasoning. The judge evaluates responses along five dimensions: explicit-implicit belief recognition, belief source attribution, belief structure graph reasoning, belief-fact discrimination, and dynamic belief tracking. Each dimension is scored from 0 to 2 (total 0--10), with higher scores requiring evidence-anchored mechanistic explanations rather than generic reasoning. A length penalty is also applied to discourage verbosity without substantive reasoning, encouraging concise and well-structured responses.

For ToM tasks with verifiable outcomes, we employ reinforcement learning from verifiable rewards (RLVR)~\cite{guo2025deepseek} on the FLARE-generated Stage-1 data constructed in Section~\ref{sec:zing-stage1-data}. The objective labels of these items provide outcome rewards for model rollouts. To improve learning efficiency, we adopt selective sampling: rather than training on all available data, we retain only mixed-outcome instances, \textit{i.e.}, cases where the model generates both correct and incorrect responses to the same question. These challenging samples provide more informative optimization signals for GRPO and improve robustness on ambiguous social reasoning scenarios.

The two reward types are mixed within the same Stage-1 GRPO training stage, with the reward assigned according to the task format: process reward improves the quality of intermediate mental-state reasoning, while verifiable reward anchors the final answer to objective correctness. Together, they reinforce the broad ToM capabilities acquired during SFT.

Following the established practice of using model-generated responses for post-training, we apply rejection sampling fine-tuning (RFT)~\cite{guo2025deepseek} after Stage-1 SFT and GRPO as a final self-mining step. We sample multiple reasoning trajectories from the trained policy and retain only those that pass correctness, majority-consistency, and reasoning-depth checks. Fine-tuning on these verified self-generated traces reinforces reasoning paths that the model can already produce but does not yet apply consistently, improving the stability and robustness of Stage-1 mental-state reasoning under sampling.


\subsection{Stage 2: Specialized Social Mind Reasoning} 
\label{sec:stage2}
\label{sec:zing-stage2}


Stage~1 establishes a broad foundation for ToM-related mental-state reasoning, but it does not cover the full specialized capability space defined by the SocialMind taxonomy in Section~\ref{sec:socialmind-taxonomy}. Stage~2 therefore starts from the Stage~1 checkpoint and targets residual capability gaps that require reasoning over affect, persona or preference, interaction dynamics, and norm-conditioned judgment. These gaps correspond to specialized parts of the taxonomy, including fine-grained emotion and perspective-sensitive affective reasoning, context-grounded action or response selection, strategic interaction, and role-, norm-, or group-boundary-aware social judgment.

Stage~2 follows a sequential post-training path. Capability-oriented SFT first injects strong-teacher reasoning traces for specialized but teachable cases, such as emotion-cause binding, persona-conditioned affect, and context-grounded social response selection. Joint OPD and mixed-reward GRPO then refine the remaining unstable capabilities by combining teacher-guided process regularization with outcome and rubric rewards, especially for cases where multiple answers appear plausible and the validity of a response depends on the intermediate social reasoning process.

\subsubsection{Stage-2 Capability Diagnosis and Data Construction}
\label{sec:stage2-data}


Stage~2 uses the same FLARE data flywheel as Stage~1, but changes the diagnostic target from foundational ToM-related operations to residual specialized SocialMind capabilities. After Stage~1 training, the model shows stronger general mental-state tracking, but remains brittle when affect, preference, social relation, or contextual constraints must determine the answer. For example, it often comes back to surface common sense templates, such as associating success with happiness or failure with disappointment, even when the story defines an atypical preference or conflicting context.  Another issue is weak emotion-cause binding: the model may select a plausible emotion while missing the true psychological trigger. Similar template-driven behavior appears in action and response items, where generic comfort or complaint is chosen instead of a context-sensitive social decision.
These failures are most salient in complex-emotion settings involving mixed feelings, irony, surprising outcomes, or value conflicts. We map these failures to trainable Stage~2 capabilities, including emotion recognition, cause attribution, visibility-aware affect reasoning, atypical-preference/persona reasoning, complex emotion understanding, social action prediction, and response selection. These capabilities instantiate the residual mentalizing and interactional parts of the broader SocialMind taxonomy.

\paragraph{Capability-oriented synthesis.}
Following the data flywheel in Section~\ref{sec:zing-data-iteration}, we use DeepSeek-Reasoner~\cite{guo2025deepseek} to synthesize targeted social-mind training cases from the Stage~2 capability-gap diagnosis. The synthesis focuses on three groups of fine-grained capabilities reflected in our benchmark taxonomy and Stage-2 diagnostic results: affective and persona-conditioned reasoning, perspective- and context-grounded interaction reasoning, and norm-, role-, and group-boundary-aware social judgment. All generated instances are converted into a unified story--question--option schema and then passed through near-duplicate control, difficulty, answerability, shortcut, and format filters.

\paragraph{Strong-teacher CoT distillation.}
We use Qwen3-235B-A22B\footnote{https://huggingface.co/Qwen/Qwen3-235B-A22B} as the teacher model and enable its thinking mode so that each accepted sample contains both an explicit specialized social-mind reasoning trace and a final boxed answer. Unlike Stage~1 distillation, which uses multi-teacher capability routing to cover broad foundational training dimensions, Stage~2 uses a same-family strong teacher to generate compatible reasoning traces for residual specialized SocialMind capabilities. The same-family teacher keeps the chat template, thinking format, and \verb|\boxed{}| answer convention compatible with the Qwen student model.
Teacher outputs are accepted only under gold consistency: for single items, the last \verb|\boxed{}| must match the gold answer; for multi-question items, all boxed answers must match the corresponding gold labels in order. The full SFT message, including prompt, story, questions, options, CoT, and final
answer, is capped at 4096 tokens. As a result, the SFT data do not simply provide answer labels; they provide correctness-constrained reasoning traces that teach the model how to compare perspectives, bind emotions to causes, condition judgments on persona and context, and select context-appropriate actions or responses.

\subsubsection{Stage-2 Supervised Fine-Tuning}
\label{sec:stage2-sft}



Stage-2 SFT is the first specialization step after Stage~1. Given the CoT-augmented Stage~2 data constructed above, its goal is to specialize the Stage~1 checkpoint toward residual SocialMind capabilities that require affective, persona-conditioned, interactional, and norm-aware reasoning. The model learns to condition its answer on the target character's belief state, available information, preference, social relation, role or norm constraint, and local context before selecting an emotion, action, judgment, or response.
We organize Stage-2 SFT into two consecutive rounds. The first round injects the main targeted capability signal using distilled specialized SocialMind data, moving the model from broad ToM abilities toward fine-grained affective, interactional, and context-sensitive social judgment. The second round is an incremental calibration step over harder cases retained by the Stage~2 filtering pipeline. It reinforces newly learned reasoning patterns while reducing the risk of forgetting the broader social-cognitive abilities acquired in Stage~1. Both rounds use full-parameter SFT with correctness-constrained CoT data. Detailed hyperparameters are reported in the appendix.


This SFT stage encourages the model to rely less on narrator-level knowledge or generic social-emotion templates, and more on the character's accessible information, preference, expectation, social relation, role or norm constraint, and local context. It therefore provides the offline specialization signal that is further stabilized by the next OPD and GRPO stage under on-policy generation.

\subsubsection{Stage-2 Joint OPD and GRPO Post-Training}
\label{sec:stage2-opd-grpo}

After Stage-2 SFT, the model has learned correctness-constrained reasoning traces for many specialized cases, but several capabilities remain unstable, especially in strategic interaction and norm-conditioned judgment. Examples include negotiation, game-theoretic reasoning, trust and deception, cooperation and conflict, group dynamics, and group-boundary reasoning. To address these gaps, we jointly apply On-Policy Distillation (OPD) and mixed-reward GRPO, moving the model from offline imitation to an online policy optimized for reasoning-process quality and answer stability.

\paragraph{Motivation for Joint OPD and GRPO.}
Our early experiments indicate that using either OPD or GRPO alone is insufficient. While OPD provides token-level distillation from a teacher model to regularize local process quality, its optimization is strictly bounded by the teacher's capability and lacks direct supervision on the final answer's correctness. Conversely, GRPO optimizes the policy directly against sequence-level task outcomes via group-relative advantages; however, it struggles when reward variance is low (e.g., all samples are correct or incorrect) and lacks fine-grained token-level constraints. Motivated by this, we integrate them into a unified on-policy objective. OPD ensures the fidelity of the reasoning process through expert guidance, while mixed-reward GRPO optimizes final generation quality via outcome and rubric rewards. This integration is crucial for Stage-2 tasks, which intrinsically rely on both logically valid mental-state trajectories and accurate final decisions.

\paragraph{Mixed-Reward GRPO Design.}
The GRPO reward design follows the same motivation as the Stage-2 specialization step: the model should not only preserve answer accuracy, but also improve the mental-state process that makes an answer socially valid. Stage~1 already uses verifiable outcome reward as a lightweight signal. It checks whether the model produces a valid reasoning block and a final answer that matches the reference label, with small penalties for uncertainty, multiple candidate answers, and excessive verbosity. This signal is useful for stabilizing the answer channel on label-verifiable social ToM items. However, it remains limited for Stage~2 because final-answer correctness alone cannot certify fine-grained social mind reasoning. A model may select the correct option through a lexical cue, a generic social prior, or narrator-level information, while failing to track what the target character actually knows, believes, feels, or intends.

Stage~2 therefore introduces rubric reward as the main process-quality signal for reasoning-heavy social-mind items. Rubrics make the latent validity conditions of social reasoning explicit. For each item, we augment the question with binary criteria generated from the question, reference answer, and ground truth,
following rubric-guided reward methods that use explicit criteria to evaluate reasoning quality~\cite{rubricguidance}. The criteria are organized into factual and process checks. Factual criteria verify answer content, such as the final belief, emotion, object, action, or social judgment. Process criteria verify the mental-state path leading to that answer: what the character could observe, whose perspective is being used, whether a belief should persist, which psychological cause supports an emotion or action, what non-literal intent is implied, or what social norm is relevant. In this way, Stage~2 rubrics extend the Stage~1 outcome-reward signal from final-answer verification to explicit process-level supervision over targeted social-mind operations.

During rollout, the reward worker extracts the model's reasoning content and final answer, removes prompt echoes and chat-template artifacts, and sends the cleaned response to an online binary judge. If $c_j\in\{0,1\}$ is the rating for criterion $j$ and $w_j$ is its weight, the rubric score $R_\mathrm{rubric}$ is computed as 
\begin{equation}
  R_\mathrm{rubric} = \frac{\sum_j w_j c_j}{\sum_j w_j},
\end{equation}
For factual criteria, the scoring follows the same item-level aggregation rule: when all factual checks are satisfied, the answer-content part receives full credit; otherwise, the unsatisfied factual checks simply contribute zero through the weighted rubric average. We do not add a separate gate that forces the outcome score to zero when a factual criterion fails.

In the mixed Stage-2 reward setup, outcome-scored items and rubric-scored items serve different purposes rather than being presented as a single closed-form reward decomposition. Outcome reward preserves answer accuracy on label-verifiable cases, rubric reward provides the main process-quality signal for targeted social-mind reasoning, and the format reward remains auxiliary. In practice, early Stage-2 RL iterations emphasize outcome reward to keep the answer channel stable; later iterations increase the relative pressure from rubric-scored items so that optimization focuses more directly on perspective tracking, emotion--cause consistency, causal grounding, and socially appropriate response selection.

\paragraph{Joint Training Objective.}
We formulate Stage-2 post-training as GRPO with an OPD regularizer on the same on-policy trajectories. Let $A_i$ denote the GRPO advantage of the $i$-th sampled response, and let $h_{i,t}=(x,y_{i,<t})$ be the token history for its $t$-th generated token. For each generated token $y_{i,t}$, the teacher provides its log-probability $\log \pi_T(y_{i,t}\mid h_{i,t})$. The OPD regularizer modifies the token-level advantage as
\begin{equation}
  \widetilde{A}_{i,t}
  =
  A_i
  -
  \lambda_{\mathrm{OPD}}
  \left[
    \log \pi_\theta(y_{i,t}\mid h_{i,t})
    -
    \log \pi_T(y_{i,t}\mid h_{i,t})
  \right],
  \label{eq:stage2_opd_grpo_advantage}
\end{equation}
where $\lambda_{\mathrm{OPD}}$ controls the strength of teacher guidance. In this formulation, GRPO uses reward differences among the student's own on-policy samples to identify more effective reasoning processes, while OPD keeps the sampled trajectory close to the teacher's token-level preference on the same visited states. As a result, Stage-2 post-training improves reasoning-process quality and answer stability without relying solely on reward optimization or teacher imitation.

\paragraph{Progressive OPD-to-GRPO Scheduling.}

We organize Stage-2 post-training as a capability-progressive optimization procedure. Earlier rounds emphasize OPD for capabilities where teacher traces are reliable, such as affective inference, emotion recognition, cause attribution, and emotion-cause binding. These rounds transfer stable affective reasoning patterns and stabilize the student's reasoning format. Later rounds increase the relative weight of GRPO for more complex SocialMind capabilities, including negotiation, game-theoretic reasoning, trust and deception, cooperation and conflict, group dynamics, identity and role, and group-boundary reasoning. These tasks often admit multiple plausible interpretations, where small differences in perspective tracking, social constraints, or interaction dynamics can affect response validity. We use the previous checkpoint to select medium-difficulty samples with mixed correct and incorrect responses, increasing reward variation within each prompt group.

Through this procedure, Stage-2 post-training progressively strengthens specialized SocialMind capabilities beyond the broad ToM foundation established in Stage~1.

\subsection{Evaluation of \modelname{}}

\subsubsection{Experiment Setup}

\paragraph{Benchmarks} We evaluate \modelname{} on five  social cognition benchmarks: ToMBench \cite{tombench}, EmoBench \cite{sabour2024emobench}, FANToM \cite{fantom}, HiToM \cite{hitom}, together with our newly proposed benchmark  SoMBench.

\begin{table}[ht]
\centering
\small
\caption{Overview of  evaluation  datasets.}
\vspace{-1em}
\label{tab:benchmarks}
\begin{tabular}{lclcc}
\toprule
\textbf{Benchmark} & \textbf{Language} & \textbf{Questions} & \textbf{Format} & \textbf{Source} \\
\midrule
ToMBench & en,zh & 5,720 & MCQ & \citep{tombench} \\
EmoBench & en,zh & 1,200 & MCQ & \citep{sabour2024emobench} \\
HiToM & en & 1,200 & MCQ & \citep{hitom} \\
FANToM & en & 11,292 & MCQ & \citep{fantom} \\ 
SoMBench & zh & 3481 & MCQ + OpenQA & - \\
\bottomrule
\end{tabular}
\end{table}

\Cref{{tab:benchmarks}} provides the statistics of these datasets. ToMBench provides a systematic evaluation of ToM reasoning across eight tasks covering 31 social-cognitive abilities. EmoBench evaluates emotional intelligence through emotion understanding and emotion application tasks. FANToM focuses on theory-of-mind reasoning in information-asymmetric multi-party conversations, requiring models to infer the beliefs and knowledge states of different participants. HiToM evaluates higher-order theory-of-mind reasoning through recursive reasoning over multiple agents' mental states. Together, these benchmarks cover complementary aspects of social cognition, including belief reasoning, emotion understanding, information accessibility, and higher-order mental-state inference. Our social mind reasoning benchmark SoMBench covers 3 primary dimensions, 17 secondary dimensions, and 71 fine-grained task paradigms, providing a unified evaluation of both cognitive and affective aspects of social intelligence.

\paragraph{Evaluation Protocol} We evaluate all models using \evalname \footnote{\evalurl}, a unified evaluation framework for ToM and social-cognition benchmarks. To ensure consistent comparisons across datasets and model families, all public benchmarks are evaluated under a unified multiple-choice (MCQ) protocol. Each question is answered once with chain-of-thought reasoning enabled, and the final prediction is extracted from the boxed answer for automatic scoring.

Specifically, we evaluate the complete bilingual release of ToMBench (5,720 questions). For EmoBench, each Emotion Understanding example is decomposed into separate emotion-understanding and emotion-cause reasoning questions, resulting in 1,200 evaluation questions. HiToM contains 1,200 English multiple-choice questions. FANToM originally includes free-form, multiple-choice, and yes/no questions; we reformulate all question types into a unified multiple-choice format using the provided reference answers, yielding 256 dialogue scenarios and 11,292 evaluation questions. SoMBench contains both multiple-choice and open-ended questions. Multiple-choice questions are evaluated using the same rule-based answer matching protocol as the public benchmarks. Open-ended questions are evaluated using an LLM-as-a-Judge protocol with manually designed scoring rubrics specifying the expected reasoning process and key semantic criteria. The judge assigns a normalized score to each response, and responses above a predefined threshold are regarded as correct.

\paragraph{Models} 
We evaluate the proposed \modelname{} model family, including \modelname{}-8B, \modelname{}-14B, \modelname{}-32B, and \modelname{}-27B (multimodal). These models are post-trained from the corresponding Qwen3\footnote{https://huggingface.co/collections/Qwen/qwen3} series and Qwen3.6\footnote{https://huggingface.co/Qwen/Qwen3.6-27B} base models, using ROLL~\cite{wang2025reinforcement} as the primary training framework.
Each \modelname{} model is compared with its corresponding Qwen3 or Qwen3.6 base model to measure the effectiveness of our post-training recipe. We also include several frontier large language models, DeepSeek-V4-Flash\footnote{https://huggingface.co/deepseek-ai/DeepSeek-V4-Flash}, DeepSeek-V4-Pro\footnote{https://huggingface.co/deepseek-ai/DeepSeek-V4-Pro}, and GPT-5.5, as reference under the same evaluation protocol.

\paragraph{Evaluation Parameters} All models are evaluated with a maximum generation length of 32,768 tokens, unless otherwise specified. We report pass@1 results using a sampling temperature of $0.6$ and top-$p = 0.95$. Each evaluation instance is answered with a single sampled response and scored according to the corresponding benchmark-specific evaluation protocol.

\begin{table}[th]
\centering
\caption{Performance comparison across models and model scales on five benchmarks.}
\footnotesize  
\setlength{\tabcolsep}{4.8pt}
\renewcommand{\arraystretch}{1.05}
\resizebox{\textwidth}{!}{%
\begin{tabular}{lrrrrrr}
\toprule
Model & HiToM & ToMBench & EmoBench & FANTom & SoMBench & Avg. \\
\midrule
Qwen3-8B & 66.33 & 74.70 & 54.50 & 79.61 & 49.24 & 64.88 \\
Qwen3-14B & 72.00 & 76.14 & 62.13 & 83.16 & 52.71 & 69.23 \\
Qwen3-32B & 74.33 & 78.30 & 63.50 & 82.91 & 52.66 & 70.34 \\
Qwen3.6-27B (multimodal) & 77.08 & 81.63 & 69.75 & 86.26 & 62.57 & 75.46 \\
\midrule
\modelname{}-8B-Stage1 & 79.08 & 75.80 & 57.63 & 83.67 & 47.34 & 68.70 \\
\modelname{}-14B-Stage1 & 80.00 & 78.36 & 64.63 & 85.87 & 53.32 & 72.44 \\
\modelname{}-32B-Stage1 & 78.42 & 78.88 & 74.50 & 85.88 & 56.56 & 74.85 \\
\modelname{}-27B-Stage1 & \textbf{82.92} & 82.55 & 76.00 & \textbf{86.97} & 66.96 & \underline{79.08} \\
\midrule
\modelname{}-8B-Stage2 & 78.08 & 77.15 & 70.13 & 85.29 & 52.48 & 72.63 \\
\modelname{}-14B-Stage2 & 80.00 & 77.47 & 74.25 & 84.96 & 53.17 & 73.97 \\
\modelname{}-32B-Stage2 & 80.42 & 79.67 & \underline{78.13} & 86.37 & 57.00 & 76.32 \\
\modelname{}-27B-Stage2 & \underline{82.00} & 82.57 & \textbf{80.00} & \underline{86.96} & \underline{67.46} & \textbf{79.80} \\
\midrule
GPT-5.5 & 77.92 & \textbf{84.56} & 74.38 & 86.65 & \textbf{68.69} & 78.44 \\
DeepSeek-V4-Pro & 76.42 & 82.33 & 71.88 & 85.36 & 63.52 & 75.90 \\
DeepSeek-V4-Flash & 74.08 & \underline{82.81} & 72.13 & 84.93 & 59.90 & 74.77 \\
\bottomrule
\end{tabular}}

\label{tab:benchmark_comparison}
\end{table}

\subsubsection{Main Evaluation Results}

Table~\ref{tab:benchmark_comparison} presents a comprehensive comparison of model performance across five social cognition benchmarks. We evaluate the \modelname{} model series at three parameter scales (8B, 14B, and 32B) against the corresponding base Qwen3 models and several leading large language models, including GPT-5.5, DeepSeek-V4-Pro, and DeepSeek-V4-Flash. Beyond the text-only \modelname{} models, we also develop \modelname{}-27B based on the Qwen3.6-27B multimodal foundation model. We draw the following key observations.




\paragraph{\modelname{} Consistently  Improves over the Corresponding Base Models.}
Our \modelname{} models demonstrate substantial improvements over the  corresponding base Qwen3 models. After Stage~1 training, all \modelname{} models have  outperformed their Qwen3 counterparts in terms of the average performance over five datasets. Notably, \modelname{}-27B-Stage1 achieves 66.96\%,  outperforming Qwen3.6-27B by 4.39 points on SoMBench. Stage~2 training further elevates the social cognitive performance, improving the average scores over the base models by 7.75, 4.74, 4.34, and 5.98 points at the 8B, 14B, 27B, and 32B scales, respectively. The largest relative gain appears at the 8B scale, suggesting that diagnosis-driven post-training is especially helpful for smaller models whose social-cognitive capabilities are less developed. Across individual benchmarks, the most notable improvements are observed on EmoBench, where \modelname{}-8B-Stage2 reaches 70.13\% and outperforms the substantially larger Qwen3-32B by 6.63 points.

\paragraph{\modelname{} Models Are Competitive with Frontier Baselines.}
\modelname{} models achieve competitive performance against frontier LLMs, with particularly strong results on emotion-related reasoning. \modelname{}-27B-Stage2 reaches 80.00\% on EmoBench, outperforming GPT-5.5, DeepSeek-V4-Flash, and DeepSeek-V4-Pro; it also performs comparably to the DeepSeek-V4 models on ToMBench and slightly surpasses GPT-5.5 on FANToM. Among the text-only variants, \modelname{}-32B-Stage2 surpasses DeepSeek-V4-Pro in the average performance, approaches GPT-5.5, and achieves the highest EmoBench result among all text-only models. 

\paragraph{Stage 1 and Stage 2 Play Complementary Roles.}
Stage~1 training consistently improves the  average performance over the base models and achieves particularly strong gains on HiToM and EmoBench, which evaluate higher-order theory-of-mind reasoning over multiple agents' mental states and emotion-related understanding and application, respectively. Stage~2 serves as a targeted specialization phase and further enhances the average  performance, but the effect is not strictly monotonic for every  benchmark. For example, \modelname{}-27B-Stage2 improves substantially on EmoBench compared with Stage 1, but slightly decreases on HiToM relative to \modelname{}-27B-Stage1. We provide a more detailed analysis of the stage-wise training effects in the following section.


\subsubsection{Stage-wise Capability Analysis }
In this section, we present detailed experimental analyses for Stage~1 and Stage~2, to provide a deeper understanding of how each stage contributes to the overall social cognitive capabilities of our \modelname{} series models.

\subsubsection*{Stage 1: Enhancing General Social Mind Ability}

Stage 1 aims to improve the model's general social cognitive ability by strengthening its capacity of inferring and representing others' mental states. To evaluate this objective, we conduct a fine-grained analysis across seven fundamental dimensions of social cognition: Belief, Desire, Intention, Perspective, Empathy, Communication, and Knowledge. Evaluation samples are aggregated from HiToM, ToMBench, EmoBench, FANToM, and SoMBench according to their corresponding capability labels, enabling a unified and dimension-level assessment beyond individual benchmark scores.

\begin{figure*}[htbp]
    \centering
    \begin{subfigure}[t]{0.3\textwidth}
        \centering
        \includegraphics[width=\textwidth]{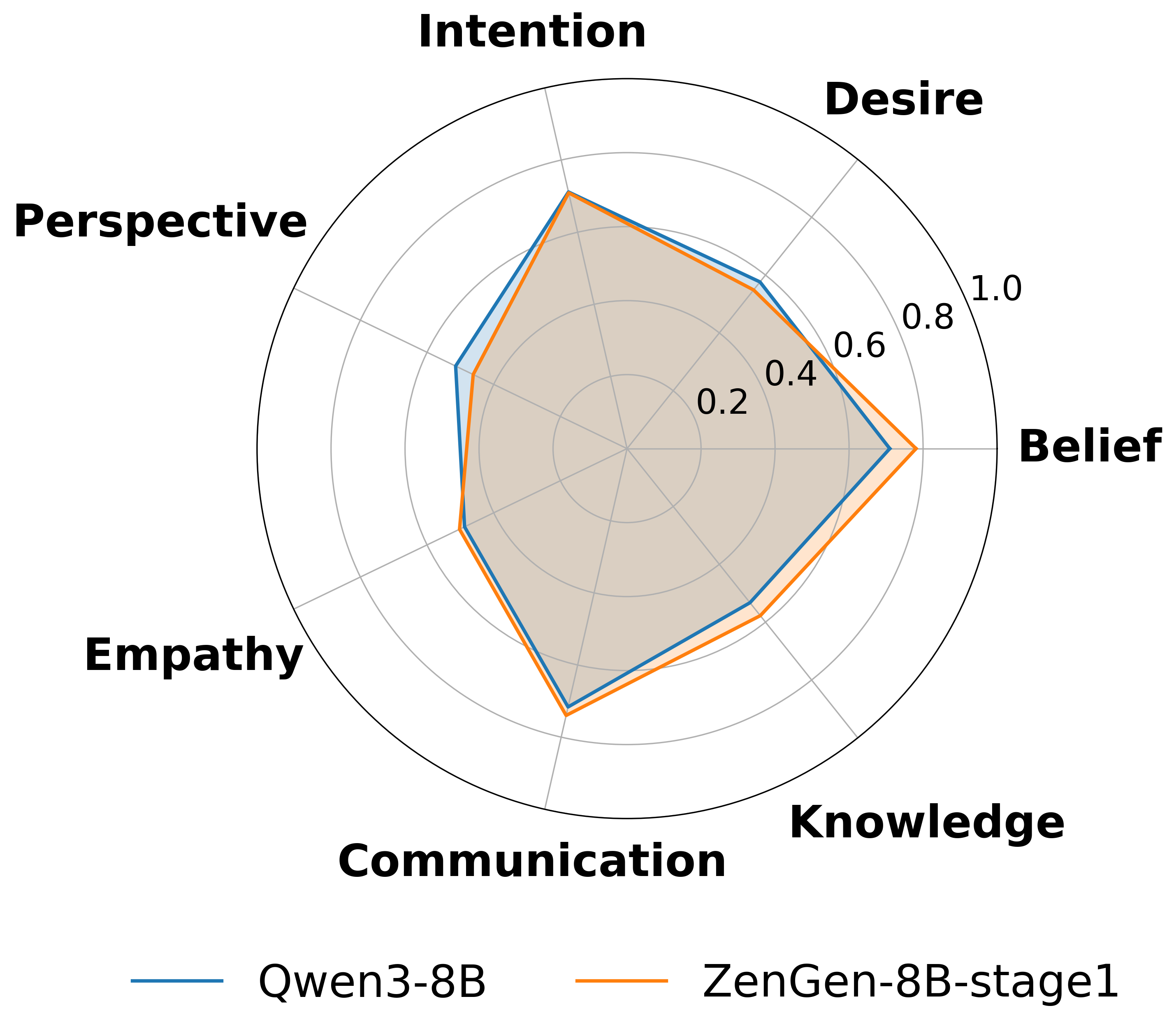}
        \caption{8B-Scale}
    \end{subfigure}
    \begin{subfigure}[t]{0.3\textwidth}
        \centering
        \includegraphics[width=\textwidth]{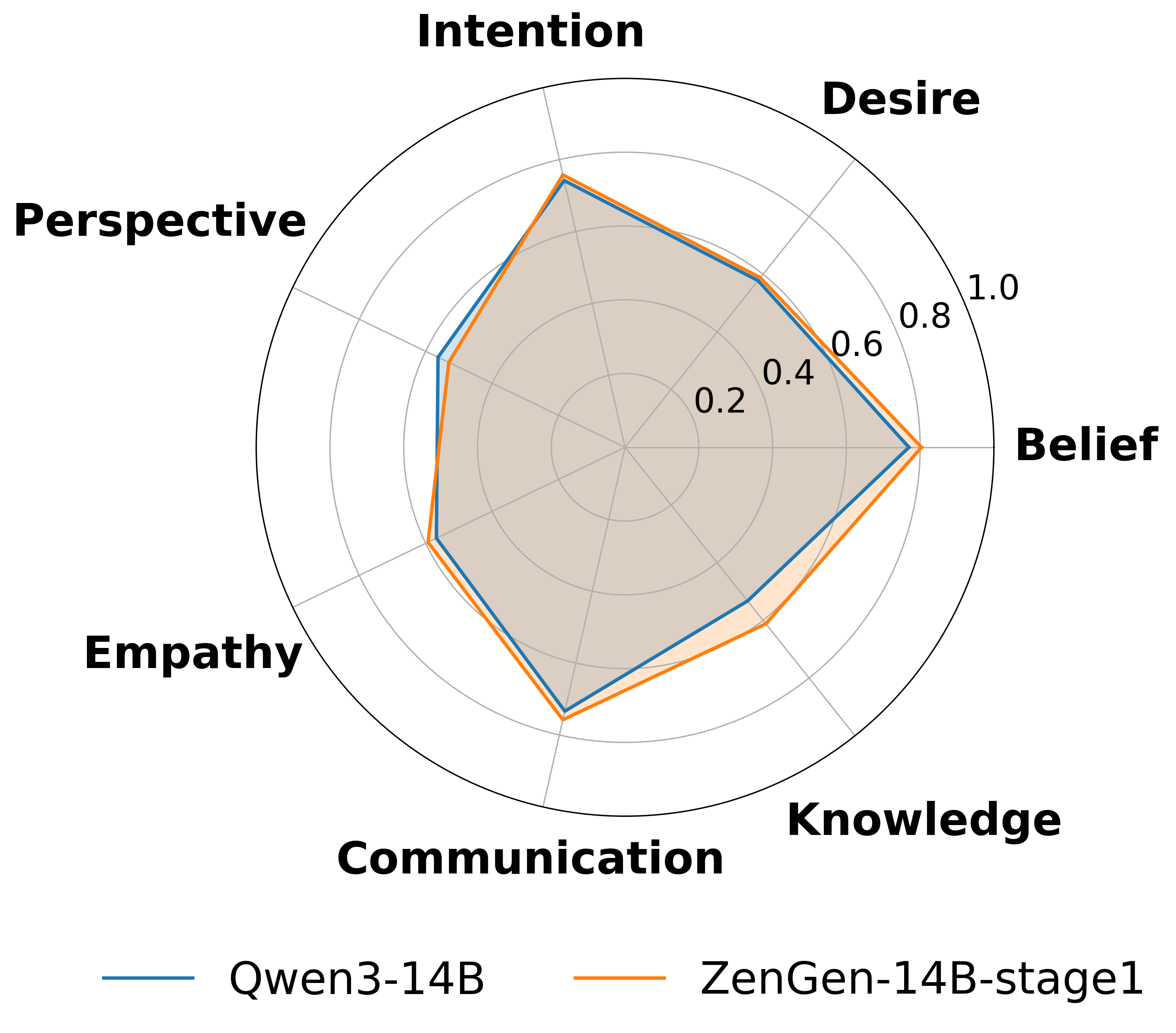}
        \caption{14B-Scale}
    \end{subfigure}

    \vspace{6pt}

    \begin{subfigure}[t]{0.3\textwidth}
        \centering
        \includegraphics[width=\textwidth]{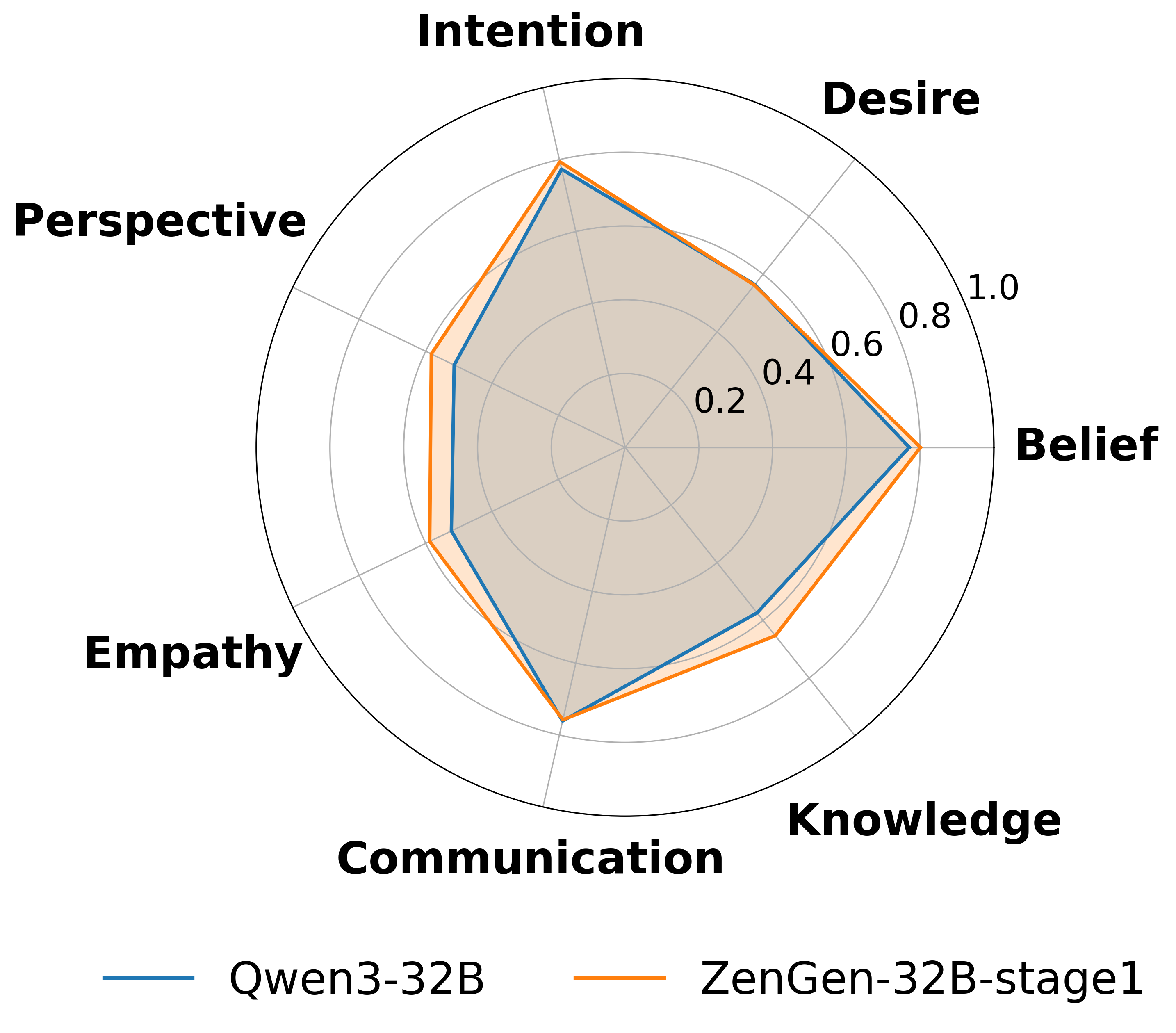}
        \caption{32B-Scale}
    \end{subfigure}
    \begin{subfigure}[t]{0.3\textwidth}
        \centering
        \includegraphics[width=\textwidth]{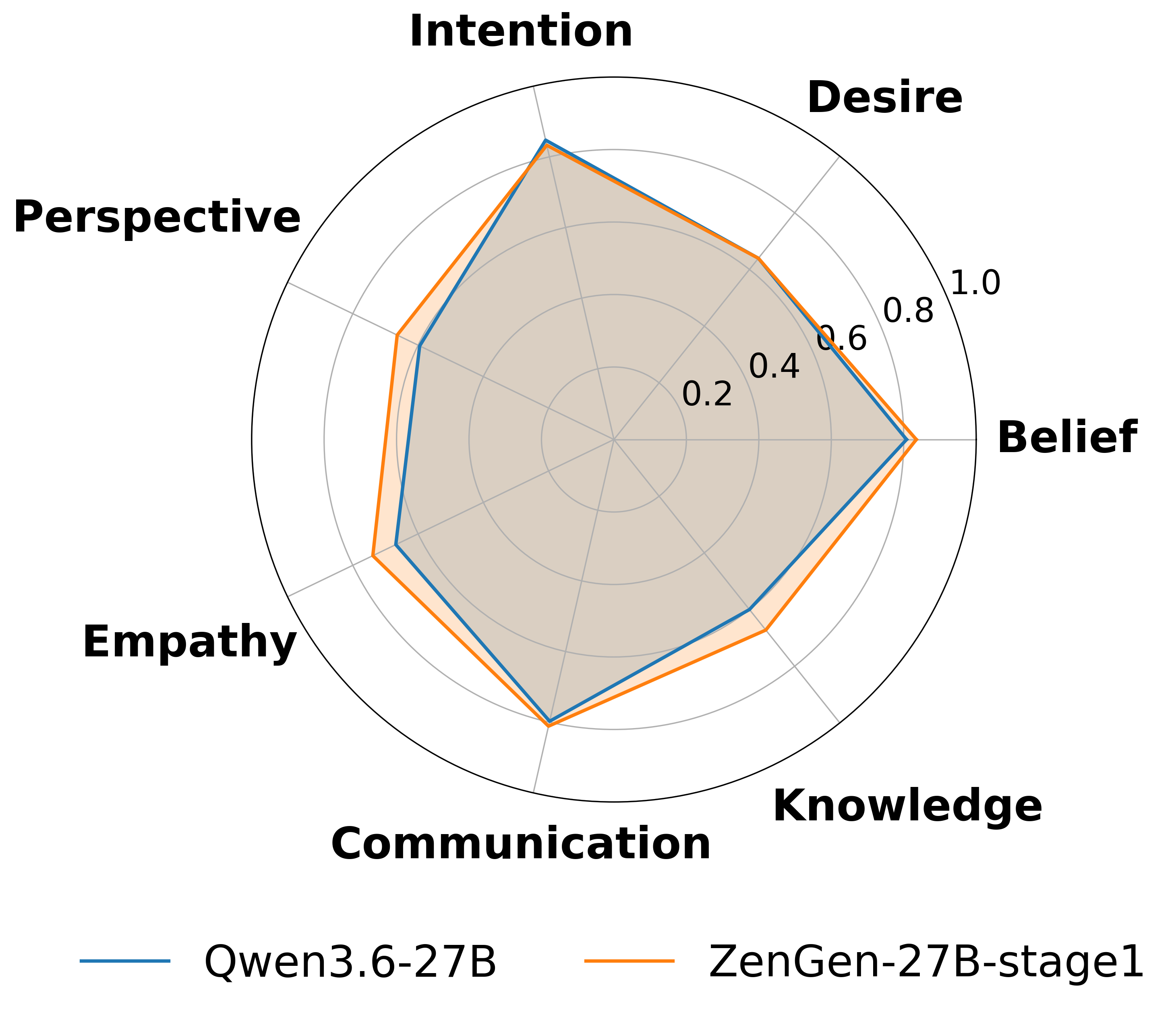}
        \caption{27B-Scale}
    \end{subfigure}

    \caption{Experimental results of the general Social Mind ability enhancement in Stage 1.}
    \label{fig:evaluation_stage1}
\end{figure*}

\paragraph{Overall Performance} Figure \ref{fig:evaluation_stage1} presents the performance of \modelname{} models before and after Stage 1 training. Stage 1 consistently improves social cognitive performance across model scales, with the most substantial gains concentrated in Belief and Knowledge reasoning. For example, \modelname{}-8B-Stage1 improves Belief reasoning from 71.00\% to 78.09\%, while \modelname{}-32B-Stage1 improves Knowledge reasoning from 57.41\% to 65.34\%. Similar improvements are also observed in Empathy for the 14B and 27B models. In contrast, gains on Desire, Perspective, and Communication are relatively modest. This  suggests that Stage 1 primarily enhances latent mental-state inference, such as belief attribution and knowledge-state tracking. The stronger improvements in Belief and Knowledge are consistent with the design objective of learning transferable mental-state representations that can generalize across diverse ToM tasks.

\paragraph{Higher-order Belief Reasoning} To further understand how Stage~1 affects recursive social reasoning, we analyze belief attribution across four reasoning orders. As shown in Table~\ref{tab:belief_orders_stage1}, the performance gains become progressively larger as reasoning depth increases. \modelname{}-8B-Stage1 improves Order-3 belief reasoning from 45.83\% to 67.50\% (+21.67 points) and Order-4 reasoning from 43.75\% to 65.83\% (+22.08 points). Similar trends are observed for the 14B, 27B and 32B models. These results indicate that Stage 1 is particularly effective in enhancing higher-order belief reasoning. Higher-order tasks require models to recursively maintain nested mental representations (e.g.,``A believes that B believes that C knows ...''), which is substantially more challenging than first-order belief attribution. The progressively larger gains from Order-1 to Order-4 therefore suggest that Stage 1 strengthens the model's ability to construct and manipulate recursive mental-state structures, rather than merely improving shallow belief recognition.

\begin{table}[ht]
\centering
\caption{Stage 1 performance on belief reasoning across four orders of belief attribution.}
\footnotesize
\setlength{\tabcolsep}{3.5pt}
\begin{tabular}{lcccccccccccc}
\toprule

\multirow{2}{*}{\textbf{Order}} 
& \multicolumn{3}{c}{\textbf{8B Scale}} 
& \multicolumn{3}{c}{\textbf{14B Scale}} 
& \multicolumn{3}{c}{\textbf{27B Scale}} 
& \multicolumn{3}{c}{\textbf{32B Scale}} \\
\cmidrule(lr){2-4} \cmidrule(lr){5-7} \cmidrule(lr){8-10} \cmidrule(lr){11-13}
& \textbf{Base} & \textbf{Stage 1} & \textbf{$\Delta$} 
& \textbf{Base} & \textbf{Stage 1} & \textbf{$\Delta$} 
& \textbf{Base} & \textbf{Stage 1} & \textbf{$\Delta$} 
& \textbf{Base} & \textbf{Stage 1} & \textbf{$\Delta$} \\
\midrule
Order-1 & 82.92 & 88.75 & +5.83 & 91.25 & 91.67 & +0.42 & 88.75 & 92.50 & +3.75 & 90.00 & 85.00 & -5.00 \\
Order-2 & 59.17 & 73.33 & +14.16 & 62.08 & 72.50 & +10.42 & 72.50 & 77.50 & +5.00 & 65.83 & 71.67 & +5.84 \\
Order-3 & 45.83 & 67.50 & +21.67 & 52.92 & 70.83 & +17.91 & 65.00 & 75.00 & +10.00 & 58.75 & 67.92 & +9.17 \\
Order-4 & 43.75 & 65.83 & +22.08 & 53.75 & 65.00 & +11.25 & 59.17 & 69.58 & +10.41 & 57.08 & 67.50 & +10.42 \\
\bottomrule
\end{tabular}

\label{tab:belief_orders_stage1}
\end{table}

\paragraph{Effect of Model Scale} We also observe that smaller models benefit more substantially from Stage 1 training. For instance, the 8B model exhibits gains of over 20 points on Order-3 and Order-4 reasoning, whereas the 32B model shows  moderate improvements. This trend suggests that Stage 1 effectively compensates for limited inherent social reasoning capacity in smaller models, while larger models may already possess relatively strong first-order reasoning abilities prior to post-training. Notably, the 32B model shows a slight decrease on Order-1 belief reasoning (90.00\% to 85.00\%), which may reflect a trade-off toward complex recursive reasoning.

In summary, the Stage~1 training strategy successfully enhances the fundamental components of social cognition, particularly belief attribution, knowledge-state reasoning, empathy, and higher-order mental-state modeling. However,the improvements remain primarily concentrated on these foundational capabilities. More specialized forms of social reasoning, including negotiation, cooperation, conflict resolution, social norms, identity, power structures, and group dynamics, still exhibit relatively limited improvement. These observations motivate Stage 2, which shifts from broad social cognitive enhancement to targeted refinement of advanced social reasoning abilities across the full spectrum of social interaction scenarios.

\subsubsection*{Stage 2: Enhancement of Specialized Social Mind Ability}

Stage 2 aims to refine the model's specialized social mind capabilities by transitioning from general mental-state inference to targeted reasoning about complex interpersonal dynamics, social structures, and strategic interactions. To evaluate this objective, we conduct a fine-grained analysis across ten advanced dimensions of social cognition: Emotion, Negotiation, Game Theory Reasoning, Cooperation-Conflict, Trust-Deception, Group Dynamics, Norm Internalization, Identity-Role, Power-Institution, and Group Boundary. Evaluation samples are aggregated from ToMBench, EmoBench, and SoMBench according to their corresponding capability labels.

\begin{figure*}[htbp]
    \centering
    \begin{subfigure}[t]{0.4\textwidth}
        \centering
        \includegraphics[width=\textwidth]{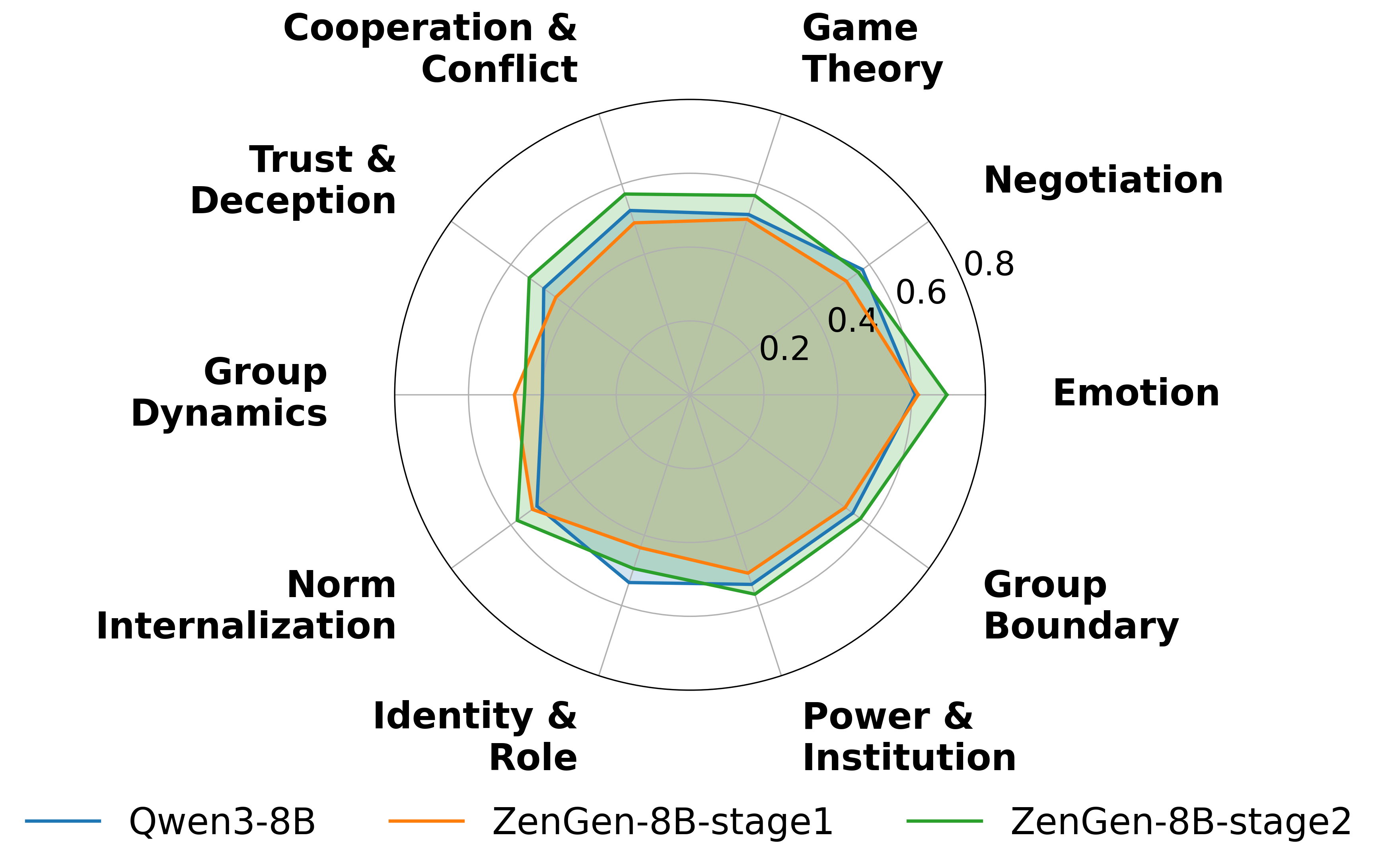}
        \caption{8B-Scale}
    \end{subfigure}
    \begin{subfigure}[t]{0.4\textwidth}
        \centering
        \includegraphics[width=\textwidth]{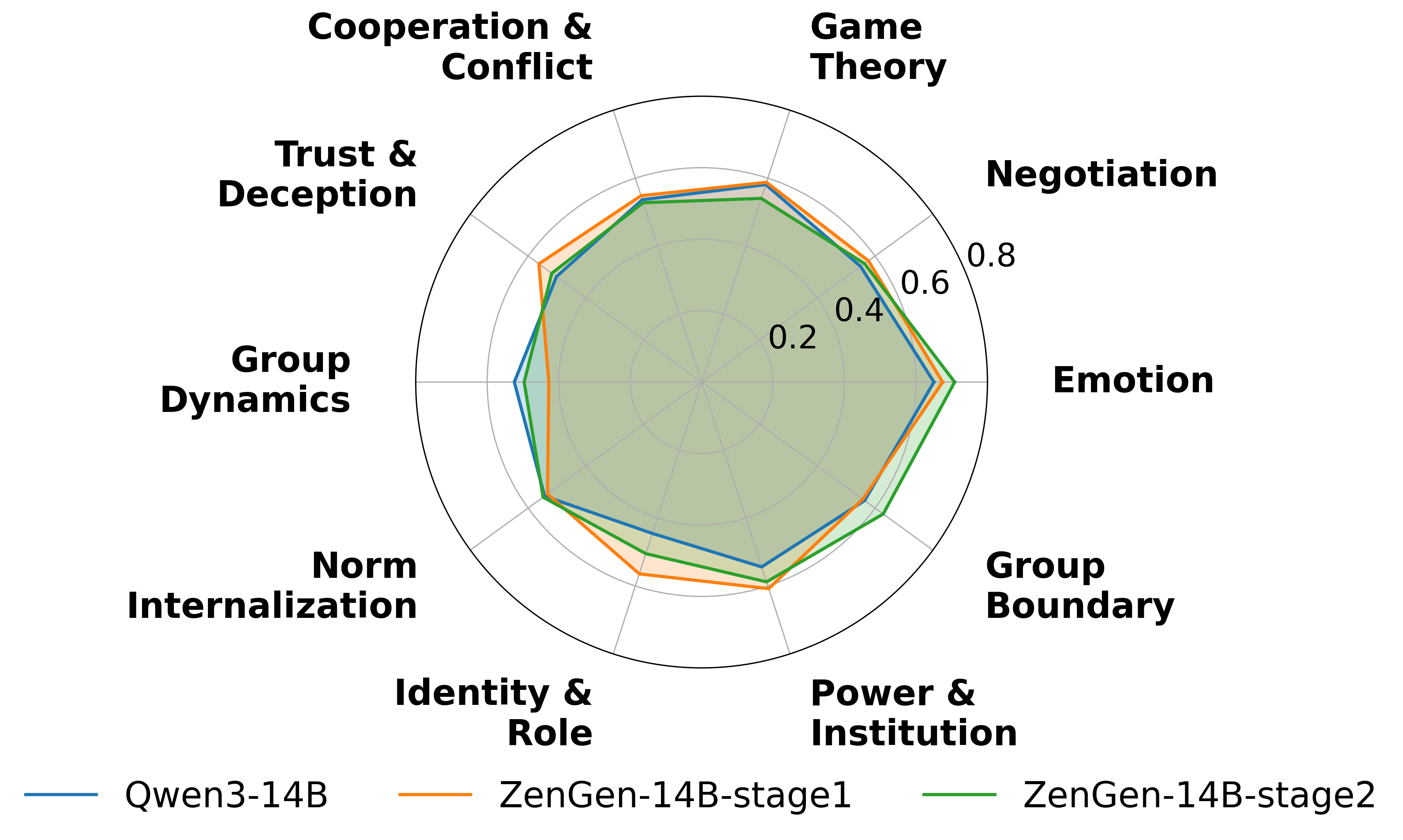}
        \caption{14B-Scale}
    \end{subfigure}

    \vspace{6pt}

    \begin{subfigure}[t]{0.4\textwidth}
        \centering
        \includegraphics[width=\textwidth]{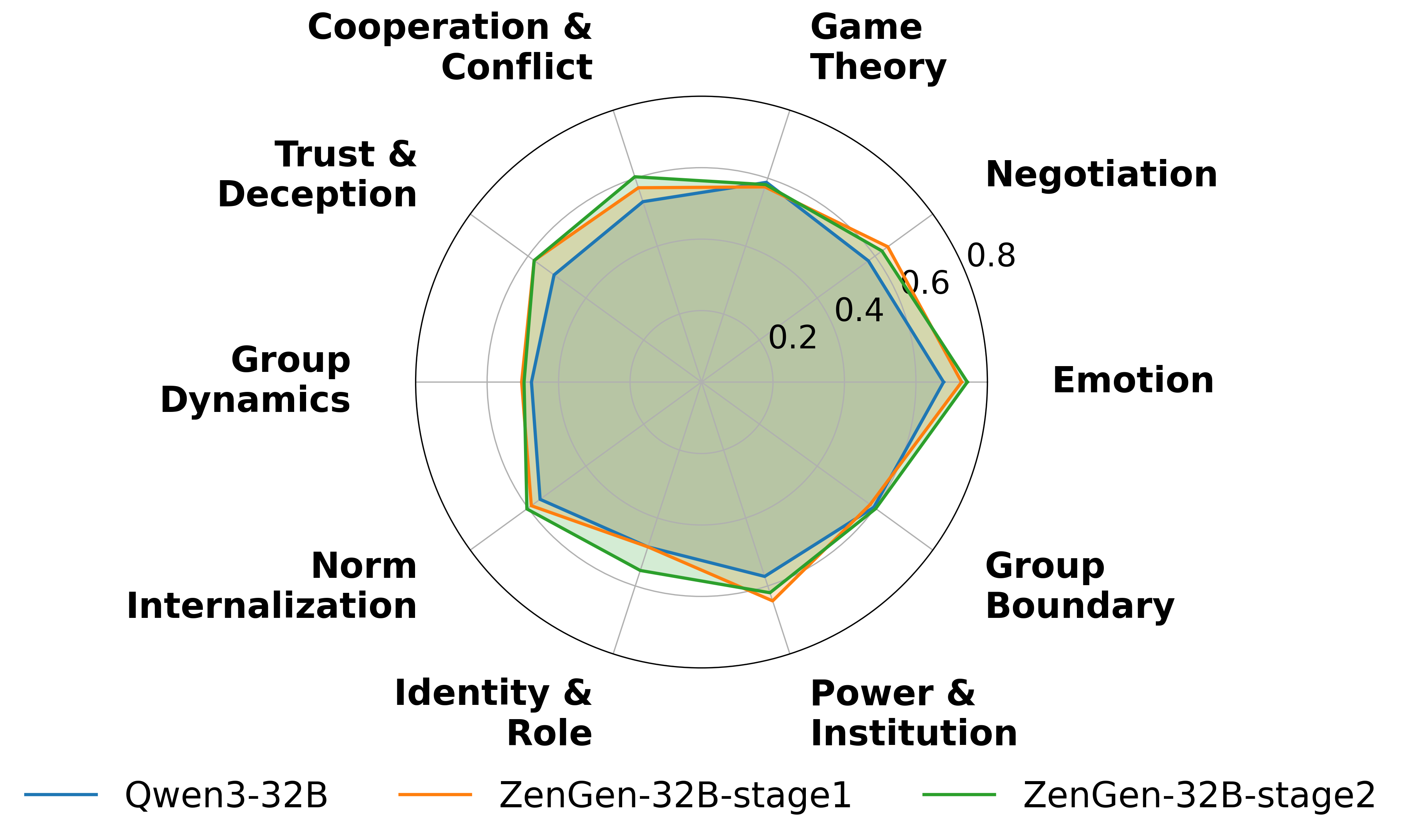}
        \caption{32B-Scale}
    \end{subfigure}
    \begin{subfigure}[t]{0.4\textwidth}
        \centering
        \includegraphics[width=\textwidth]{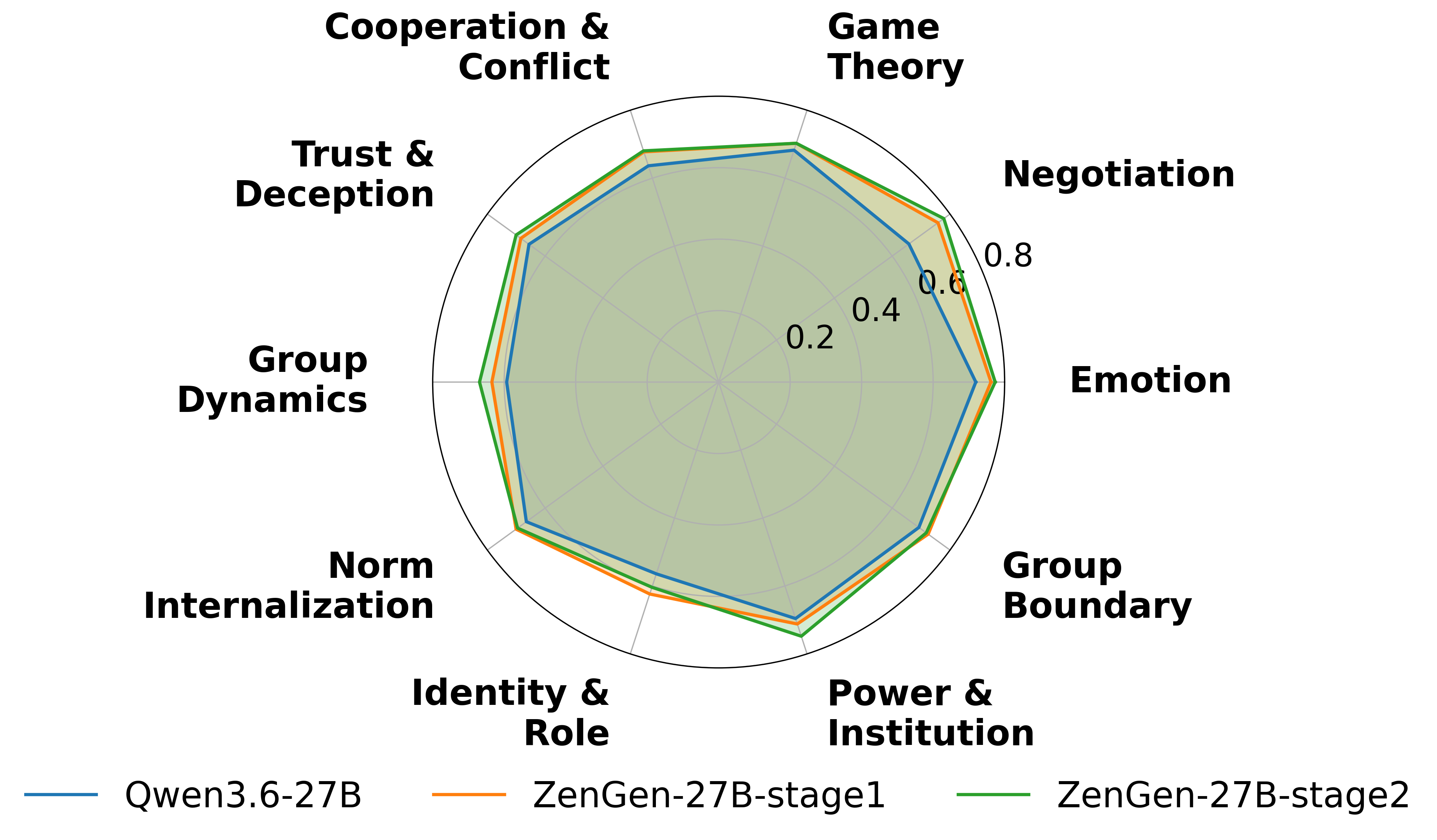}
        \caption{27B-Scale}
    \end{subfigure}

    \caption{Experimental results of the specialized Social Mind ability enhancement in Stage 2.}
    \label{fig:gpteval}
\end{figure*}

\paragraph{Overall Performance} Figure~\ref{fig:gpteval} presents the performance of \modelname{} models across their corresponding base models, Stage 1 and Stage 2 models. Overall, Stage 2 consistently improves specialized social mind capabilities across model scales, with the most substantial gains  in Emotion reasoning. For example, \modelname{}-8B-Stage2 improves Emotion from 61.78\% (Stage 1) to 69.54\% (+7.76 points), while \modelname{}-27B-Stage2 achieves the highest score among all \modelname{} models on Emotion at 77.35\%, surpassing both Stage 1 (76.18\%) and the base model (71.93\%). Notable improvements of Stage 2 are also observed in Norm Internalization for the 8B model (52.79\% to 57.87\%, +5.08 points) and Game Theory Reasoning for the 8B model (50.00\% to 56.76\%, +6.76 points).

In contrast, gains on Negotiation, Trust-Deception, and Group Dynamics are relatively modest, with some dimensions showing performance degradation compared to Stage 1. For instance, \modelname{}-32B-Stage2 drops from 64.43\% to 62.42\% on Negotiation despite substantial Stage 1 gains. This result suggests that Stage 2 primarily enhances capabilities that are directly tractable through targeted supervision, such as emotion understanding and structured social reasoning, while dimensions that require complex strategic reasoning or nuanced social perception remain challenging.

\begin{table*}[ht]
\centering
\caption{Stage 2 performance on fine-grained emotion dimensions.}
\label{tab:stage2_emotion_finegrained}
\setlength{\tabcolsep}{3.5pt}
\resizebox{\textwidth}{!}{
\begin{tabular}{lcccccccccccc}
\toprule
\multirow{2}{*}{Dimension} 
& \multicolumn{3}{c}{8B} 
& \multicolumn{3}{c}{14B} 
& \multicolumn{3}{c}{27B}
& \multicolumn{3}{c}{32B} \\
\cmidrule(lr){2-4} \cmidrule(lr){5-7} \cmidrule(lr){8-10} \cmidrule(lr){11-13}
& Base & Stage 1 & Stage 2 
& Base & Stage 1 & Stage 2 
& Base & Stage 1 & Stage 2 
& Base & Stage 1 & Stage 2 \\
\midrule
Basic Emotion Recognition 
& 55.20\% & 56.86\% & 71.10\% 
& 62.37\% & 63.93\% & 73.70\% 
& 70.17\% & 76.51\% & \textbf{79.42\%} 
& 64.45\% & 73.80\% & 77.65\% \\

Complex Emotion Understanding 
& 50.08\% & 50.25\% & 67.11\% 
& 55.70\% & 57.69\% & 70.08\% 
& 68.93\% & 76.36\% & \textbf{79.34\%} 
& 58.35\% & 69.42\% & 76.03\% \\

Implicit Emotion Inference 
& 53.68\% & 52.48\% & 65.68\% 
& 58.31\% & 62.38\% & 66.12\% 
& 67.00\% & 71.73\% & \textbf{74.70\%} 
& 62.49\% & 68.54\% & 70.19\% \\

Emotion Regulation Strategy Understanding 
& 57.66\% & 63.40\% & 70.07\% 
& 65.69\% & 68.20\% & 75.39\% 
& 70.70\% & 75.39\% & \textbf{79.56\%} 
& 65.38\% & 76.43\% & 78.83\% \\
\bottomrule
\end{tabular}
}
\end{table*}

\paragraph{Fine-grained Emotion Reasoning} To further examine how Stage 2 affects emotion-related social cognition, we analyze performance across four fine-grained emotion sub-dimensions: Basic Emotion Recognition, Complex Emotion Understanding, Implicit Emotion Inference, and Emotion Regulation Strategy Understanding. As shown in Table~\ref{tab:stage2_emotion_finegrained}, the performance gains are most pronounced on Basic Emotion Recognition and Complex Emotion Understanding. \modelname{}-8B-Stage2 improves Basic Emotion Recognition from 56.86\% (Stage 1) to 71.10\% (+14.24 points) and Complex Emotion Understanding from 50.25\% to 67.11\% (+16.86 points). Similar trends are observed for the 14B, 27B, and 32B models, with the 27B model achieving the highest scores across all four emotion sub-dimensions.

These results indicate that Stage 2 is particularly effective at enhancing emotion-related reasoning across multiple levels of complexity. Basic emotion recognition tasks require models to identify overt emotional expressions, while complex emotion understanding and implicit inference demand deeper reasoning about contextual cues, social norms, and underlying psychological states. The substantial gains across all four emotion sub-dimensions suggest that Stage 2 strengthens the model's ability to construct and manipulate emotion-related mental representations, ranging from surface-level recognition to nuanced emotional understanding.

\paragraph{Effect of Model Scale} We also observe that the 8B model benefits more substantially from Stage 2 training on emotion-related tasks, whereas the 27B model achieves higher absolute performance but exhibits more moderate relative improvements. For instance, the 8B model gains over 16 points on Complex Emotion Understanding, while the 27B model improves from 76.36\% to 79.34\% on the same dimension (+2.98 points). This trend aligns with the observation from Stage 1 that smaller models have greater potential for improvement, while larger models may approach saturation on these capabilities.


Overall, the Stage 2 results demonstrate that our specialized training strategy successfully enhances the core components of advanced social mind reasoning, particularly emotion understanding, game-theoretic reasoning, and norm internalization. However, gains on more complex dimensions—including negotiation, trust-deception, group dynamics, and power-institution reasoning—remain limited, indicating that these capabilities may require more sophisticated training strategies, richer interaction data, or longer training horizons. Nevertheless, the consistent improvements across emotion sub-dimensions and the complementary gains relative to Stage 1 validate the effectiveness of our two-stage progression, establishing a solid foundation for future work on comprehensive social intelligence.

\section{\agentname{}: Grounding Social Intelligence at Deployment Time}
\label{sec:actio}

\subsection{Motivation and Challenges}
\label{sec:actio-motivation}

The goal of \agentname{} is to ground social reasoning at deployment time through an agentic runtime. At the problem level, deployed social reasoning is not merely answer generation, but situated judgment management. A valid judgment depends on intermediate interpretations of perspective, mental state, social relation, and norm context. These interpretations may be incomplete, unstable, or mutually inconsistent, and a single forward pass gives little visibility into where the judgment came from or how it should be repaired.

Existing work addresses parts of this problem but does not fully specify the deployment-time process. Task-specific theory-of-mind or social agents sharpen particular abilities, but they are usually tied to narrow task settings~\cite{kim2025thoughttracing, yang2025tomagent, zhang2025autotom}. General ReAct-style agents and broader agent frameworks provide a useful loop for interleaving reasoning and action~\cite{xi2023agentsurvey, park2023generativeagents, wang2024agentsurvey}, but their actions are typically organized around generic search, memory, retrieval, or tool use. For social reasoning, the technical question is therefore not simply how to add tools, but how to regulate the process through which intermediate interpretations are formed, grounded, and revised at deployment time.

This process is difficult because intermediate interpretations in social reasoning are interdependent rather than isolated. An intention attribution can depend on another agent's belief state, a norm-sensitive reading can change which part of the narrative is relevant, and new context can strengthen one interpretation while undermining another. The runtime must therefore manage not only what information is added, but how assumptions are made explicit, checked against the narrative, and revised when they become unstable. A useful runtime should preserve this process in a form that remains observable, revisable, and traceable during inference.

Accordingly, \agentname{} translates this runtime requirement into a set of distinct support roles. It separates deployment-time assistance into procedural guidance for structuring the reasoning path, runtime representation of attributed mental states, additional normative and social context for grounding interpretations, and historical traces for reusing previously effective reasoning patterns. These supports address different failure modes and should not be collapsed into a single memory, retrieval, or tool-use channel. \agentname{} therefore designs corresponding support modules for them. Once multiple supports are available, the remaining challenge is to keep their use selective, ordered, and consistent with the narrative. \agentname{} addresses this challenge with a harness structure that gates when each support is invoked, regulates how its output is integrated, and validates the resulting reasoning trace during inference.

The remainder of this section presents this architecture in detail. Section~\ref{sec:system-overview} describes the harness-controlled inference loop, Sections~\ref{sec:actio-skill}--\ref{sec:actio-rag} introduce the support modules, and Section~\ref{sec:actio-harness} describes how gating, validation, repair, and trace-based optimization bind them into a controllable deployment-time system.

\subsection{System Overview}
\label{sec:system-overview}

\begin{figure}[!tbp]
\centering
\scalebox{1}[0.88]{%
  \includegraphics[width=0.95\linewidth]{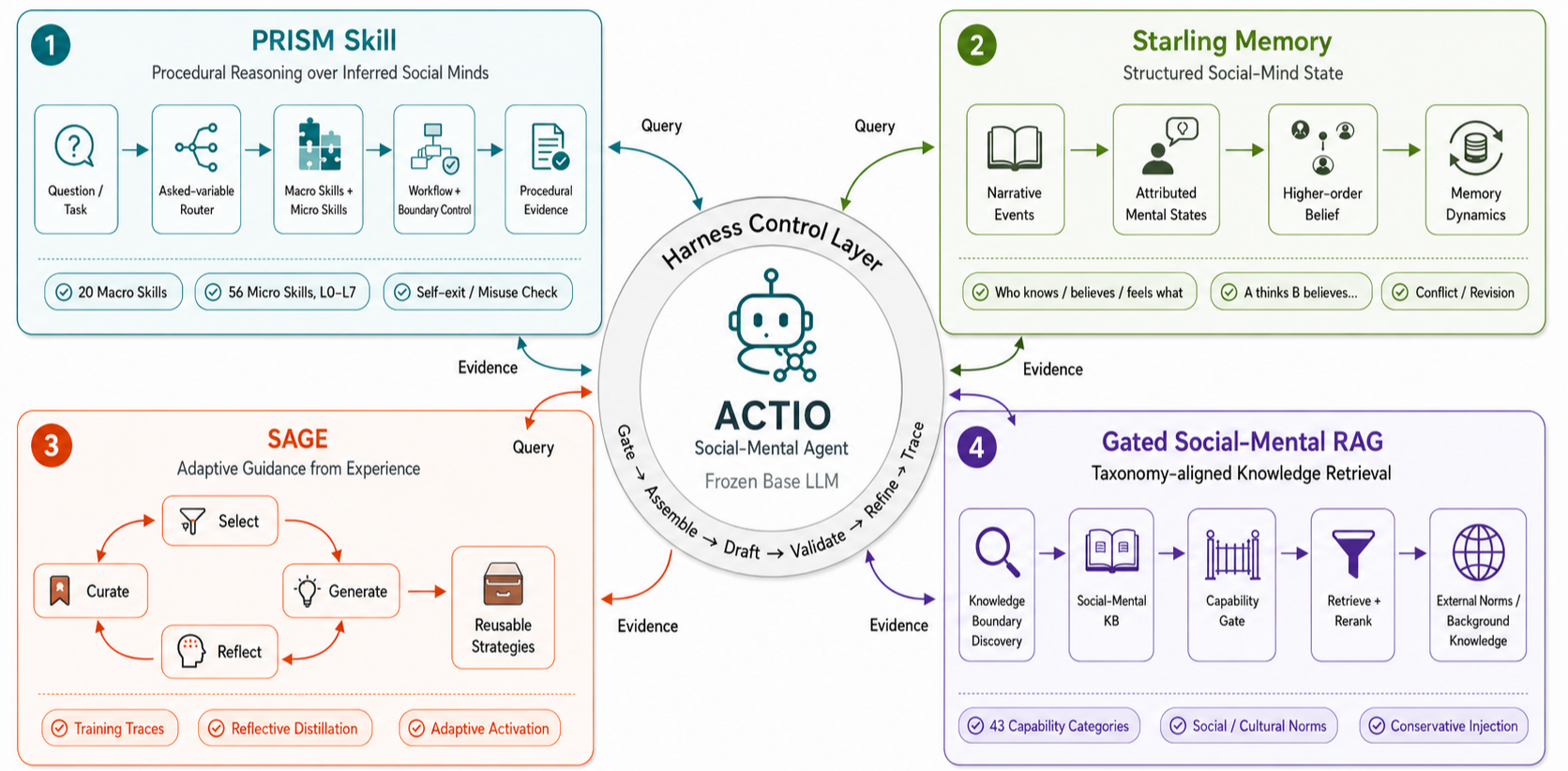}
}
\caption{System overview of \agentname{}. A frozen base LLM is wrapped by a harness;
four support modules, including PRISM, Starling memory, SAGE and gated RAG, are invoked by the control system.}
\label{fig:overview}
\end{figure}

Following the motivation above, \agentname{} implements an observable, revisable, and traceable runtime by wrapping a frozen base language model with a harness-controlled inference pipeline. The base model remains responsible for language understanding and answer generation, while the harness regulates how deployment-time support is selected, integrated, validated, and recorded. As illustrated in Figure~\ref{fig:overview}, for each item, the control layer parses the input, selects the active support modules, assembles their outputs into a typed context, obtains a draft from the base model, applies social-reasoning validation checks, and returns the final answer together with an execution trace.

The architecture consists of a control layer and four support modules. The four support modules divide deployment-time assistance into complementary roles, each contributing a specific capability to the harness-controlled runtime:
  \begin{itemize}
      \item \textbf{PRISM} provides procedural guidance for structuring the reasoning path, enabling \emph{explicit
      capability modeling} by making the required social-reasoning procedure visible rather than leaving it implicit
      in the base model's free-form generation.
      \item \textbf{Starling Memory} provides runtime representations of attributed mental states, enabling
      \emph{runtime state representation} by maintaining beliefs, intentions, knowledge states, and higher-order
      relations outside the frozen model.
      \item \textbf{SAGE} provides historical traces of prior reasoning cases, enabling \emph{experience-based
      generalization} by distilling reusable reasoning strategies from past interactions.
      \item \textbf{Gated Social-Mental RAG} provides additional normative and social context when the narrative alone is
      insufficient, enabling \emph{selective contextual grounding} by introducing external knowledge under
      controlled invocation.
  \end{itemize}
The control layer coordinates these modules through a typed interface that identifies each support's source, target mental variable, and intended use.

The runtime is summarized formally as follows. Given an item $x$, the gate selects the queried mental variable $v$, execution path $p$, and active module set $g$ under the runtime control policy $\bm{\theta}$; the composer then constructs the corresponding context $c_{x,g,p}$:
\begin{equation}
\begin{aligned}
(v,p,g) &= \pi_{\mathrm{gate}}(x;\bm{\theta}), \\
c_{x,g,p} &= \mathrm{Compose}(x,g,p), \\
\hat{y},\,\tau
&= \mathrm{Finalize}\!\left(
    \mathrm{Validate}_{v}\!\left(
        \mathrm{Draft}_{M}(c_{x,g,p})
    \right)
\right).
\end{aligned}
\label{eq:forward}
\end{equation}
Here, $\bm{\theta}$ denotes the runtime control policy, and $\tau$ records the selected path, activated support, and validation outcome. These traces support offline analysis and policy refinement, as detailed in Section~\ref{sec:actio-harness}. This formulation keeps the base model fixed while making the system's runtime decisions explicit and traceable.


\subsection{PRISM: A Hierarchical Mentalizing Skill Library}
\label{sec:actio-skill}

PRISM is the procedural support module in \agentname{}. It turns the reasoning procedure required by a social-reasoning item into an explicit, bounded skill path that can be selected and checked at runtime. This design is grounded in the view that social cognition involves partially distinct operations, such as perception, belief, desire, emotion, pragmatics, and norm reasoning~\cite{schaafsma2015deconstructing,wellman2004scaling,happe2017structure}. In free-form generation, these operations can be conflated: perceptual access may be treated as attention, knowledge as memory, or social harm as faux-pas attribution. PRISM addresses such confusions by routing each item to a procedure aligned with the queried mental or social variable, thereby enabling \emph{explicit capability modeling}.

\begin{figure}[t]
\centering
\includegraphics[width=0.95\linewidth]{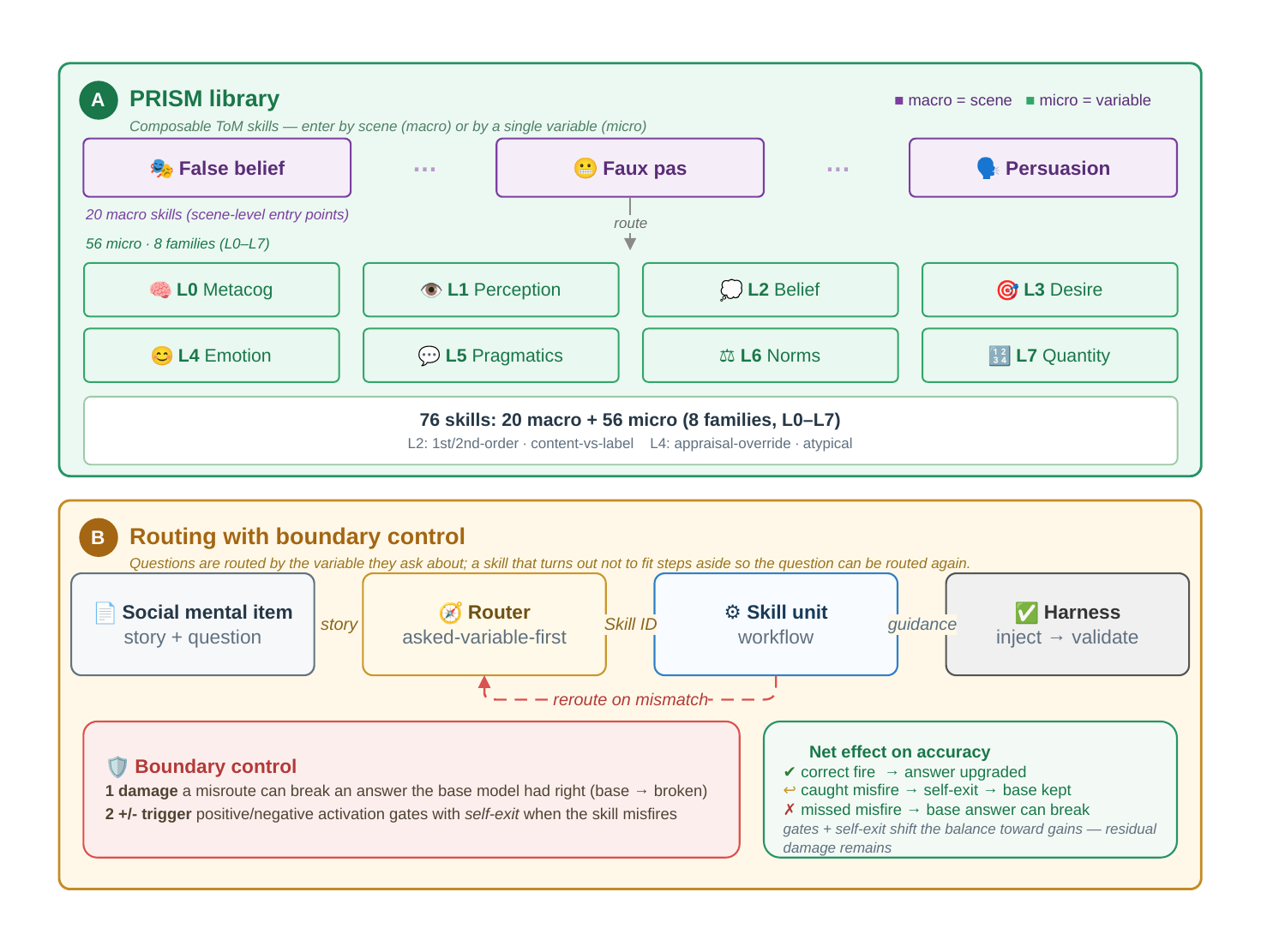}
\caption{Overview of PRISM. \textbf{(A)}~The library is two-level: $20$ macro skills are scene-level entry points (false belief, faux pas, persuasion, \ldots) that route a story to a recognizable scenario; micro skills are atomic capabilities organized into layers L0--L7, deployed as $56$ skills across eight semantic families (one per layer) that separate easily conflated capabilities, for a $76$-skill routing pool. \textbf{(B)}~A dedicated LLM router reads the asked mental variable (not the story surface) and selects one skill unit; each unit carries positive and negative triggers and a self-exit rule, and the dashed edge marks the reroute path taken on mismatch.}
\label{fig:skill}
\end{figure}

A useful procedural support module must satisfy three requirements in the harness-controlled runtime. It must cover recurring social-reasoning procedures while preserving fine-grained capability distinctions; it must accommodate stable persona or preference cues when behavior prediction is underdetermined; and it must keep skill use bounded so that an off-target procedure can be stopped, rerouted, or diagnosed. PRISM meets these requirements through a hierarchical skill library, a persona-aware extension, and routing with boundary control. Figure~\ref{fig:skill} summarizes the module, and the following subsections describe these components in turn.

\subsubsection{Theory-grounded Hierarchical Skill Library}
\label{subsubsec:skill_library}

Social reasoning tasks often combine several reasoning operations, so procedural support must cover recurring scenarios without collapsing the capabilities that make them difficult. Faux-pas reasoning, for example, may require speaker-knowledge tracking, listener-affect inference, and norm judgment, while persuasion involves goals, relationships, and pragmatic strategy. Rather than organizing the library solely by surface scenarios, PRISM separates scene-level routing from capability-level reasoning in a hierarchical skill structure.


\begin{figure}[!tbp]
\centering
\includegraphics[width=\linewidth]{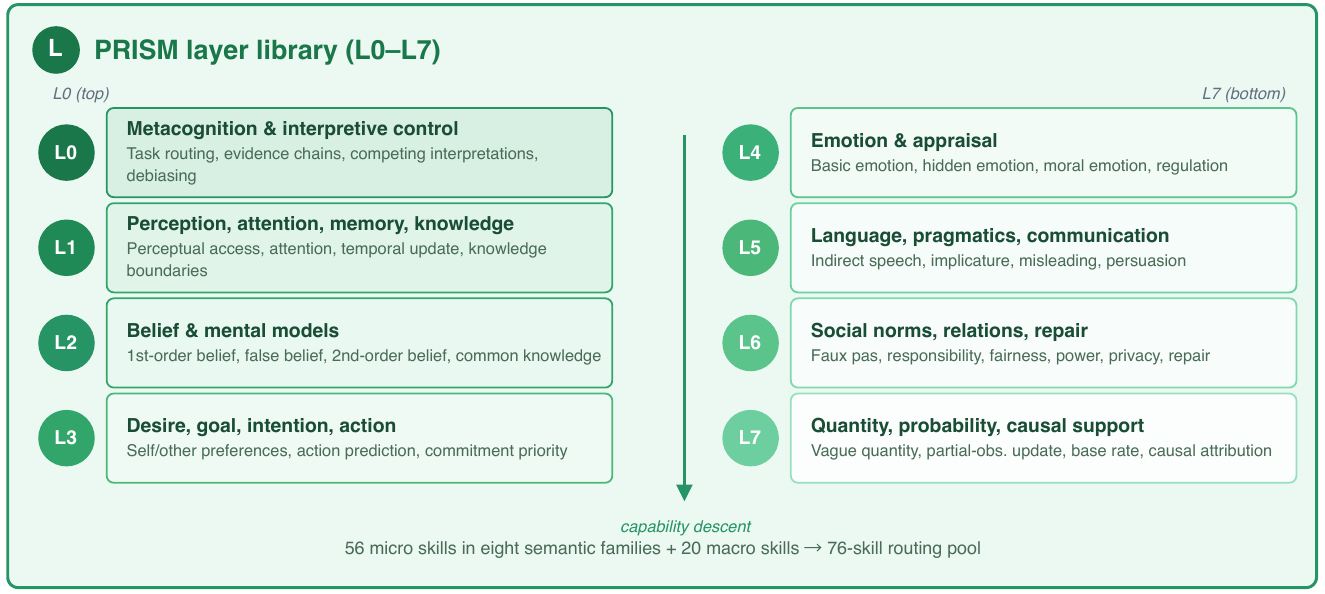}
\caption{PRISM layer library (L0--L7): each row states a layer's core capability and typical decision variables. The deployed library instantiates $56$ micro skills in eight semantic families (one per layer), together with $20$ macro skills forming a $76$-skill routing pool.}
\label{fig:skill_layers}
\end{figure}

The library combines scene-level macro skills with capability-level micro skills. Macro skills provide reusable workflows for recurring scenario structures, while micro skills encode fine-grained reasoning operations organized into layers L0--L7 (Figure~\ref{fig:skill_layers}). The current library contains $20$ macro and $56$ micro skills across eight semantic families, forming a routing pool of $76$ skills.


The hierarchy separates closely related decision variables rather than simply expanding the skill inventory. For example, perceptual access differs from attention, knowledge from memory, and listener harm from faux-pas attribution. At the family level, L2 distinguishes first-order, higher-order, and content-versus-label belief reasoning, while L4 distinguishes appraisal-based, atypical, and moral-emotion inference.


\subsubsection{Persona-aware Skill Extension}
\label{subsubsec:persona_skill}



The core mentalizing skills identify the mental-state and normative constraints relevant to a case, but these constraints may leave multiple behaviors plausible. Persona-aware skills provide a bounded extension for such cases: they use explicit and stable persona or preference cues when the task requires behavior prediction or action selection.

The extension modulates rather than overrides the core procedure. It cannot supersede explicit evidence or infer stable traits from isolated actions or temporary states; when the required persona evidence is absent, the extension is bypassed, and reasoning remains with the core mentalizing skills.


\subsubsection{Skill Routing, Boundary Control, and Outcome Diagnosis}
\label{subsubsec:skill_routing}

Because PRISM enters the harness as procedural support rather than as an independent answerer, skill use must be selective, bounded, and inspectable. The effectiveness of a skill depends on both correct selection and appropriate use. An off-target skill may redirect reasoning toward the wrong mental variable or change an otherwise correct base prediction. PRISM therefore combines asked-variable routing, explicit applicability boundaries, and separate diagnostics for routing quality and downstream answer effects.
When invoked by the harness, PRISM returns a typed support block containing the selected skill identifier, target variable, workflow guidance, boundary conditions, and failure checks. The composer uses this block as procedural guidance for the base model.

Let $\mathcal{S}=\mathcal{S}_{\mathrm{macro}}\cup\mathcal{S}_{\mathrm{micro}}$ denote the PRISM library. Each skill $s\in\mathcal{S}$ is represented as
\[
s=\bigl(\textit{var}(s),\textit{pos}(s),\textit{neg}(s),
\textit{wf}(s),\textit{fc}(s),\textit{exit}(s)\bigr),
\]
where $\textit{var}(s)$ denotes the target mental variable, $\textit{pos}(s)$ and $\textit{neg}(s)$ specify applicability conditions, $\textit{wf}(s)$ defines the reasoning procedure, $\textit{fc}(s)$ checks common failure patterns, and $\textit{exit}(s)$ determines whether execution should stop or reroute.

\paragraph{Asked-variable routing.}
Given an item $x$, the router first identifies the queried mental variable $v$ and then selects a compatible skill:
\begin{equation}
v=\pi_{\mathrm{var}}(x),\qquad
s^{*}=\pi_{\mathrm{router}}(x,\mathcal{S};v).
\label{eq:skill_routing}
\end{equation}
The selection is guided by the question target and the skill's applicability conditions rather than by surface similarity to the story. The same narrative may therefore route differently depending on whether the question asks about factual truth, speaker intent, or listener affect.



\paragraph{Explicit boundary control.}
Each skill includes positive and negative triggers, a workflow, a failure check, and a self-exit rule, as summarized in Table~\ref{tab:skill_unit}. The triggers define when the skill is applicable, while the failure check and self-exit allow execution to stop when the selected procedure no longer matches the item. This prevents a locally useful workflow from being applied beyond its intended conditions.


\begin{table}[t]
\centering
\small
\caption{Components of a PRISM skill unit. Applicability conditions and self-exit jointly bound skill activation.}
\label{tab:skill_unit}
\begin{tabularx}{\linewidth}{>{\raggedright\arraybackslash}p{0.28\linewidth} X}
\toprule
\textbf{Component} & \textbf{Role} \\
\midrule
Target variable & Mental variable addressed by the skill \\
Positive trigger & Conditions supporting activation \\
Negative trigger & Conditions excluding activation \\
Workflow & Structured reasoning procedure \\
Failure check & Detection of common reasoning errors \\
Self-exit / reroute & Stop or reroute when the skill does not match \\
\bottomrule
\end{tabularx}
\end{table}

If the selected skill triggers its self-exit condition, it is removed from the candidate set and the item is rerouted:
\begin{equation}
s_{\mathrm{next}}
=
\pi_{\mathrm{router}}
\bigl(x,\mathcal{S}\setminus\{s^{*}\};v\bigr).
\label{eq:skill_selfexit}
\end{equation}
The process terminates after at most $K$ rerouting attempts, bounding the additional inference cost and preventing an open-ended search over the library.

\paragraph{Routing and outcome diagnosis.}
Routing quality and answer quality are evaluated separately. We use \emph{family hit} and \emph{strict hit} to measure whether the router selects an appropriate capability family and skill, and \emph{repair} and \emph{damage} to measure how skill augmentation changes the base prediction. Let $f^{*}(x)$ denote the annotated capability family and $\mathcal{S}^{*}(x)$ the set of acceptable skills for item $x$. The four metrics are
\begin{align}
\text{family-hit}
&=
\Pr\!\left[\textit{fam}(s^{*})=f^{*}(x)\right],
\notag\\
\text{strict-hit}
&=
\Pr\!\left[s^{*}\in\mathcal{S}^{*}(x)\right],
\notag\\
\text{repair}
&=
\Pr\!\left[
y_{s^{*}}\text{ correct}
\mid
y_{\mathrm{base}}\text{ wrong}
\right],
\notag\\
\text{damage}
&=
\Pr\!\left[
y_{s^{*}}\text{ wrong}
\mid
y_{\mathrm{base}}\text{ correct}
\right].
\label{eq:skill_diagnosis}
\end{align}

The first two metrics diagnose selection errors independently of the final prediction, whereas repair and damage quantify the positive and negative effects of applying the selected skill. This distinction allows a correct answer produced under an incorrect route to be separated from a genuine routing success, and is used in the skill analysis of Section~\ref{subsec:skill_experiment}.

\subsection{Starling Memory: Runtime State Representation for Social Reasoning}
\label{sec:actio-starling}


\emph{Starling Memory} is the runtime state representation module in \agentname{}. It maintains an explicit interaction-level state outside the frozen model. This state records who holds a belief, knowledge state, desire, intention, or commitment, together with the evidence and temporal context. Starling is complementary to SAGE (Section~\ref{sec:actio-memory}): Starling tracks the social state within an ongoing interaction, whereas SAGE stores reusable reasoning strategies distilled across cases. The harness can query Starling when a judgment depends on perspective, information access, temporal change, or higher-order belief, making these intermediate states available as updateable runtime support rather than implicit free-form reasoning.

The design has two parts. An attribution-first representation supports perspective and higher-order belief, while a versioned update mechanism handles temporal change and revision.

\subsubsection{Attribution-first State Representation}


The basic unit of Starling is an attributed \emph{statement} rather than an agent-independent fact. Each statement records the agent who holds the state, the content being held, and the attitude toward that content, such as belief, knowledge, desire, intention, or commitment. It also records polarity, temporal scope, supporting evidence, confidence, and revision metadata. Conflicting or retracted statements are retained as linked versions rather than overwritten. The same content can therefore be represented differently for different agents, and recursive beliefs can be encoded by allowing one statement to refer to another. This schema supports explicit queries over perspective, provenance, and nested mental states.

Starling uses two complementary representations for higher-order belief. \emph{Symbolic nesting} encodes explicitly expressed attitudes about other attitudes. \emph{Perception-grounded reconstruction} instead derives a character's belief from what the character could observe over time, particularly when an unobserved state change leaves the character with a stale belief. For higher-order queries, the system composes the relevant observation histories across agents while preserving the distinction between direct observation and reported information.

\subsubsection{Temporal Update and Revision}
Runtime state must be updateable without losing its provenance. Starling therefore treats updates as versioned changes rather than destructive overwrites.
New observations first enter a revisable store, while records become more stable only after repeated support, replay, or consistency checks. Direct writes to the stable layer are disallowed, reducing the risk that a single extraction error becomes persistent state.

Replay prioritizes records according to salience, novelty, unresolved conflict, and task relevance. Contradictions, corrections, and retractions create linked records that preserve the original evidence and revision path. This mechanism keeps the represented state inspectable while allowing it to change as the interaction evolves.

\subsection{SAGE: Experience Memory and Adaptive Routed Activation}
\label{sec:actio-memory}

\begin{figure}[!tbp]
    \centering
    \includegraphics[width=1.0\linewidth]{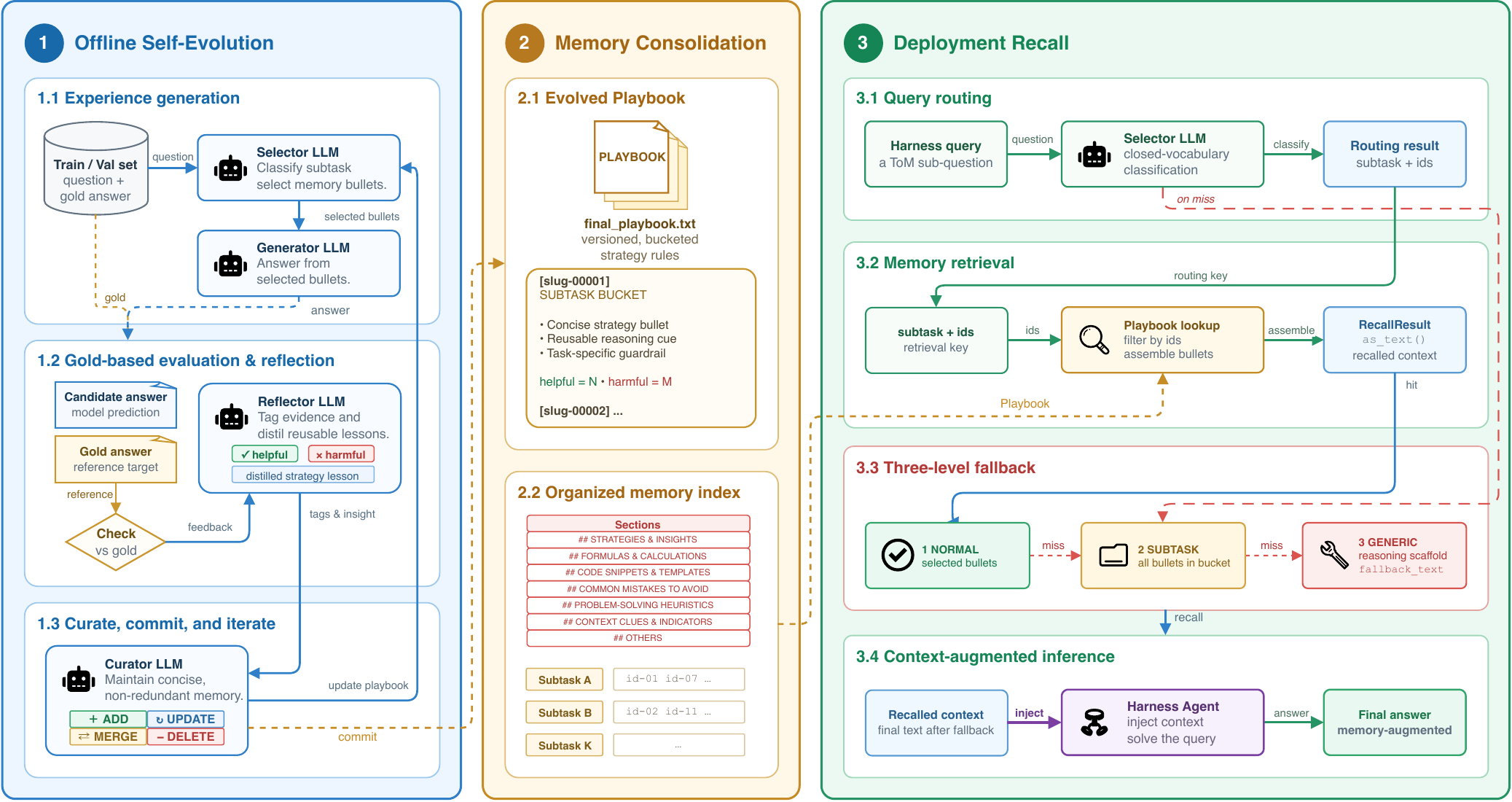}
    \caption{
        Overview of SAGE. The offline loop distills and consolidates reusable
        reasoning strategies, while the deployment-time path selectively recalls
        relevant entries through a bounded fallback cascade.
    }
    \label{fig:memory}
\end{figure}

SAGE is the experience memory module in \agentname{}. Unlike Starling Memory, which represents the social state within an ongoing interaction, SAGE accumulates reusable reasoning experience across cases. Rather than retaining raw interaction histories, SAGE converts informative successes and failures into compact guidance with explicit usage and boundary conditions. This enables \emph{experience-based generalization}: prior reasoning experience can support new scenarios without updating the base model. As shown in Figure~\ref{fig:memory}, SAGE consists of an offline distillation and consolidation loop and a deployment-time selector that recalls a small number of relevant strategies.

\subsubsection{Reflective Strategy Distillation}

SAGE extends the Generator--Reflector--Curator loop of
ACE~\cite{zhang2026agentic} with a dedicated \textbf{Selector}. The Selector decouples memory activation from answer generation, allowing the loop to evaluate strategies under a bounded context rather than conditioning on the whole store. For each
training case paired with a reference outcome, the Selector retrieves a small
subset of existing strategies, and the Generator produces an answer using the
selected guidance. The Reflector compares the answer with the reference,
labels the selected entries as helpful, harmful, or neutral, and may propose a
missing strategy. The Curator then adds, revises, or merges entries in the
store. Only the experience store is updated; the base model remains fixed.

The loop abstracts episode-specific feedback into reusable guidance. For
example, an error caused by assuming that an absent character observed a state
change can be converted into the strategy: ``when a character misses a state
change, preserve the last state that the character observed.'' 
The stored entry includes both the strategy and the conditions under which it should be applied, allowing the same reasoning pattern to be activated for different stories that share the underlying failure mode.

\subsubsection{Experience Entry and Consolidation}

Let $\mathcal{M}=\{m_1,\ldots,m_n\}$ denote the experience store. Each entry is
represented as
\begin{equation}
m=
\bigl(
\mathrm{dim}(m),
\mathrm{str}(m),
\mathrm{use}(m),
\mathrm{bnd}(m),
h^{+}(m),
h^{-}(m)
\bigr),
\qquad
\mathrm{dim}(m)\in\mathcal{D},
\quad
|\mathcal{D}|=7,
\label{eq:mem_entry}
\end{equation}
where $\mathcal{D}$ contains six social-mental dimensions, including \texttt{Emotion},
\texttt{Desire}, \texttt{Intention}, \texttt{Knowledge}, \texttt{Belief}, and
\texttt{Non-Literal Communication}, together with \texttt{Others}.

The field $\mathrm{dim}(m)$ provides a routing label, while
$\mathrm{str}(m)$ stores answer-independent reasoning guidance. The usage and
boundary conditions, $\mathrm{use}(m)$ and $\mathrm{bnd}(m)$, specify when the
strategy should and should not be applied. The counters $h^{+}(m)$ and
$h^{-}(m)$ record how often the entry has helped or harmed predictions during
distillation. Edit provenance is stored separately to track additions,
revisions, and merges.

The Curator periodically consolidates the store by merging semantically
redundant entries, generalizing strategies that are tied too closely to a
single case, and pruning entries that are rarely useful or repeatedly harmful.
A fixed memory budget limits the retained store and prevents accumulated
experience from producing unbounded context cost. 
Retention is guided by the helpful--harmful history of each entry and its redundancy with other entries.
Consolidation therefore
operates on reusable cross-case strategies rather than on the interaction-level
mental states maintained by Starling.

\subsubsection{Selective Activation and Fallback}

At deployment time, the experience store is frozen. Given an item $x$, the
Selector predicts its social-mental dimension and retrieves at most $k$ entries
according to their dimension, usage, boundary conditions, and previous
helpfulness. Only the selected strategies are injected into the reasoning
context; the remaining store is not exposed to the base model. This selective
activation bounds the amount of memory added to each item.

When exact strategy selection returns no entry, SAGE uses a three-level fallback
cascade:
\begin{enumerate}
    \item \textbf{Exact recall.}
    Inject the entries selected for the current item.

    \item \textbf{Dimension fallback.}
    Use entries associated with the predicted social-mental dimension.

    \item \textbf{Generic scaffold.}
    If the dimension bucket is also empty, inject a short, answer-independent
    scaffold that identifies the queried perspective and separates reality from
    each character's mental state.
\end{enumerate}

Let
\[
B(v)=\{m\in\mathcal{M}:\mathrm{dim}(m)=v\}
\]
denote the entries associated with dimension $v$, and let $g_0$ denote the
generic scaffold. If $A(x)$ is the Selector's exact recall result and
$\hat{v}(x)$ is the predicted dimension, the injected guidance is
\begin{equation}
A^{\dagger}(x)=
\begin{cases}
A(x),
& A(x)\neq\varnothing,\\
B(\hat{v}(x)),
& A(x)=\varnothing
  \ \wedge\
  B(\hat{v}(x))\neq\varnothing,\\
\{g_0\},
& \text{otherwise}.
\end{cases}
\label{eq:mem_fallback}
\end{equation}

We define exact-recall coverage as
\[
\mathrm{cov}
=
\Pr_{x}\!\left[A(x)\neq\varnothing\right],
\]
which measures how often the Selector retrieves a strategy without using
fallback guidance. The cascade prevents an empty recall from removing all
support, but it does not guarantee that the fallback is relevant. Recall
coverage and downstream answer effects are therefore evaluated separately.

\subsection{Gated Social-Mental RAG: Taxonomy-Guided Retrieval and Feedback Adaptation}
\label{sec:actio-rag}


Gated RAG is the contextual grounding module in \agentname{}. It supplies external normative and social knowledge when a judgment cannot be grounded from the narrative alone, enabling \emph{selective contextual grounding}. This support is useful for cases that depend on social norms, cultural expectations, role obligations, or pragmatic conventions that are unstated in the item. However, external knowledge must remain subordinate to the narrative: generic norms can override what a character actually knows or feels, retrieval can shift the target from a mental-state judgment to a moral verdict, and surface-similar entries can dilute attention to the queried variable. The module therefore combines taxonomy-guided retrieval with validation-based feedback and is activated only when external knowledge is expected to help.

\subsubsection{Taxonomy-Guided Retrieval}

Retrieval is organized by a social-mind taxonomy
$\mathcal{C}=\{c_1,\ldots,c_{43}\}$ covering belief and knowledge, emotion,
desire and intention, and social pragmatics. The same taxonomy labels both
queries and knowledge entries, providing a shared routing space and making retrieval capability-aligned rather than purely surface-similarity based.

Given an item $x$, the module first predicts its relevant categories
$\hat{L}(x)\subseteq\mathcal{C}$. For each category $c$, let $B(c)=\{e\in\mathcal{K}:c\in L(e)\}$ denote the bucket of entries assigned to that
  category. The candidate pool for item $x$ is
  \[
  P(x)=\bigcup_{c\in\hat{L}(x)}B(c).
  \]
If $\hat{L}(x)=\varnothing$ or $P(x)=\varnothing$, the retrieval path is bypassed and the base reasoner proceeds without external knowledge. Otherwise, the module ranks entries within $P(x)$ and returns a bounded set of entries as contextual support. The returned entries enter the composer as a bounded context block with their taxonomy labels and knowledge types, rather than as answers or decision rules.

\subsubsection{Validation-Guided Credit Assignment and Adaptive Search}

Validation data are used to estimate the contribution of retrieved entries.
Instead of treating the knowledge store as fixed, the RAG module records whether
an entry tends to repair, damage, or leave a prediction unchanged. This feedback
is used to refine both the retained entries and their subsequent retrieval:
\begin{equation}
u(e)=\mathrm{Credit}\!\left(e;\mathcal{V}\right),\qquad
\mathcal{K},\pi_{\mathrm{ret}}
\leftarrow
\mathrm{Refine}\!\left(\mathcal{K},\pi_{\mathrm{ret}};\{u(e)\}\right),
\label{eq:rag_credit}
\end{equation}
where $\mathcal{V}$ is the validation set, $u(e)$ is the empirical contribution
of entry $e$, and $\pi_{\mathrm{ret}}$ denotes the retrieval policy.

Here $\mathcal{V}$ is a held-out set of labeled items, disjoint from the
evaluation data, that serves as the ground-truth reference for judging whether
retrieval actually helps: because each item carries a known answer, the module
can score an entry by its effect on prediction correctness rather than on
surface plausibility, and keeping $\mathcal{V}$ separate from the final test data
prevents the feedback signal from reusing the items on which the system is
scored. The contribution $u(e)$ is then estimated by contrasting the validation
outcomes in which $e$ participates in retrieval against the outcomes expected
without it: an entry associated with corrected predictions accrues positive
credit, one associated with newly introduced errors accrues negative credit, and
cases with no effect on correctness are neutral, so that aggregating over
$\mathcal{V}$ gives a single estimate of how much the entry helps or hurts on
average. Finally, $\mathrm{Refine}(\cdot)$ turns these credit signals into two
coupled updates driven by the same $\{u(e)\}$. On the knowledge store
$\mathcal{K}$ it retains consistently useful entries, flags marginal ones for
revision, and suppresses those with negative net utility. On the retrieval
policy $\pi_{\mathrm{ret}}$ it reweights how entries are ranked within the
candidate pool and how much each taxonomy category is trusted during that
ranking, so that future retrieval favors entries and categories with validated
positive utility and down-ranks those that have proven harmful; categories whose
entries carry no retained utility are thereby skipped without a separate rule.

The harness decides whether RAG should be invoked, while the RAG module uses
taxonomy alignment and these validation outcomes to determine where to search and
which entries to return. External knowledge thus supplements, rather than
replaces, the narrative evidence used for social-mental reasoning.

\subsection{Agent Harness}
\label{sec:actio-harness}

The support modules in Sections~\ref{sec:actio-skill}--\ref{sec:actio-rag} provide procedural, state, experiential, and contextual support, but their usefulness depends on when and how they are introduced. We summarize the main control risks as two failure modes: 
\begin{itemize}
  \item \emph{Mismatch}: a piece of evidence is applied to a task or background it does not fit.
  \item \emph{Conflict}: two pieces of evidence push the draft toward contradictory conclusions.
\end{itemize}
\noindent To address these issues, we introduce the harness as the control layer. This layer governs how evidence is gated, assembled, and validated, keeping support integration selective, ordered, and narrative-consistent.

\begin{figure}[!tbp]
\centering
\definecolor{cHarnessRun}{HTML}{2C5F8A}
\definecolor{cHarnessOff}{HTML}{B8860B}
\definecolor{cHarnessSup}{HTML}{555555}
\resizebox{0.98\linewidth}{!}{%
\begin{tikzpicture}[
  font=\small\sffamily,
  >={Stealth[length=2.5mm, width=1.5mm]},
  online/.style={
    rectangle, rounded corners=3pt, draw=cHarnessRun, line width=0.9pt,
    fill=cHarnessRun!7, align=center, text width=18mm,
    minimum height=13mm, inner xsep=3pt, inner ysep=3pt},
  offline/.style={
    rectangle, rounded corners=3pt, draw=cHarnessOff, line width=1.0pt,
    fill=cHarnessOff!9, align=center, text width=26mm,
    minimum height=13mm, inner xsep=3pt, inner ysep=3pt},
  supp/.style={
    rectangle, rounded corners=3pt, draw=cHarnessSup, line width=0.9pt,
    fill=gray!8, align=center, text width=22mm,
    minimum height=11mm, inner xsep=3pt, inner ysep=3pt},
  flow/.style={->, line width=0.9pt, draw=cHarnessRun},
  oflow/.style={->, line width=0.9pt, draw=cHarnessOff},
  fbedge/.style={->, line width=0.9pt, draw=cHarnessOff, dashed},
  supwire/.style={line width=0.8pt, draw=cHarnessSup!70},
  supedge/.style={->, line width=0.9pt, draw=cHarnessSup},
  lbl/.style={font=\scriptsize\itshape, text=gray!55!black,
              inner xsep=1.5pt, inner ysep=1pt},
  paneltitle/.style={font=\scriptsize\bfseries, fill=white,
                     inner xsep=2pt, inner ysep=1pt}
]
\node[online] (item)   at (0,    0) {\textbf{ToM item}\\{\scriptsize story+question}};
\node[online] (gate)   at (2.4,  0) {\textbf{Gate}\\{\scriptsize path, modules}};
\node[online] (assem)  at (4.8,  0) {\textbf{Context}\\{\scriptsize assembly}};
\node[online] (draft)  at (7.2,  0) {\textbf{Base-LLM}\\{\scriptsize draft}};
\node[online] (valid)  at (9.6,  0) {\textbf{ToM}\\{\scriptsize validation}};
\node[online] (repair) at (12.0, 0) {\textbf{Repair \&}\\{\scriptsize finalize}};
\node[online] (ans)    at (14.4, 0) {\textbf{Answer}};

\draw[flow] (item)   -- (gate);
\draw[flow] (gate)   -- (assem);
\draw[flow] (assem)  -- (draft);
\draw[flow] (draft)  -- (valid);
\draw[flow] (valid)  -- (repair);
\draw[flow] (repair) -- (ans);

\node[offline] (bundle)  at (2.4,  3.2) {\textbf{Policy Bundle}\\{\scriptsize $\bm{\theta}$}};
\node[offline] (optnode) at (8.4,  3.2) {\textbf{Offline Optimizer}\\{\scriptsize credit assignment}};
\node[offline] (trace)   at (14.4, 3.2) {\textbf{Trace Store}\\{\scriptsize execution trace $\tau$}};

\draw[oflow] (trace)   -- (optnode);
\draw[oflow] (optnode) -- (bundle);
\draw[fbedge] (ans.north)    -- (trace.south)
  node[midway, right=2pt, lbl] {trace $\tau$};
\draw[fbedge] (bundle.south) -- (gate.north)
  node[midway, left=2pt, lbl] {gating policy};

\node[supp] (skill)    at (3.6,  -3.5) {\textbf{PRISM}};
\node[supp] (starling) at (6.0,  -3.5) {\textbf{Starling}\\{\scriptsize Memory}};
\node[supp] (sage)     at (8.4,  -3.5) {\textbf{SAGE}};
\node[supp] (rag)      at (10.8, -3.5) {\textbf{Gated RAG}};

\coordinate (bus) at ($(assem.south) + (0,-2.0)$);
\draw[supwire] (skill.north)    -- (skill.north |- bus);
\draw[supwire] (starling.north) -- (starling.north |- bus);
\draw[supwire] (sage.north)     -- (sage.north |- bus);
\draw[supwire] (rag.north)      -- (rag.north |- bus);
\draw[supwire] (skill.north |- bus) -- (rag.north |- bus);
\draw[supedge] (bus) -- (assem.south);

\begin{scope}[on background layer]
  \node[draw=cHarnessRun, line width=1pt, rounded corners=5pt, fill=cHarnessRun!3,
        inner xsep=4mm, inner ysep=4mm, fit=(item)(ans)] (boxOnline) {};
  \node[draw=cHarnessOff, line width=1pt, rounded corners=5pt, fill=cHarnessOff!3,
        inner xsep=4mm, inner ysep=4mm, fit=(bundle)(optnode)(trace)] (boxOffline) {};
  \node[draw=cHarnessSup, line width=1pt, rounded corners=5pt, fill=gray!3,
        inner xsep=4mm, inner ysep=4mm, fit=(skill)(starling)(sage)(rag)] (boxSupp) {};
\end{scope}
\node[paneltitle, text=cHarnessRun, anchor=west]
  at ([xshift=3mm]boxOnline.north west) {Online runtime (inner loop)};
\node[paneltitle, text=cHarnessOff, anchor=west]
  at ([xshift=3mm]boxOffline.north west) {Offline optimizer (outer loop)};
\node[paneltitle, text=cHarnessSup, anchor=west]
  at ([xshift=3mm]boxSupp.north west) {Support modules invoked by control system};
\end{tikzpicture}%
}
\caption{Overall architecture of the \agentname{} harness: an online runtime (inner loop) gates, assembles, drafts, validates, and finalizes per item; an offline optimizer (outer loop) turns execution traces into a policy bundle that the runtime loads next. Four support modules, including PRISM, SAGE, Starling memory, and gated RAG, are invoked by the control system.}
\label{fig:harness}
\end{figure}
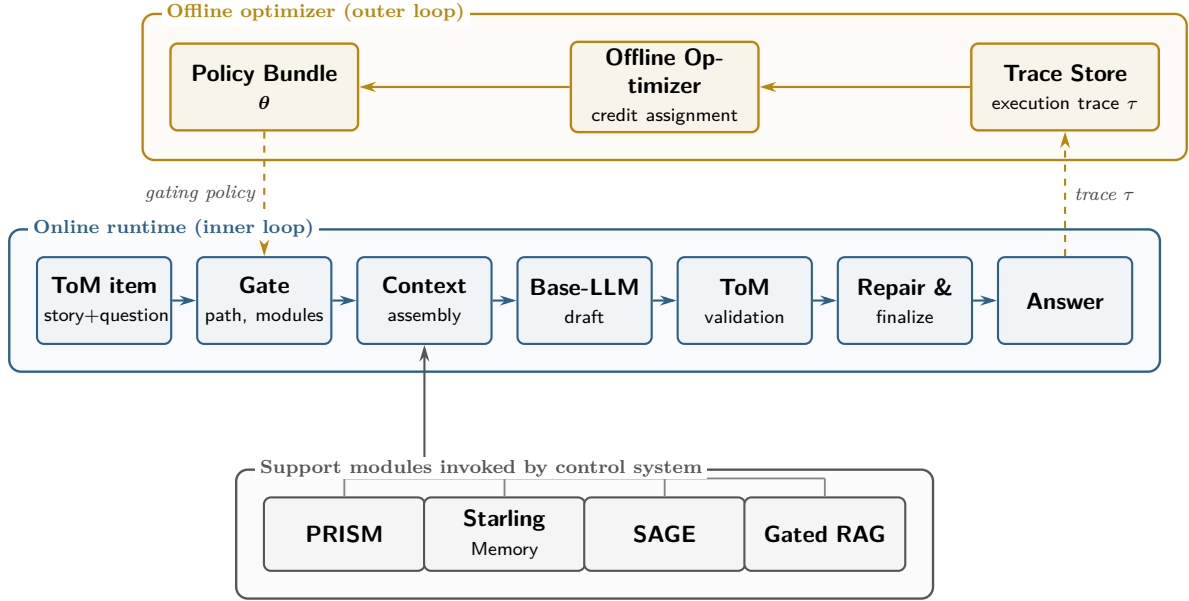

\paragraph{Design: Two Coupled Loops.} The harness is organized as two coupled loops (Figure~\ref{fig:harness}): the \emph{online runtime} (inner loop) that builds, per item, a mapping from a social situation to its corresponding answer, and the \emph{offline optimizer} (outer loop) that turns accumulated traces into the policies the runtime loads next. The two loops communicate only through a trace store (runtime to optimizer) and a policy bundle (optimizer to runtime); the base model's weights stay frozen. This separation keeps per-item behavior reproducible and auditable while still allowing the system to improve over time.

\paragraph{Online Runtime.} The online runtime is the loop that loads the current policy $\bm{\theta}$ and executes a controlled forward pass without learning. It consists of three stages: 

\begin{itemize}
  \item \emph{Gating} analyzes the item to infer the queried mental variable and the information the model lacks, selectively engages the modules whose evidence fills that gap, and chooses one of three execution paths, including direct reasoning, selective augmentation, and staged reasoning.
  \item \emph{Context assembly} organizes the narrative, the queried perspective, and any gated module evidence into one context; narrative evidence is the basis of social-mind perception, and every other source enters as support constrained by the narrative evidence.
  \item \emph{Validation} detects the social reasoning failure modes the harness targets, including perspective drift, norm overriding, and over-mentalization. When a violation is detected, the runtime either repairs the draft under bounded constraints
  or reroutes the item locally.
\end{itemize}
Each run emits a trace $\tau$ for the offline optimizer.

\paragraph{Offline Optimizer.} The optimizer manages the trace store and updates the policy bundle $\bm{\theta}$ from it. It flags a trace as faulty in two cases: the final answer is wrong, or the answer is right, but the trace contains internal errors such as a validator flag, an unjustified gate, or a detected mismatch or conflict. In this paper, $\phi(\tau)$ denotes trace-level loss, where a larger $\phi(\tau)$ indicates weaker correctness and consistency of the trace–answer pair. The update adjusts $\bm{\theta}$ to penalize the gating and assembly decisions that produced faulty traces; $\bm{\theta}$ is a token-level updatable control configuration, including a weighting over gates, execution paths, and assembly rules. Concretely, each faulty trace is attributed to the gate and path decisions that caused it, and the corresponding weights are demoted: gates that opened wrongly are tightened, and paths that produced errors are demoted. The objective prioritizes breaking error-producing strategies over bridging capability gaps:
\begin{equation}
\bm{\theta}_{t+1}=\arg\min_{\bm{\theta}}\; \mathbf{E}_{\tau\sim\pi_{\bm{\theta}}}\bigl[\phi(\tau)\bigr],\quad \phi_{\text{wrong-answer}}\gg\phi_{\text{trace-only}}.
\end{equation}

\subsection{Evaluation of \agentname{}}
\label{sec:actio-results}


\subsubsection{Experimental Setup}
\label{subsec:exp-setup}

\paragraph{Benchmarks}
We evaluate \agentname{} on three public theory-of-mind and social-reasoning
benchmarks that together span the competences the three challenges target.
\emph{ToMBench} covers a broad set of theory-of-mind subtasks, including false belief,
faux pas, emotion, desire, intention, and knowledge, and serves as our primary
multi-capability testbed~\cite{tombench}. \emph{HiToM} nests false-belief
reasoning from the zeroth to the fourth order and isolates higher-order
recursive belief~\cite{hitom}. 
\emph{TactfulToM}~\cite{emnlp/LiuPHS25} probes tact, white lies, and
politeness, and provides a norm-sensitive test of social-pragmatic reasoning. The three range from
perceptual belief tracking to norm-sensitive social judgment, so that no single
competence dominates the evaluation.

\paragraph{Models}
The base models evaluated 
are five instruction models without parameter updates drawn from three families (Qwen, DeepSeek, and GPT), divided by parameter scale into two tiers: a mid-scale tier including GPT-5.4-mini and Qwen3.6-27B, and a large-scale tier including DeepSeek-V4-Pro, Qwen3.7-Plus, and GPT-5.5.
The main results
tables (Section~\ref{subsec:main-results}) report one table per benchmark: each
row is a base model, grouped by scale tier, and the columns compare the frozen
base model, each support module in isolation (PRISM, memory and RAG), and the full
harness combining the three. 

\paragraph{Protocol}
All models run under a single direct-answer protocol: 
the token budget is fixed at 32k, thinking mode is disabled, temperature is set to zero, and each system produces one final prediction per item. Final answers are scored by each
benchmark's own protocol. This keeps the comparison between base model, isolated
module, and full harness attributable to the support rather than to a change in
decoding or sampling.

\paragraph{Evaluation Questions}
We test the harness at the system level first, then examine each challenge's
dedicated support in turn. The four questions map one-to-one onto the four
subsequent subsections:

\begin{itemize}
  \item \textbf{EQ1 (System-level gain; Section~\ref{subsec:main-results}).}
  Does adding deployment-time support improve a frozen base model across
  benchmarks, and how do the gains vary across base models?
  \item \textbf{EQ2 (Capability structuring;
  Section~\ref{subsec:skill_experiment}).} Does PRISM together with its router
  address capability conflation, and where does it help
  rather than harm?
  \item \textbf{EQ3 (Mental-state representation;
  Section~\ref{subsec:starling_experiment}).} Does Starling improve higher-order belief tracking, and under which
  conditions does explicit state representation help?
  \item \textbf{EQ4 (Experience accumulation;
  Section~\ref{subsec:memory_is_effective_but_needs_selection}).} Does SAGE provide reusable reasoning guidance across cases, and how do selective activation and fallback affect its gains? 
\end{itemize}

\noindent
Each subsequent subsection answers one question and states both the supporting
evidence and the boundary conditions under which the support fails.

\subsubsection{Overall System-Level Performance}
\label{subsec:main-results}

Table~\ref{tab:actio-main} compares the base models, isolated support
configurations, and the full harness across three benchmarks and five base
models. We focus here on the overall effect of the full harness; module-specific
effects are analyzed in the following subsections.

\paragraph{Deployment-time support improves most model--benchmark pairs.}
The full harness improves 14 of the 15 model--benchmark pairs, with an average
gain of $3.70$ percentage points (pp). The mean improvements are $3.33$ pp on
ToMBench, $2.79$ pp on HiToM, and $4.97$ pp on TactfulToM. The only decrease
occurs for Qwen3.7-Plus on HiToM, where accuracy changes from $75.00$ to
$74.25$. These results show that deployment-time support provides broad, though
not universal, gains across the evaluated tasks.

\paragraph{The gains vary substantially across base models.}
Averaged over the three benchmarks, the full-harness gains are $5.50$ pp for
Qwen3.6-27B, $4.74$ pp for GPT-5.4-mini, $4.41$ pp for DeepSeek-V4-Pro,
$2.34$ pp for GPT-5.5, and $1.50$ pp for Qwen3.7-Plus. The variation is not
monotonic in base accuracy: models with similar base performance can benefit
differently depending on the benchmark. The largest individual gain is
$9.87$ pp for Qwen3.6-27B on TactfulToM, while several high-base conditions
show smaller improvements.

\paragraph{Harness effectiveness arises from selective gating and complementary module coverage.}
The broad improvement stems from the control layer gating module activation
per item rather than applying a fixed scaffold: the gate reads the asked
mental variable and evidence gap, selects an active module set, and the
composer assembles their typed evidence before the validator filters the
draft. This selectivity lets the full harness recover cases where an isolated
module hurts---on HiToM, PRISM alone reduces Qwen3.6-27B by $0.45$\,pp and
Qwen3.7-Plus by $0.83$\,pp, whereas the full stack gains $2.89$\,pp on the
former and limits the drop to $0.75$\,pp on the latter. The four modules
target distinct base-model failure modes: PRISM makes implicit mentalizing
procedures explicit and yields its largest single-module gains on the
norm-sensitive TactfulToM ($+8.61$\,pp for Qwen3.6-27B, $+6.63$\,pp for
DeepSeek-V4-Pro); Starling and SAGE jointly track attributed state and reuse
distilled strategies, lifting GPT-5.4-mini by $5.80$\,pp and GPT-5.5 by
$3.00$\,pp on ToMBench; and Gated RAG supplies external social-cultural
knowledge for high-base conditions (Qwen3.7-Plus $+3.24$\,pp on TactfulToM,
GPT-5.5 $+0.99$\,pp on HiToM). This complementarity among procedural,
state-tracking, strategy-reuse, and knowledge gaps is what makes the
supports broadly useful but not uniformly additive. The full harness is best or tied for best in 8 of the 15 pairs, while an isolated module performs better in the remaining cases. This indicates that the supports are broadly useful but not uniformly additive, motivating the module-level and routing analyses that follow.

\begin{table}[t]
\centering
\small
\setlength{\tabcolsep}{5pt}
\caption{Main accuracy (\%) across three benchmarks under the direct-answer
protocol. Each benchmark forms a block of rows comparing the base model,
isolated support configurations, and the full harness across five base models.
For each evaluation, the best result is shown in \textbf{bold}, and
results below the corresponding base model are \underline{underlined}.}
\label{tab:actio-main}
\begin{tabular}{l l ccccc}
\toprule
Benchmark & Config & GPT-5.4-mini & Qwen3.6-27B & DeepSeek-V4-Pro & Qwen3.7-Plus & GPT-5.5 \\
\midrule
           & Base   & 73.64 & 79.20 & 78.53 & 81.08 & 83.25 \\
           & PRISM  & 74.27 & 81.89 & 79.41 & 81.19 & 84.85 \\
  ToMBench & RAG    & 74.37 & 81.43 & 79.23 & 82.66 & 84.34 \\
           & Memory & 79.44 & 79.69 & \textbf{80.17} & 81.15 & \textbf{86.25} \\
           & Full   & \textbf{79.97} & \textbf{82.94} & 79.97 & \textbf{83.85} & 85.63 \\
\midrule
           & Base   & 54.58 & 67.28 & 65.17 & 75.00 & 69.41 \\
           & PRISM  & 60.08 & \underline{66.83} & 66.33 & \underline{74.17} & 69.74 \\
  HiToM    & RAG    & 56.25 & 69.12 & 67.50 & 76.08 & \textbf{70.40} \\
           & Memory & 59.58 & 69.83 & 67.17 & \textbf{76.25} & 70.33 \\
           & Full   & \textbf{60.58} & \textbf{70.17} & \textbf{70.33} & \underline{74.25} & 70.05 \\
\midrule
             & Base   & 46.78 & 60.39 & 57.92 & 78.94 & 74.58 \\
             & PRISM  & 49.01 & 69.00 & \textbf{64.55} & 81.52 & 76.95 \\
  TactfulToM & RAG    & 47.84 & 61.97 & 60.51 & \textbf{82.18} & 77.07 \\
             & Memory & \textbf{49.11} & 65.28 & 62.46 & \underline{78.73} & \textbf{79.81} \\
             & Full   & 48.66 & \textbf{70.26} & \textbf{64.55} & 81.41 & 78.59 \\
\bottomrule
\end{tabular}
\end{table}

\subsubsection{Capability Structuring: PRISM Routing and Effect Analysis}
\label{subsec:skill_experiment}

We evaluate whether PRISM identifies the capability queried by an item and
whether the selected skill improves the final prediction. Routing accuracy is
measured on a stratified 320-item ToMBench subset, while the task-level and
router-configuration diagnostics use Qwen3.5-27B as a separate solver.

\paragraph{PRISM provides reliable capability separation at the family level.}
The router achieves a $94.4\%$ family-level hit rate, indicating that it
usually identifies the relevant mental variable. The strict hit rate, which
requires the selected skill to match the annotated acceptable skill set, is
$74.4\%$. This $20.0$\,pp gap suggests that the remaining routing errors are
concentrated in fine-grained selection among related skills, rather than in
confusion between broad capability families.

\begin{figure}[!tbp]
\centering
\includegraphics[width=0.95\linewidth]{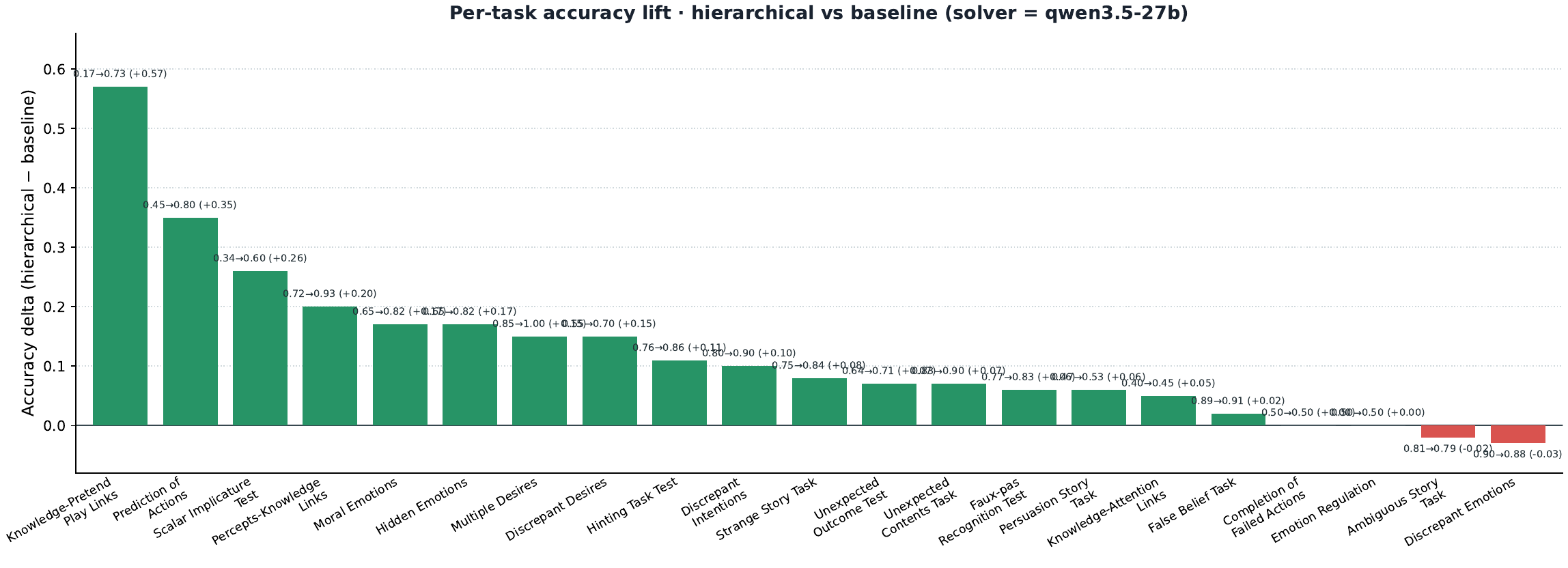}
\caption{Task-level accuracy changes produced by PRISM on ToMBench, using
Qwen3.5-27B as the diagnostic solver. Positive and negative values denote
improvement and damage relative to the base condition.}
\label{fig:skill_lift}
\end{figure}

\paragraph{Skill alignment determines downstream gain and damage.}
PRISM yields its largest gains when the queried variable is supported by a
well-matched reasoning procedure. On TactfulToM, it improves Qwen3.6-27B by
$8.61$\,pp and DeepSeek-V4-Pro by $6.63$\,pp; on ToMBench, it improves
Qwen3.6-27B by $2.69$\,pp. The task-level analysis in
Figure~\ref{fig:skill_lift} further shows that these gains vary across
capabilities.

PRISM is not uniformly beneficial. On HiToM, it reduces accuracy by
$0.45$\,pp for Qwen3.6-27B and by $0.83$\,pp for Qwen3.7-Plus. Such cases
indicate that an unnecessary or misaligned procedure can interfere with tasks
that rely primarily on recursive state tracking or are already handled well by
the base model.

\begin{figure}[t]
\centering
\includegraphics[width=0.9\linewidth]{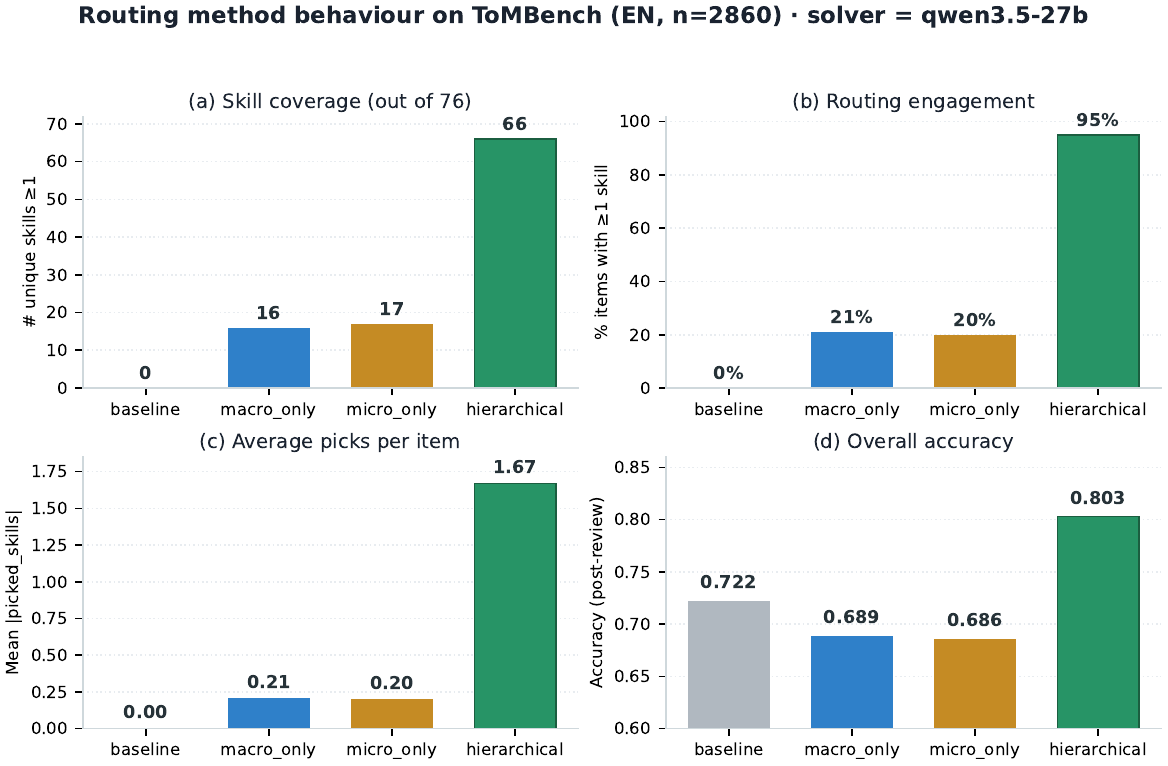}
\caption{Routing diagnostics on ToMBench using Qwen3.5-27B
($n=2860$). The panels report skill-library coverage, item-level activation,
the mean number of selected skills, and post-review accuracy across routing
configurations.}
\label{fig:skill_routing}
\end{figure}

\paragraph{Fine-grained routing remains the main source of unrealized gain.}
The Qwen3.5-27B diagnostics in Figure~\ref{fig:skill_routing} show that
hierarchical routing covers more of the skill library and achieves higher
post-review accuracy than flat macro-only or micro-only routing. The remaining
gap to oracle selection indicates further room to improve item-level skill
choice.

Overall, PRISM reduces broad capability conflation and improves reasoning when
the selected procedure aligns with the queried mental variable. Its remaining
failures mainly arise from fine-grained skill confusion and unnecessary
activation, motivating the boundary controls in
Section~\ref{subsubsec:skill_routing}.

\subsubsection{Mental-State Representation: Starling and Higher-Order Belief}
\label{subsec:starling_experiment}

We evaluate whether Starling improves higher-order belief tracking and identify
the conditions under which explicit state representation is useful. HiToM
contains false-belief questions from zeroth to fourth order, allowing the effect
of representation depth to be examined directly. For this diagnostic
experiment, we use \modelname{}-14B and DeepSeek-V4-Pro under a
\emph{same-model-in-the-loop} protocol: the same model extracts the structured
mental state and produces the final answer, with Starling providing an
intermediate perception-grounded representation.

\begin{figure}[!tbp]
\centering
\includegraphics[width=0.95\linewidth]{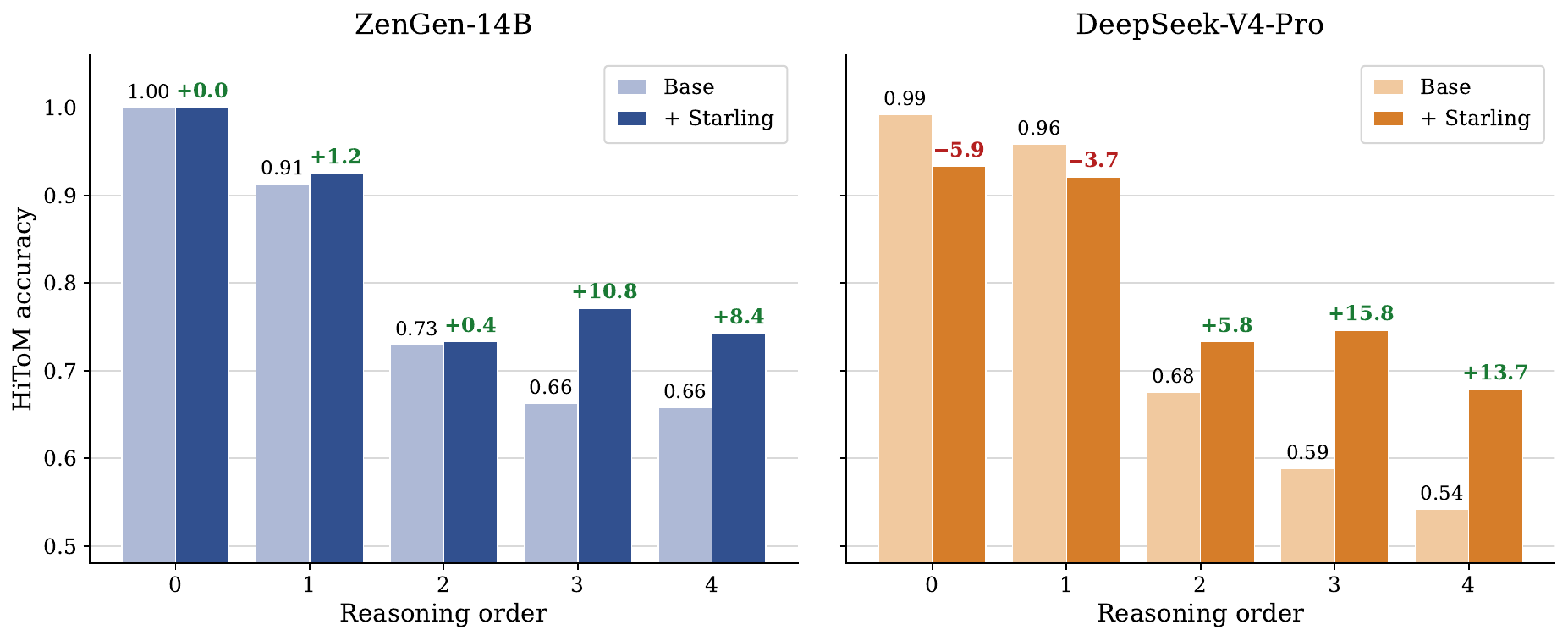}
\caption{HiToM accuracy by reasoning order under the same-model-in-the-loop
protocol. Bars compare the base condition with Starling for \modelname{}-14B and
DeepSeek-V4-Pro; labels report the accuracy change in percentage points.}
\label{fig:starling_gain}
\end{figure}

\paragraph{Starling primarily improves deep higher-order belief tracking.}
Starling improves overall HiToM accuracy from $79.3\%$ to $83.4\%$ on
\modelname{}-14B and from $75.1\%$ to $80.3\%$ on DeepSeek-V4-Pro. As shown in
Figure~\ref{fig:starling_gain}, the gains are concentrated at the third and
fourth orders: \modelname{}-14B improves by $10.8$ and $8.4$\,pp, while
DeepSeek-V4-Pro improves by $15.8$ and $13.7$\,pp. In contrast, zeroth- and
first-order questions show little benefit or a small decline. The results
indicate that explicit tracking becomes most useful when the answer depends on
multiple levels of perceptual access and belief attribution.

\paragraph{The benefit depends on task depth and representation quality.}
Starling provides no clear aggregate gain on the broader ToMBench evaluation,
where many items require only shallow belief, desire, or intention inference.
Its advantage is therefore concentrated in cases where explicit temporal and
perspective tracking is more reliable than direct model reasoning. Conversely,
when the mental-state extractor is unreliable, the structured representation
can propagate incorrect states and reduce final accuracy. This effect is
especially visible on out-of-distribution narratives, whereas semantic gating
reduces the worst observed loss by bypassing Starling when its representation
is unlikely to help.

The \modelname{}-14B comparison uses paired runs. The DeepSeek-V4-Pro results show the
same deep-order pattern, but its base and Starling runs were not collected
under a strictly paired setting and are therefore treated as corroborative
evidence rather than a precise cross-model comparison.

Overall, Starling improves higher-order belief reasoning when the task requires
explicit tracking of who observed which state and when. Its limited effect on
shallow tasks and its sensitivity to extraction errors support the selective
activation mechanism described in Section~\ref{sec:actio-starling}.

\subsubsection{Experience Accumulation: Selective Activation and Transfer}
\label{subsec:memory_is_effective_but_needs_selection}

We evaluate SAGE from two perspectives: whether selective activation and
fallback improve the use of accumulated experience, and whether the distilled
strategies transfer beyond the data on which they were produced. This
diagnostic experiment uses DeepSeek-V4-Flash under a fixed decoding protocol.
SAGE is curated only on the ToMBench training split and is frozen before
evaluation. HiToM and TactfulToM are used as held-out transfer benchmarks;
TactfulToM is split by conversation to avoid context leakage.

\begin{figure}[!tbp]
\centering
\includegraphics[width=0.70\linewidth]{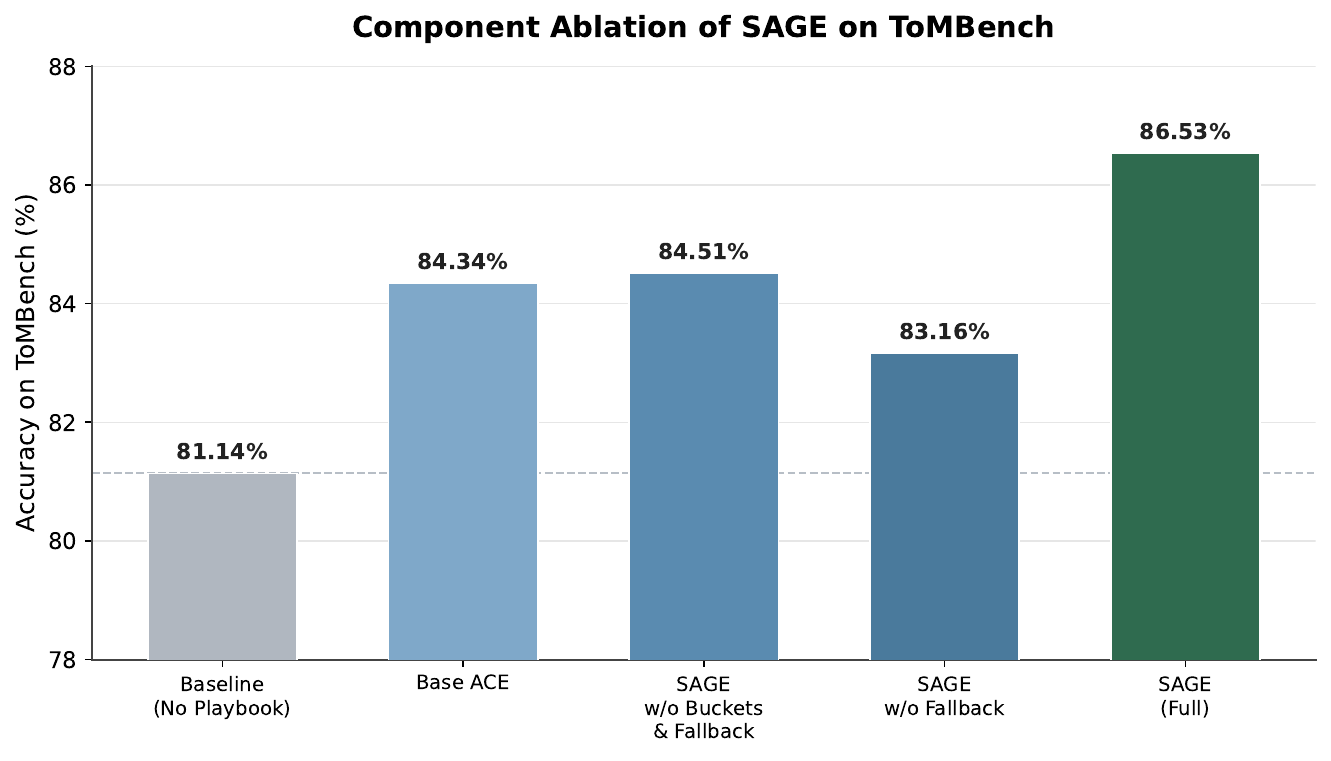}
\caption{Component ablation of SAGE on the ToMBench test split. Dimension
routing without fallback reduces effective recall, while the full activation
cascade achieves the highest accuracy.}
\label{fig:ablation}
\end{figure}

\paragraph{Selective activation requires a fallback mechanism.}
Figure~\ref{fig:ablation} shows the contribution of the main SAGE components on
ToMBench. The ACE curation loop improves accuracy from $81.14\%$ to
$84.34\%$, showing that distilled strategies provide useful guidance. Adding
the Selector yields a further increase to $84.51\%$. Dimension routing alone,
however, reduces accuracy to $83.16\%$, indicating that an incorrect or empty
route can suppress useful experience.

The full SAGE configuration adds the fallback cascade and reaches $86.53\%$,
a gain of $5.39$\,pp over the memory-free baseline and $3.37$\,pp over
dimension routing without fallback. Thus, structured selection improves memory
precision only when recall failures can back off to broader guidance.

\paragraph{The distilled strategies remain useful across datasets.}
Table~\ref{tab:generalization} evaluates the frozen strategy store without
further curation. SAGE improves the in-domain ToMBench result by $5.39$\,pp
and transfers positively to HiToM by $2.08$\,pp. It also produces a smaller
gain of $0.65$\,pp on TactfulToM. The held-out results are averaged over
three runs.

The stronger transfer to HiToM suggests that belief-tracking strategies
distilled from ToMBench are reusable on related higher-order reasoning tasks.
The smaller TactfulToM gain indicates that transfer weakens when the target
benchmark requires a different form of social-pragmatic reasoning. These
results provide evidence that SAGE retains cross-case reasoning guidance,
although its benefit depends on the alignment between stored experience and the
target task.

\begin{table}[t]
\centering
\small
\begin{tabular}{l c c c}
\toprule
\textbf{Benchmark} &
\textbf{Baseline} &
\textbf{+ SAGE} &
\textbf{$\Delta$} \\
\midrule
ToMBench\textsuperscript{\dag} & $81.14$ & $86.53$ & $+5.39$ \\
HiToM                         & $82.64$ & $84.72$ & $+2.08$ \\
TactfulToM                    & $80.95$ & $81.60$ & $+0.65$ \\
\bottomrule
\end{tabular}
\caption{Accuracy (\%) with a SAGE strategy store curated on the Chinese
ToMBench training split and frozen during evaluation. HiToM and TactfulToM
are held-out English benchmarks, and their results are averaged over three
runs. \textsuperscript{\dag}The ToMBench row is the in-domain reference.}
\label{tab:generalization}
\end{table}

Overall, SAGE accumulates reusable reasoning guidance rather than
episode-specific answers. Its gains depend on selective activation: rigid
dimension routing may reduce recall, whereas the fallback cascade preserves
coverage and produces the strongest in-domain result. Positive transfer to
held-out benchmarks further shows that part of the accumulated experience can
be reused across cases and datasets.

\section{Conclusion}
\label{sec:conclusion}

This report presents an integrated technical framework for social mind large language models. The central claim is that social intelligence must be measured, internalized, and grounded. \datasetname{} makes the capability space explicit through a psychology-grounded benchmark. \modelname{} improves the base model through targeted distillation and rubric-based reinforcement training. \agentname{} supports deployment-time reasoning with skills, memory, retrieval, and verification. The current results are promising but also diagnostic: strong social intelligence requires not only higher average scores, but also better measurement, finer process supervision, and more reliable control over when external support should be used.

 \section{Limitations and Future Work}
  \label{sec:limitations-future-work}

  The current report makes progress toward social intelligence as a measurable, trainable, and deployable
  capability, but several aspects of the target remain only partially covered. The present framework works best when
  social reasoning can be decomposed into labeled capabilities and evaluated in controlled scenarios. This is useful
  for diagnosis, yet social intelligence often appears in more compositional forms, where multiple mental states,
  goals, relations, and norms must be coordinated at once. Future work should therefore move beyond isolated
  capability labels toward richer social episodes in which the relevant interpretation only emerges after combining
  several cues.

  A second limitation lies in supervision granularity. Much of the current training signal still comes from final
  answers, short rationales, or selectively distilled traces, which are useful but do not fully represent how social
  reasoning unfolds over time. A promising direction is cognitive trajectory synthesis: constructing training data
  that preserves intermediate belief updates, perspective shifts, norm reweighting, and repair steps across a social
  episode. Such trajectories would provide richer process supervision and help the model learn not only what answer
  is correct, but how a socially grounded judgment is formed, revised, and stabilized.

  A third limitation concerns norm and context robustness. Social norms are not fixed labels; they depend on role,
  relationship history, institutional setting, cultural expectation, and the surrounding interaction. The current
  work handles this in a controlled way, but more robust social intelligence will require better behavior under
  ambiguity, conflict, and context shift. Future work should therefore emphasize calibration under uncertainty,
  clearer criteria for when a model should hedge or defer, and more explicit analysis of how normative judgments
  change across social frames.

  Taken together, the next stage should move from isolated measurement and supervision toward richer social
  episodes, more process-level training signals, and more robust behavior under changing relational and normative
  context.

\bibliography{bibs/train}
\bibliographystyle{plain}

\clearpage
\appendix

\section{Authors}
\label{sec:authors}
This project was completed through the joint effort of contributors across research design, technical implementation, evaluation and analysis, infrastructure and resource support, and report preparation. The full list of contributors is provided below in alphabetical order.

Ao Xiang, 
Bi Jingping, 
Chen Jiahui, Chen Lehan, Chen Yilin, Cheng Xueqi$^{*}$, 
Fan Yixing, 
Gan Kairong, Gao Haowen, Gao Jinhua, Gao Shuxuan, Gong Chang, Guo Jiafeng, Guo Ruijie$^{*}$, 
Han Zhouyu, He Guangfu, He Yichun, 
Jiang Shuo, Jing Shaoling, Jing Ya, 
Lei Chenhao, Lei Yan, Li Anqi, Li Chengao, Li Haoyu, Li Shitian, Liang Xinjian, Liu Zhaoge, Lyu Xingyu, 
Nie Zhuwei, 
Pang Liang, 
Quan Zeping, 
Shan Shiguang, Shen Huawei, 
Tang Xinran, Tian Feng, 
Wang Qian, Wang Ruiping, Wang Xiaohong, 
Xia Zaiyu, Xiao Yi, Xu Jiayuan, Xu Kehan, Xu Qianqian, Xu Tianyu, Xu Yongjun, Xue Yuanhai,
Yang Haoming, Yang Jun, Yao Di, Yu Xiaoming, 
Zhang Futong, Zhang Jie, Zhang Shixuan, Zhang Yuxuan, Zhao Xinyu, Zhao Zhuoran, Zhong Yunfei, Zhu Shengyu.

\vspace{0.3em}

$^{*}$Correspondence to:
\texttt{cxq@ict.ac.cn},
\texttt{guoruijie@ict.ac.cn}

\section{Detailed Task Paradigms}
\label{sec:appendix_tasks}

This appendix documents the full set of 71 fine-grained task paradigms organized under the three primitives and 17 secondary dimensions of the SocialMental framework. Each paradigm is presented with (i) a one-sentence definition of the target capability, (ii) the principal theoretical anchor(s), and (iii) the discrimination logic between adjacent paradigms within the same sub-dimension. Paragraph identifiers (e.g., A.1.1.3) follow the L1.L2.L3 hierarchy of the framework.

\subsection{Mentalizing Primitive}
\label{app:mentalizing}

The Mentalizing primitive (7 sub-dimensions, 30 paradigms) evaluates the recognition and inference of internal mental states---belief, desire, intention, emotion, perspective, empathy, personality/preference---in self and others. The seven ATOMS-derived ToM categories that CogToM \cite{tong2026cogtom} treats as its full taxonomy are entirely subsumed here.

\subsubsection{Belief (7 paradigms)}
\label{app:belief}

\subsubsubsection{Belief Attribution Inference.} Identifying the content, source (direct experience, hearsay, inference, instruction), and structure of an agent's belief at a specific moment, while distinguishing ``A believes X'' from ``X is true.'' Anchored in classical belief-attribution research \cite{tom}.

\subsubsubsection{Knowledge-State Distinction (Known / Unknown).} Discriminating among the four states---fully known, partially known, unknown, and erroneously known---of an agent's knowledge regarding a proposition, using epistemic markers (``I heard,'' ``I think,'' ``I'm not sure'') as evidence.

\subsubsubsection{Belief Inconsistency Detection.} Detecting logical contradictions or content conflicts between beliefs held by the same agent across turns or contexts, and choosing a clarification rather than confrontation strategy.

\subsubsubsection{First-Order False-Belief Reasoning.} The canonical Sally--Anne paradigm \cite{hitom}: predicting an agent's action based on their (factually incorrect) belief rather than the true state of the world.

\subsubsubsection{Belief-Certainty Inference.} A meta-cognitive variant: identifying the agent's confidence level in a belief and judging whether that confidence is calibrated to the available evidence.

\subsubsubsection{Cognitive-Bias Recognition.} Detecting that an agent's reasoning is shaped by a systematic bias drawn from one of five families: decision/judgment biases (anchoring, availability, framing), social-attribution biases (fundamental attribution error, actor-observer asymmetry), belief-maintenance biases (confirmation bias, motivated reasoning), memory biases (peak-end, consistency), and statistical-intuition biases (base-rate neglect, conjunction fallacy).

\subsubsubsection{Bidirectional Mental-State Tracking.} Maintaining parallel mental-state trajectories for two or more agents whose states recursively depend on each other (``A acts on a model of B's intent, and B acts on a model of A's expected behavior'').

\subsubsection{Desire (3 paradigms)}
\label{app:desire}

\subsubsubsection{Deep Desire / Motivation Abductive Inference.} Inferring the durable underlying motivation behind a sequence of behaviors, anchored in Maslow's needs hierarchy, self-determination theory (autonomy/competence/relatedness) \cite{bessi}, McClelland's tri-need theory, and Deci--Ryan's motivation continuum.

\subsubsubsection{Preference-Tendency Recognition.} Recognizing stable, cross-situational preferences and value priorities from choice patterns, anchored in Schwartz's 10-value model, Hofstede's six cultural dimensions, Haidt's six moral foundations, and the Inglehart materialist/post-materialist axis.

\subsubsubsection{Belief--Desire--Action Reasoning.} Integrating belief and desire to predict the most likely action under the BDI framework, and reverse-inferring belief--desire pairs from observed actions.

\subsubsection{Intention (3 paradigms)}
\label{app:intention}

\subsubsubsection{Immediate-Action Intention Recognition.} Identifying the immediate communicative goal of a speaker---informational, action-directive, social/relational, or affective---from form and context cues.

\subsubsubsection{Intention--Behavior Consistency Judgment.} Detecting whether surface speech form aligns with underlying intent (e.g., a request phrased as a question, an emotional need phrased as factual inquiry).

\subsubsubsection{Multi-Level Intention Hierarchy Parsing.} Decomposing an immediate intention into the higher-level goal structure it serves (e.g., ``how to write a resume'' $\rightarrow$ ``land a target role'' $\rightarrow$ ``career development'').

\subsubsection{Emotion (5 paradigms)}
\label{app:emotion}

\subsubsubsection{Basic Emotion Recognition.} Identifying the dominant emotion in an utterance using multi-layered taxonomies: Ekman's six basic emotions, Plutchik's eight-emotion wheel with dyadic compounds, dimensional PAD/Russell models, self-conscious emotions (shame, guilt, pride, embarrassment), moral emotions (righteous anger, gratitude, awe, contempt), and epistemic emotions (curiosity, confusion, insight).

\subsubsubsection{Composite Emotion Parsing.} Recognizing the simultaneous presence of multiple (sometimes contradictory) emotions in a single experience (``proud yet anxious,'' ``relieved yet bereaved'').

\subsubsubsection{Implicit Emotion Inference.} Inferring emotions that are not lexically marked, from topic avoidance, tone shifts, over-rationalized framings, or selective silence.

\subsubsubsection{Emotion-Regulation Strategy Understanding.} Classifying the strategy an agent is using to manage emotion, strictly anchored in Gross's five-family process model \cite{gross1998emotion}: situation selection, situation modification, attentional deployment (including rumination as a maladaptive variant), cognitive change (reappraisal, acceptance, self-distancing), and response modulation (suppression, expressive enhancement). Adaptive versus maladaptive judgments are required.

\subsubsubsection{Emotional Display Rules.} Recognizing context-, culture-, and role-governed rules about when emotions should be expressed, suppressed, masked, or amplified in social interaction.

\subsubsection{Perspective (5 paradigms)}
\label{app:perspective}

\subsubsubsection{Self--Other Cognitive-Boundary Recognition.} Maintaining a clean separation between what the model knows and what the user/agent in the scenario knows; specifically, avoiding the curse-of-knowledge failure mode where model-known information is implicitly assumed shared.

\subsubsubsection{Second- and Higher-Order Mental-State Attribution.} Representing recursive mental states, such as what one agent believes another agent knows, intends, or expects, and using those states to predict interactional behavior.

\subsubsubsection{Joint Attention.} Recognizing shared attentional focus and common-ground cues, including cases where agents fail to notice that attention is not jointly established.

\subsubsubsection{Capability-Boundary Awareness and Disclosure.} Identifying what an agent or system can and cannot know or do in a given situation, and disclosing those limits without overclaiming.

\subsubsubsection{Mistake Admission Integration.} Recognizing when new evidence requires admitting and correcting a prior error, then integrating the correction into subsequent reasoning or response.

\subsubsection{Empathy (4 paradigms)}
\label{app:empathy}

\subsubsubsection{Cognitive-Empathy Inference.} Inferring an agent's emotional response from their situation, role, and value orientation via perspective-taking---independent of emotional contagion. Theoretically anchored in Davis's IRI four-factor model (perspective-taking and fantasy as cognitive-empathy components).

\subsubsubsection{Affective-Empathy Response Generation.} Producing a response that mirrors and validates the agent's emotional experience, with intensity matched to the experience and structural sensitivity to whether the agent needs sympathy, compassion, or rational distance.

\subsubsubsection{Emotion-Contagion Mechanism Modeling.} Recognizing the bidirectional emotional transmission in interaction (the user's affect shaping the conversational climate; the system's tone shaping the user's affect) and deploying counter-contagious linguistic stabilization where appropriate.

\subsubsubsection{Well-being-Oriented Altruistic Motivation.} Inferring or producing help-oriented behavior driven by concern for another agent's welfare, rather than by self-interest or purely instrumental goals.

\subsubsection{Personality / Preference (3 paradigms)}
\label{app:personality}

\subsubsubsection{Trait-Based Preference Inference.} Inferring stable trait-level dispositions from interaction patterns, anchored in Big Five / HEXACO (with explicit attention to the H factor for honesty--humility), the Dark Tetrad (narcissism, Machiavellianism, psychopathy, sadism) for interpersonal-risk detection, attachment styles, and Vaillant's defense-mechanism hierarchy.

\subsubsubsection{Context-Conditioned Preference Realization Prediction.} Predicting how an inferred trait will translate into context-specific behavior, recognizing that traits express differently in work versus casual, familiar versus unfamiliar, and low- versus high-pressure situations.

\subsubsubsection{Dynamic User-Profile Updating.} Updating inferred preferences, traits, and interpersonal expectations as new evidence accumulates across turns, while avoiding stale or overgeneralized profiles.

\subsection{Strategic Navigation Primitive}
\label{app:strategic}

The Strategic Navigation primitive (6 sub-dimensions, 27 paradigms) evaluates the dynamic, multi-step processes of dyadic and group interaction.

\subsubsection{Communication (6 paradigms)}
\label{app:communication}

\subsubsubsection{Implicature Inference.} Recovering non-literal meaning generated by adherence to or violation of Gricean maxims \cite{grice}, with extensions to Sperber--Wilson relevance theory and Levinson's Q/I/M neo-Gricean refinements.

\subsubsubsection{Indirect Speech-Act Recognition.} Recognizing mismatch between speech-act form and function (``Can you close the window?'' as a directive, not a capability inquiry), grounded in Brown--Levinson politeness theory \cite{brown1987politeness} and Searle's indirect speech-act framework \cite{searle}, with attention to irony/sarcasm, hedging, euphemism, and metaphor.

\subsubsubsection{Speech-Act Intention Parsing.} Full Austin/Searle parsing of locution, illocution, and perlocution, supplemented with conversation-analytic mechanisms: turn-taking, adjacency-pair preference organization, repair-preference hierarchies, and discourse-marker meta-pragmatic functions.

\subsubsubsection{Conversational Presupposition Recognition.} Identifying the implicit propositions an utterance treats as common ground (``When did you stop smoking?'' presupposes ``you used to smoke''), and challenging presuppositions when warranted rather than accepting them.

\subsubsubsection{Social Signal Weighting and Focus Tracking.} Weighting social cues such as tone, status, salience, gaze, and topic focus, and tracking which cue should dominate interpretation at each point in the interaction.

\subsubsubsection{Historical Tracking and Anaphora Resolution.} Maintaining history-dependent meanings across turns, including resolving pronouns, ellipsis, and references to earlier events, commitments, or utterances.

\subsubsection{Negotiation (3 paradigms)}
\label{app:negotiation}

\subsubsubsection{Divergence-Localization Inference.} Identifying the type of underlying divergence (factual, value-based, interest-based, conceptual), since each type warrants a distinct resolution strategy.

\subsubsubsection{Negotiation-Strategy Prediction.} Predicting each party's likely strategy choice (competitive, collaborative, compromising, avoiding, accommodating) given their interests, power positions, relationship horizon, and cultural backdrop.

\subsubsubsection{Consensus-Path Reasoning.} Identifying viable paths from divergence to consensus---information sharing, interest reconstruction, framing change, incremental accumulation, third-party mediation---and their feasibility under given constraints.

\subsubsection{Game-Theoretic Reasoning (3 paradigms)}
\label{app:game}

\subsubsubsection{Zero-Sum vs.\ Non-Zero-Sum Distinction.} Diagnosing game structure, including specific game types: Stag Hunt (high-payoff cooperation vs.\ safe alternative), Battle of the Sexes (coordination with divergent preference), Chicken (mutual escalation risk), signaling games (separating vs.\ pooling equilibria), and the sequential/simultaneous, complete/incomplete information distinctions.

\subsubsubsection{Non-Dilemma (Win--Win) Game Reasoning.} Recognizing situations where the incentive structure already supports cooperation, and identifying the residual challenges (signaling, trust establishment, surplus allocation).

\subsubsubsection{Collective-Action Dilemma Reasoning.} Reasoning about prisoner-dilemma-type situations, free-rider problems, and tragedy of the commons, integrating evolutionary game theory \cite{axelrod} (Tit-for-Tat, Pavlov, generous variants), Nowak's five reciprocity mechanisms, Ostrom's eight design principles for commons governance, and critical-mass theory.

\subsubsection{Cooperation \& Conflict (7 paradigms)}
\label{app:cooperation_conflict}

\subsubsubsection{Conflict-Cause Identification.} Diagnosing the root cause(s) of conflict, anchored in Jehn's three-type distinction (task, relationship, process), Coser's realistic-versus-symbolic conflict distinction, and Galtung's conflict triangle of structural, attitudinal, and behavioral components \cite{galtung1969violence}.

\subsubsubsection{Conflict-Escalation Path Prediction.} Predicting the escalation trajectory along Glasl's nine-stage model (Hardening $\rightarrow$ Debate $\rightarrow$ Actions-not-Words $\rightarrow$ Images-and-Coalitions $\rightarrow$ Loss of Face $\rightarrow$ Threat Strategies $\rightarrow$ Limited Destruction $\rightarrow$ Fragmentation $\rightarrow$ Together-into-the-Abyss), with attention to escalation-spiral mechanisms (attribution bias, mirror perception, self-fulfilling prophecy, security dilemma).

\subsubsubsection{Conflict De-escalation Strategy Generation.} Generating stage-appropriate de-escalation strategies (active cooling, framing reconstruction, third-party introduction, interest-versus-position reframing) given the current Glasl phase.

\subsubsubsection{Cooperation-Motivation Recognition.} Recognizing the type of motivation underlying cooperation (complementary capacity, scale economy, defensive alliance, affective bond, shared identity, normative obligation) and inferring its stability under change.

\subsubsubsection{Alliance-Stability Judgment.} Assessing alliance robustness against internal (allocation fairness, trust, commitment credibility) and external (common-threat persistence, substitute availability, exit costs) pressures.

\subsubsubsection{Cooperation-Breakdown Prediction.} Identifying the early-warning signals of a deteriorating cooperative relationship and the threshold conditions for collapse.

\subsubsubsection{Relationship Rupture Recognition and Repair.} Detecting signs of relational rupture, such as loss of trust, face threat, or withdrawal, and identifying appropriate repair moves.

\subsubsection{Trust \& Deception (5 paradigms)}
\label{app:trust_deception}

\subsubsubsection{Trust-Formation Condition Judgment.} Evaluating which of the four trust dimensions (competence trust, benevolence trust, integrity trust, value-congruence trust) are met by available evidence, and identifying the limiting dimension.

\subsubsubsection{Deception-Intent Recognition.} Recognizing the structural intent of deception, classified across the lie typology (white lies, prosocial lies, self-protection, self-gain, pathological lying) and aware of Levine's Truth-Default Theory (the difficulty of deception detection without trigger events)\cite{bessi}.

\subsubsubsection{Lie Detection.} Concrete statement-level credibility assessment using cognitive cues (cognitive-load markers, narrative linearity, missing self-corrections), emotional cues (Ekman micro-expressions, ``duping delight,'' guilt leakage), and control cues (over-rehearsed delivery, excessive eye contact as over-compensation).

\subsubsubsection{Manipulation-Strategy Recognition.} Identifying manipulation tactics organized into three families: emotional manipulation (FOG triad, gaslighting, love bombing, intermittent reinforcement, triangulation, silent treatment), social manipulation (DARVO, negging, social isolation, information control), and cognitive manipulation (false dilemma, straw man, moving goalposts, ad hominem). Trust-repair pathways (apology, compensation, commitment, penalty) are also evaluated.

\subsubsubsection{Adversarial Input Recognition and Constructive Redirection.} Detecting manipulative, hostile, or adversarial inputs and redirecting the interaction constructively while preserving task relevance and social quality.

\subsubsection{Group Dynamics (3 paradigms)}
\label{app:group}

\subsubsubsection{Conformity-Pressure Judgment.} Recognizing normative versus informational conformity pressure (Asch paradigm), with extensions to obedience to authority (Milgram), groupthink (Janis), and minority influence (Moscovici).

\subsubsubsection{Group-Polarization Reasoning.} Predicting the direction and magnitude of opinion shift in homogeneous-group discussion (risky vs.\ cautious shift), and recognizing related group-dynamic mechanisms: bystander effect (Latan\'e--Darley five stages), deindividuation (Zimbardo), social facilitation vs.\ inhibition (Triplett--Zajonc), social loafing (Ringelmann effect).

\subsubsubsection{Social-Comparison Inference.} Recognizing upward and downward social comparison \cite{festinger1954theory} and inferring its impact on self-evaluation, emotion, and motivation.

\subsection{Norm Internalization \& Dynamic Balance Primitive}
\label{app:norm}

The Norm Internalization primitive (4 sub-dimensions, 14 paradigms) evaluates how structural and cultural rules shape mind and behavior, with explicit emphasis on the dynamic weighing of conflicting norms rather than static norm classification.

\subsubsection{Norm Internalization (4 paradigms)}
\label{app:norm_internalization}

\subsubsubsection{Social-Norm Recognition.} Identifying which norms are operative in a given setting, distinguishing descriptive (what people do) versus injunctive (what is approved) norms \cite{cialdini}, and applying Bicchieri's conditional-preference model and Cialdini's focus theory of normative conduct.

\subsubsubsection{Norm-Violation Detection.} Detecting norm violations and classifying severity using Kohlberg's three-level moral-development framework \cite{kohlberg1969stage}, Turiel's social-domain theory (moral vs.\ conventional vs.\ personal), and Haidt's social-intuitionist model \cite{haidt2001emotional}.

\subsubsubsection{Cross-Cultural Norm-Transfer Reasoning.} Reasoning about norm conflict and adaptation in cross-cultural contact, anchored in Berry's four-quadrant acculturation model, Oberg's four-stage culture-shock model, and Hall's high-/low-context distinction.

\subsubsubsection{Contextual Appropriateness Judgment.} Judging whether an utterance or action is socially appropriate given the setting, timing, role relationship, cultural background, and interactional stakes.

\subsubsection{Identity and Role (2 paradigms)}
\label{app:identity}

\subsubsubsection{Identity--Behavior Association Inference.} Mapping identity labels (professional, familial, group-membership) to typical behavior scripts in a culture-specific way.

\subsubsubsection{Social Role-Expectation Inference.} Identifying the role-specific expectation system (teacher, parent, leader, friend, new employee) operative in a situation and evaluating the agent's awareness of and compliance with it, including tensions arising when a single agent simultaneously occupies multiple roles with conflicting expectations.

\subsubsection{Power and Institution (5 paradigms)}
\label{app:power}

\subsubsubsection{Status--Power Asymmetry Inference.} Identifying interpersonal status asymmetry across position, expertise, age, and reputation, and inferring its effect on language choice and behavioral space.

\subsubsubsection{Power Asymmetry Recognition (Bases).} Identifying power bases at a more general level, anchored in French--Raven's six bases (coercive, reward, legitimate, referent, expert, informational) \cite{french1959bases}, Foucault's disciplinary, biopolitical, and governmentality forms of power \cite{foucault}, Gramsci's cultural hegemony, and Lukes's three-dimensional power.

\subsubsubsection{Power-Exercise Strategy Reasoning.} Reasoning about top-down power exercise using Yukl--Falbe's 11 influence tactics (rational persuasion, inspirational appeals, consultation, collaboration, ingratiation, personal appeals, exchange, coalition, legitimating, pressure, informing) and Keltner's power-paradox of empathy erosion.

\subsubsubsection{Power-Resistance Mechanism Modeling.} Reasoning about bottom-up resistance, integrating Scott's full ``weapons of the weak'' typology \cite{scott}: foot-dragging, pilfering, feigned ignorance, gossip, mystification, religious heterodoxy, and the public-vs.-hidden transcript distinction; collective resistance through political opportunity structure and contention repertoires.

\subsubsubsection{Institutional vs.\ Individual Power Distinction.} Distinguishing power deriving from institutional position from power deriving from individual attributes, and applying the distinction to choose between institutional-reform and interpersonal-management strategies.

\subsubsection{Group Boundary (3 paradigms)}
\label{app:group_boundary}

\subsubsubsection{In-group / Out-group Distinction.} Recognizing the activation of in-group / out-group categorization and its effects on attribution, trust, and cooperation, with attention to the arbitrariness of the categorization base.

\subsubsubsection{Group-Membership Determination.} Evaluating membership criteria (ascribed, achieved, subjective-affiliative) and resolving boundary disputes in immigration, minority-identity, and subcultural contexts.

\subsubsubsection{Stereotype / Prejudice / Discrimination Differentiation.} Distinguishing the three layers---cognitive (stereotype), affective (prejudice), and behavioral (discrimination)---anchored in Allport's prejudice scale, Fiske--Cuddy's Stereotype Content Model (warmth $\times$ competence), the modern-versus-traditional prejudice distinction, implicit-bias measurement, and structural/institutional discrimination.

\section{Seed-Generation Prompt Example}
\label{app:expansion_prompt}

We illustrate the seed-generation prompt using paradigm \textbf{1.1.1 Belief Attribution Inference} as a representative example. Every paradigm has its own structured prompt following the same template; only the dimension-specific content (definitions, context constraints, question requirements, and distractor types) varies. The original prompts are written in Chinese; we present an English translation for readability.

\begin{quote}\small

\textbf{Dimension Definition.}
\textit{L1 Primitive --- Mentalizing:} the capacity to represent, infer, and track mental states (beliefs, desires, intentions, emotions, etc.) of self and others.
\textit{L2 Dimension --- Belief:} the internal representation an agent holds about world states, possibly diverging from reality; requires tracking not only what an agent \emph{knows} but also what they \emph{believe they know}.
\textit{L3 Paradigm --- 1.1.1 Belief Attribution Inference:} the ability to identify the content, source (direct experience, hearsay, inferential reasoning, testimony), and structure of an agent's belief, while strictly distinguishing ``the agent believes X'' from ``X is true.''

\textbf{Difficulty Requirements} (expert-level).
\emph{Scenario complexity:} at least three subjects each forming beliefs through heterogeneous and cross-referenced information sources; at least one subject holds a false belief that is internally coherent within their evidence frame; dynamic belief revision must occur at one or more time points; some beliefs must be embedded as implicit presuppositions rather than explicit statements.
\emph{Distractor difficulty:} the strongest distractors must differ from the correct option only along the belief-attribution / fact-confusion or source-attribution boundary; distractors must exploit cognitive biases such as Belief--Reality Confusion, Curse of Knowledge, or Source-Monitoring Error; no option may be eliminable by surface common-sense alone.

\textbf{Context Design.}
The context must cover at least three of the four belief-source types; contain at least one false-belief subject whose error is internally consistent with their information frame; include implicit-presupposition signals that require pragmatic inference; place the action in a realistic multi-party social context (project reporting, cross-department collaboration, multi-specialist consultation) with named characters and concrete dialogue/behavioral cues; contain a belief-update juncture where new information triggers belief revision in at least one subject.

\textbf{Four Questions per Context.}
\begin{itemize}
    \item \textit{Single-choice} --- belief-content attribution: identify what a specific subject believes at a specific moment, requiring the model to distinguish ``subject believes X'' from ``X is true.''
    \item \textit{Multiple-choice} --- belief-source attribution: identify the formation source of each subject's belief across the four source types, requiring parallel source-monitoring.
    \item \textit{Judgment-with-rationale} --- implicit-belief extraction: infer an unspoken but operative belief from a subject's questioning style, behavioral choice, or causal reasoning, and judge its consistency with the subject's explicit statements.
    \item \textit{Open-ended} --- belief-graph construction and dynamic tracking: synthesize the belief content, source, and structure of all subjects into a complete belief graph; trace the belief-revision trajectory of a focal subject after new information arrives; analyze the formation mechanism of belief biases along the content--source--structure interaction.
\end{itemize}

\textbf{Distractor Design.}
\begin{itemize}
    \item \textit{Belief--Reality Confusion:} equate objective fact with subject belief.
    \item \textit{Curse of Knowledge:} project the evaluator's global information onto the subject as their belief.
    \item \textit{Source-Monitoring Error:} correctly identify belief content but misattribute its source.
    \item \textit{Surface-Statement Equivalence:} equate an agent's overt utterance with their full belief, ignoring implicit presuppositions.
    \item \textit{Belief-Update Neglect (Belief Perseverance Bias):} use the agent's initial belief in place of the post-update belief, or assume full revision while ignoring anchoring residue.
\end{itemize}

\textbf{Output schema.} Each generated test case must be returned as a single JSON object containing: \texttt{context} (800--1200 Chinese characters), \texttt{questions} (a list of four items, one per question type), and per-question fields \texttt{prompt}, \texttt{options}/\texttt{conclusion\_options}, \texttt{answer}, and \texttt{reasoning} for the judgment and open-ended types.

\end{quote}

\section{Evaluation Prompts}
\label{app:eval_prompts}

We illustrate the zero-shot evaluation protocol with the prompt for the single-choice question type. The prompts for the other three item types (multiple-choice, judgment-with-rationale, open-ended application) and the LLM-as-judge templates used to score open-ended and judgment answers follow the same JSON-only structured-output format. All prompts are bilingual; we show the English translation here.

\begin{quote}\small\ttfamily
{[}System{]} You are a strict answering machine. Output only JSON. Do not output any additional text.

\medskip
{[}User{]} Answer the following single-choice question. Output only JSON, in the form \{"answer": "B"\}.

\medskip
\textbf{[Scenario]}\\
\{q.context\}

\medskip
\textbf{[Question]} \{q.prompt\}

\medskip
\textbf{[Options]}\\
A. \{Option A\}\\
B. \{Option B\}\\
C. \{Option C\}\\
D. \{Option D\}
\end{quote}

\noindent To control for option-position bias, every question is evaluated five times: four trials with the four cyclic rotations of the option order, and one additional trial with a random shuffle distinct from the rotations. The average accuracy across the five trials is reported as the per-paradigm score.

\end{document}